%% file: mlaut.tex
\documentclass[oneside,english]{article}

\input{cfg/config.tex}
\input{cfg/global_macros.tex}

\title{Machine Learning Automation Toolbox (MLaut)}

\author[1,2]{
Viktor Kazakov
\thanks{\url{viktor.kazakov.18@ucl.ac.uk}}
}

\author[3,4]{
Franz J.~Kir\'{a}ly
\thanks{\url{f.kiraly@ucl.ac.uk}}
}

\affil[1]{
Lab of Innovative Finance and Technology (LIFTech),\newline
Civil, Environmental and Geomatic Engineering Department,\newline
University College London,
Gower Street,
London WC1E 6BT, United Kingdom
}

\affil[2]{
European Bank for Reconstruction and Development\newline
One Exchange Square,
London EC2A 2JN, United Kingdom
}

\affil[3]{
Department of Statistical Science,
University College London,\newline
Gower Street,
London WC1E 6BT, United Kingdom
}

\affil[4]{The Alan Turing Institute,\newline
The British Library, Kings Cross,
London NW1 2DB, United Kingdom
}

\begin{document}
	
	\maketitle

\begin{abstract}
	In this paper we present MLaut (Machine Learning AUtomation Toolbox) for the python data science ecosystem. MLaut automates large-scale evaluation and benchmarking of machine learning algorithms on a large number of datasets. MLaut provides a high-level workflow interface to machine algorithm algorithms, implements a local back-end to a database of dataset collections, trained algorithms, and experimental results, and provides easy-to-use interfaces to the scikit-learn and keras modelling libraries. Experiments are easy to set up with default settings in a few lines of code, while remaining fully customizable to the level of hyper-parameter tuning, pipeline composition, or deep learning architecture.

As a principal test case for MLaut, we conducted a large-scale supervised classification study in order to benchmark the performance of a number of machine learning algorithms - to our knowledge also the first larger-scale study on standard supervised learning data sets to include deep learning algorithms. While corroborating a number of previous findings in literature, we found (within the limitations of our study) that deep neural networks do not perform well on basic supervised learning, i.e., outside the more specialized, image-, audio-, or text-based tasks.
\end{abstract} 

%\newpage

\tableofcontents{}
\newpage

	\input{inputs/introduction.tex}
	\input{inputs/supervised_learning.tex}
	\input{inputs/benchmarking.tex}
	\input{inputs/api_design_and_features.tex}
	\input{inputs/ml_comparison_study.tex}

	\newpage

    \bibliographystyle{plainnat}
	\bibliography{MLAUT}

	\newpage
	\appendix

	\input{inputs/appendix.tex}

\end{document}

%% file: cfg/config.tex
\usepackage[utf8]{inputenc}

\usepackage{courier}

\usepackage{mathtools}
\usepackage{amsmath,amssymb,amsthm}
\usepackage{thmtools}

\usepackage{enumitem}

\usepackage{pdflscape} % for rotating pages
\usepackage{geometry}
\usepackage{blindtext}
\usepackage{adjustbox}

\usepackage{multirow}
\usepackage{array}
\usepackage{rotating}
\usepackage{makecell}
\usepackage{capt-of}
\usepackage{tabularx}
\usepackage{longtable}

\usepackage{hyperref}
\usepackage{url}

\usepackage{bbm}

\usepackage{authblk}

\usepackage[numbers]{natbib}

\usepackage{color}

\usepackage{graphicx}
\graphicspath{ {images/} }

\usepackage{booktabs}

\usepackage{xcolor}
\usepackage{tcolorbox}

\usepackage{listings}
\usepackage[]{algorithm2e} %for writing algoithms

\definecolor{str_color}{RGB}{125,105,97}
\definecolor{code_num_color}{rgb}{0.5,0.5,0.5}
\definecolor{comments_color}{RGB}{77,102,74}
\definecolor{keyword_color}{RGB}{77,52,77}
\DeclareFontFamily{OT1}{pzc}{}
\DeclareFontShape{OT1}{pzc}{m}{it}{<-> s * [1.10] pzcmi7t}{}
\DeclareMathAlphabet{\mathpzc}{OT1}{pzc}{m}{it}
\usepackage{caption}

%% file: cfg/global_macros.tex
\newcommand{\code}[1]{{\fontfamily{pcr}\selectfont #1}}

\newcommand{\PredFunction}{f}

\newcommand{\TrainingSet}{\mathcal{D}}
\newcommand{\TestSet}{\mathcal{T}}

\newcommand{\NewX}{X^*}
\newcommand{\NewY}{Y^*}

\newcommand{\LossFunction}{L}

\newcommand{\sterror}{\widehat{SE}}

\newcommand{\PredictedLabel}{\hat{Y}}

\newcommand{\sgn}{\operatorname{sgn}}

\newcommand{\Var}{\operatorname{Var}}

\newcommand{\calX}{\mathcal{X}}
\newcommand{\calY}{\mathcal{Y}}

\newcommand{\EE}{\ensuremath{\mathbb{E}}}

\newcommand{\RR}{\ensuremath{\mathbb{R}}}

\newcommand{\OOne}{\ensuremath{\mathbbm{1}}}

\makeatletter
\providecommand*{\diff}%
        {\@ifnextchar^{\DIfF}{\DIfF^{}}}
\def\DIfF^#1{%
        \mathop{\mathrm{\mathstrut d}}%
                \nolimits^{#1}\gobblespace
}
\def\gobblespace{%
        \futurelet\diffarg\opspace}
\def\opspace{%
        \let\DiffSpace\!%
        \ifx\diffarg(%
                \let\DiffSpace\relax
        \else
                \ifx\diffarg\[%
                        \let\DiffSpace\relax
                \else
                        \ifx\diffarg\{%
                                \let\DiffSpace\relax
                        \fi\fi\fi\DiffSpace}
\makeatother

\newcommand{\myToprule}{\noalign{\hrule height 1.0pt}}
\newcommand{\myMidrule}{\noalign{\hrule height 1.0pt}}
\newcommand{\myBottomrule}{\noalign{\hrule height 1.0pt}}

%% file: inputs/introduction.tex
\section{Introducing MLaut}
\label{sec:intro}

MLaut~\cite{mlaut_2018} is a modelling and workflow toolbox in python, written with the aim of simplifying large scale benchmarking of machine learning strategies, e.g., validation, evaluation and comparison with respect to predictive/task-specific performance or runtime. Key features are:

\begin{enumerate}
	\itemsep-0.2em
	\item[(i)] automation of the most common workflows for benchmarking modelling strategies on multiple datasets including statistical post-hoc analyses, with user-friendly default settings
	\item[(ii)] unified interface with support for scikit-learn strategies, keras deep neural network architectures, including easy user extensibility to (partially or completely) custom strategies
	\item[(iii)] higher-level meta-data interface for strategies, allowing easy specification of scikit-learn pipelines and keras deep network architectures, with user-friendly (sensible) default configurations
	\item[(iv)] easy setting up and loading of data set collections for local use (e.g., data frames from local memory, UCI repository, openML, Delgado study, PMLB)
	\item[(v)] back-end agnostic, automated local file system management of datasets, fitted models, predictions, and results, with the ability to easily resume crashed benchmark experiments with long running times
\end{enumerate}

MLaut may be obtained from pyPI via \texttt{pip install mlaut}, and is maintained on GitHub at \texttt{github.com/alan-turing-institute/mlaut}. A Docker implementation of the package is available on Docker Hub via \texttt{docker pull kazakovv/mlaut}.

\subsection*{Note of caution: time series and correlated/associated data samples}
MLaut implements benchmarking functionality which provides statistical guarantees under assumption of either independent data samples, independent data sets, or both. This is mirrored in Section~\ref{sec:theory.setting} by the crucial mathematical assumptions of statistical independence (i.i.d.~samples), and is further expanded upon in Section~\ref{sec:theory.performance}.\\
{\bf In particular, it should be noted that naive application of the validation methodology implemented in MLaut to samples of time series, or other correlated/associated/non-independent data samples (within or between datasets), will in general violate the validation methodologies' assumptions, and may hence result in misleading or flawed conclusions about algorithmic performance.}\\
The BSD license under which MLaut is distributed further explicitly excludes liability for any damages arising from use, non-use, or mis-use of MLaut (e.g., mis-application within, or in evaluation of, a time series based trading strategy).

\subsection{State-of-art: modelling toolbox and workflow design}
\label{sec:intro.toolboxes}

A hierarchy of modelling designs may tentatively be identified in contemporary machine learning and modelling ecosystems, such as the python data science environment and the R language:
\begin{enumerate}
	\itemsep-0.2em
	\item[Level 1.] implementation of specific methodology or a family of machine learning strategies, e.g., the most popular packages for deep learning, Tensorflow \cite{martin_abadi_tensorflow:_2015}, MXNet \cite{chen_mxnet:_2015}, Caffe \cite{shelhamer_caffe_nodate} and CNTK \cite{seide_cntk:_2016}.
	\item[Level 2.] provision of a unified interface for methodology solving the same ``task'', e.g., supervised learning aka predictive modelling. This is one core feature of the Weka~\cite{jagtap_census_2013}, scikit-learn~\cite{pedregosa_scikit-learn:_2011} and Shogun~\cite{sonnenburg_shogun_2010} projects which both also implement level 1 functionality, and main feature of the caret~\cite{wing_caret:_2018} and mlr~\cite{bischl_mlr:_2016} packages in R which provides level 2 functionality by external interfacing of level 1 packages.
	\item[Level 3.] composition and meta-learning interfaces such as tuning and pipeline building, more generally, first-order operations on modelling strategies. Packages implementing level 2 functionality usually (but not always) also implement this, such as the general hyper-parameter tuning and pipeline composition operations found in scikit-learn and mlr or its mlrCPO extension. Keras~\cite{chollet_keras_2015} has abstract level 3 functionality specific to deep learning, Shogun possesses such functionality specific to kernel methods.
	\item[Level 4.] workflow automation of higher-order tasks performed with level 3 interfaces, e.g., diagnostics, evaluation and comparison of pipeline strategies. Mlr is, to our knowledge, the only existing modelling toolbox with a modular, class-based level 4 design that supports and automates re-sampling based model evaluation workflows. The Weka GUI and module design also provides some level 4 functionality.\\
	A different type of level 4 functionality is automated model building, closely linked to but not identical with benchmarking and automated evaluation - similarly to how, mathematically, model selection is not identical with model evaluation. Level 4 interfaces for automated model building also tie into level 3 interfaces, examples of automated model building are implemented in auto-Weka~\cite{hall2009weka}, auto-sklearn~\cite{feurer2015efficient}, or extensions to mlrCPO~\cite{thomas2018automatic}.
\end{enumerate}

In the Python data science environment, to our knowledge, there is currently no widely adopted solution with level 4 functionality for evaluation, comparison, and benchmarking workflows. The reasonably well-known skll~\cite{skll} package provides automation functionality in python for scikit-learn based experiments but follows an unencapsulated scripting design which limits extensibility and usability, especially since it is difficult to use with level 3 functionality from scikit-learn or state-of-art deep learning packages.

Prior studies conducting experiments which are level 4 use cases, i.e., large-scale benchmarking experiments of modelling strategies, exist for supervised classification, such as \cite{fernandez-delgado_we_2014, wainer_comparison_2016}. Smaller studies, focusing on a couple of estimators trained on a small number of datasets have also been published \cite{huang_comparing_2003}.
However, to the best of our knowledge: none of the authors released a toolbox for carrying out the experiments; code used in these studies cannot be directly applied to conduct other machine learning experiments; and, deep neural networks were not included as part of the benchmark exercises.

At the current state-of-art, hence, there is a distinct need for level 4 functionality in the scikit-learn and keras ecosystems. Instead of re-creating the mlr interface or following a GUI-based philosophy such as Weka, we have decided to create a modular workflow environment which builds on the particular strengths of python as an object oriented programming language, the notebook-style user interaction philosophy of the python data science ecosystem, and the contemporary mathematical-statistical state-of-art with best practice recommendations for conducting formal benchmarking experiments - while attempting to learn from what we believe works well (or not so well) in mlr and Weka.

\subsection{Scientific contributions}

MLaut is more than a mere implementation of readily existing scientific ideas or methods. We argue that the following contributions, outlined in the manuscript, are scientific contributions closely linked to its creation:

\begin{enumerate}
	\itemsep-0.2em
	\item[(1)] design of a modular ``level 4'' software interface which supports the predictive model validation/comparison workflow, a data/model file input/output back-end, and an abstraction of post-hoc evaluation analyses, at the same time.
	\item[(2)] a comprehensive overview of the state-of-art in statistical strategy evaluation, comparison and comparative hypothesis testing on a collection of data sets. We further close gaps in said literature by formalizing and explicitly stating the kinds of guarantees the different analyses provide, and detailing computations of related confidence intervals. %[cite section]
	\item[(3)] as a principal test case for MLaut, we conducted a large-scale supervised classification study in order to benchmark the performance of a number of machine learning algorithms, with a key  sub-question being whether more complex and/or costly algorithms tend to perform better on real-world datasets. On the representative collection of UCI benchmark datasets, kernel methods and random forests perform best. %[cite section]
	\item[(4)] as a specific but quite important sub-question we empirically investigated whether common off-shelf deep learning strategies would be worth considering as a default choice on the ``average'' (non-image, non-text) supervised learning dataset. The answer, somewhat surprising in its clarity, appears to be that they are not - in the sense that alternatives usually perform better. However, on the smaller tabular datasets, the computational cost of off-shelf deep learning architectures is also not as high as one might naively assume. This finding is also subject to a major caveat and future confirmation, as discussed in Section~\ref{subsection:estimators.keras} and Section~\ref{subsection:conclusions}.
\end{enumerate}

Literature relevant to these contribution will be discussed in the respective sections.

%\newpage

\subsection{Overview: usage and functionality}
\label{subsec:ease_of_use}

We present a short written demo of core MLaut functionality and user interaction, designed to be convenient in combination with jupyter notebook or scripting command line working style. Introductory jupyter notebooks similar to below may be found as part of MLaut's documentation~\cite{mlaut_2018}.

The first step is setting up a database for the dataset collection, which has to happen only once per computer and dataset collection, and which we assume has been already stored in a local MLaut HDF5 database. The first step in the core benchmarking workflow is to define hooks to the database input and output files:

\begin{lstlisting}[language=Python]
input_io = data.open_hdf5(...) #path to input HDF5 file
out_io = data.open_hdf5(...) #path to output HDF5 file
\end{lstlisting}

After the hooks are created we can proceed to preparing fixed re-sampling splits (training/test) on which all strategies are evaluated. By default MLaut creates a single evaluation split with a uniformly sampled $\dfrac{2}{3}$ of the data for training and $\dfrac{1}{3}$ for testing.

\begin{lstlisting}[language=Python]
data.split_datasets(hdf5_in=..., hdf5_out=..., dataset_paths=...)
\end{lstlisting}

For a simple set-up, a standard set of estimators that come with sensible parameter defaults can be initialized. Advanced commands allow to specify hyper-parameters, tuning strategies, keras deep learning architectures, scikit-learn pipelines, or even fully custom estimators.

\begin{lstlisting}[language=Python]
est = ['RandomForestClassifier','BaggingClassifier']
estimators = instantiate_default_estimators(estimators=est)
>>> estimators
<mlaut.estimators.ensemble_estimators.Random_Forest_Classifier>
<mlaut.estimators.ensemble_estimators.Bagging_Classifier>
\end{lstlisting}

The user can now proceed to running the experiments. Training, prediction and evaluation are separate; partial results, including fitted models and predictions, are stored and retrieved through database hooks. This allows intermediate analyses, and for the experiment to easily resume in case of a crash or interruption. If this happens, the user would simply need to re-run the code above and the experiment will continue from the last checkpoint, without re-executing prior costly computation.

\begin{lstlisting}[language=Python]
>>> orchest.run(modelling_strategies=estimators)
RandomForestClassifier trained on dataset 1
RandomForestClassifier trained on dataset 2
...
\end{lstlisting}

The last step in the pipeline is executing post-hoc analyses for the benchmarking experiments. The \code{AnalyseResults} class allows to specify performance quantifiers to be computed and comparison tests to be carried out, based on the intermediate computation data, e.g., predictions from all the strategies.

\begin{lstlisting}[language=Python]
analyze.prediction_errors(score_accuracy, estimators)
\end{lstlisting}

The \code{prediction\_errors()} method returns two sets of results: \code{errors\_per\_estimator} dictionary which is used subsequently in further statistical tests and \code{errors\_per\_dataset \_per\_estimator\_df} which is a dataframe with the loss of each estimator on each dataset that can be examined directly by the user.

We can also use the produced errors in order to perform the statistical tests for method comparison. The code below shows an example of running a t-test.

\begin{lstlisting}[language=Python]
_, t_test_df = analyze.t_test(errors_per_estimator)
>>> t_test_df
Estimator 1       Estimator 2
t_stat p_val      t_stat p_val
Estimator 1 ...    ...        ...    ...
Estimator 2 ...    ...        ...    ...
...
\end{lstlisting}

Data frames or graphs resulting from the analyses can then be exported, e.g., for presentation in a scientific report.

\subsection*{Authors contributions}
MLaut is part of VK's PhD thesis project, the original idea being suggested by FK. MLaut and this manuscript were created by VK, under supervision by FK. The design of MLaut is by VK, with suggestions by FK. Sections~\ref{sec:intro}, \ref{sec:theory} and~\ref{Sec:benchmarking} were substantially edited by FK before publication, other sections received only minor edits (regarding content). The benchmark study of supervised machine learning strategies was conducted by VK.

\subsection*{Acknowledgments}
We thank Bilal Mateen for critical reading of our manuscript, and especially for suggestions of how to improve readability of Section~\ref{sec:theory.performance}.

FK acknowledges support by The Alan Turing Institute under EPSRC grant EP/N510129/1. 

%% file: inputs/supervised_learning.tex
\newpage

\section{Benchmarking supervised learning strategies on multiple datasets - generative setting}
\label{sec:theory}

This section introduces the mathematical-statistical setting for the mlaut toolbox - supervised learning on multiple datasets. Once the setting is introduced, we are able to describe the suite of statistical benchmark post-hoc analyses that mlaut implements, in Section~\ref{Sec:benchmarking}. \\

\subsection{Informal workflow description}

Informally, and non-quantitatively, the workflow implemented by mlaut is as follows: multiple prediction strategies are applied to multiple datasets, where each strategy is fitted to a training set and queried for predictions on a test set. From the test set predictions, performances are computed: performances by dataset, and also overall performances across all datasets, with suitable confidence intervals. For performance across all datasets, quantifiers of comparison (``is method A better than method B overall?'') are computed, in the form statistical (frequentist) hypothesis tests, where p-values and effect sizes are reported.\\

The remainder of this Section~\ref{sec:theory} introduces the \emph{generative setting}, i.e., statistical-mathematical formalism for the data sets and future situations for which performance guarantees are to be obtained. The reporting and quantification methodology implemented in the mlaut package is described in Section~\ref{Sec:benchmarking} in mathematical language, usage and implementation of these in the mlaut package is described in Section~\ref{sec:api}.\\

From a statistical perspective, it should be noted that only a single train/test split is performed for validation. This is partly due to simplicity of implementation, and partly due to the state-of-art's incomplete understanding of how to obtain confidence intervals or variances for re-sampled performance estimates. Cross-validation strategies may be supported in future versions.\\

A reader may also wonder about whether, even if there is only a single set of folds, should there not be three folds per split (or two nested splits), into tuning-train/tuning-test/test\footnote{What we call the ``tuning-test fold'' is often, somewhat misleadingly, called a ``validation fold''. We believe the latter terminology is misleading, since it is actually the final test fold which validates the strategy, not second fold.}. The answer is: yes, if tuning via re-sample split of the training set is performed. However, in line with current state-of-art understanding and interface design, tuning is considered as part of the prediction strategy. That is, the tuning-train/tuning-test split is strategy-intrinsic. Only the train/test split is extrinsic, and part of the evaluation workflow which mlaut implements; a potential tuning split is encapsulated in the strategy. This corresponds with state-of-art usage and understanding of the wrapper/composition formalism as implemented for example with GridSearchCV in sklearn.

\subsection{Notational and mathematical conventions}

To avoid confusion between quantities which are random and non-random, we always explicitly say if a quantity is a random variable. Furthermore, instead of declaring the type of a random variable, say $X$, by writing it out as a measurable function $X:\Omega\rightarrow \calX$, we say ``$X$ is a random variable taking values in $\calX$'', or abbreviated ``$X$ t.v.in $\calX$'', suppressing mention of the probability space $\Omega$ which we assume to be the same for all random variables appearing.

This allows us easily to talk about random variables taking values in certain sets of functions, for example a prediction functional obtained from fitting to a training set.
Formally, we will denote the set of functions from a set $\calX$ to a set $\calY$ by the type theoretic arrow symbol $\calX \rightarrow \calY$, where bracketing as in $[\calX\rightarrow \calY]$ may be added for clarity and disambiguation. E.g., to clarify that we consider a function valued random variable $f$, we will say for example ``let $f$ be a random variable t.v.in $[\calX\rightarrow \calY]$''.

An observant reader familiar with measure theory will notice a potential issue (others may want to skip to the next sub-section): the set $[\calX\rightarrow \calY]$ is, in general, not endowed with a canonical measure. This is remedied as follows: if we talk about a random variable taking values in $[\calX\rightarrow \calY]$, it is assumed that the image of the corresponding measurable function $X: \Omega \rightarrow [\calX\rightarrow \calY]$, which may not be all of $[\calX\rightarrow \calY]$, is a measurable space. This is, for example, the case we substitute training data random variables in a deterministic training functional $f$, which canonically endows the image of $f$ with the substitution push-forward measure.

\subsection{Setting: supervised learning on multiple datasets}
\label{sec:theory.setting}

We introduce mathematical notation to describe $D$ datasets, and $K$ prediction strategies.
As running indices, we will consistently use $i$ for the $i^{th}$ dataset, $j$ for $j^{th}$ (training or test) data point in a given data set, and $k$ for the $k^{th}$ estimator.

The data in the $i$-th dataset are assumed to be sampled from mutually independent, generative/population random variables $(X^{(i)},Y^{(i)})$, taking values in feature-label-pairs $\calX^{(i)} \times \calY$, where either $\calY = \RR$ (regression) or $\calY$ is finite (classification). In particular we assume that the label type is the same in all datasets.

The actual data are i.i.d.~samples from the population $(X^{(i)},Y^{(i)})$, which for notational convenience we assume to be split into a training set $\TrainingSet_i = \left((X_{tr,1}^{(i)}, (X_{tr,1}^{(i)}), \dots, ((X_{tr,N_i}^{(i)}, (X_{tr,N_i}^{(i)})\right)$ and a test set $\TestSet_i = \left((X_{1}^{(i)}, Y_{1}^{(i)}), \dots, (X_{M_i}^{(i)}, Y_{M_i}^{(i)})\right).$ Note that the training and test set in the $i$-th dataset are, formally, not ``sets'' (as in common diction) but ordered tuples of length $N_i$ and $M_i$. This is for notational convenience which allows easy reference to single data points.
By further convention, we will write $Y_{\star}^{(i)} := \left(Y_{1}^{(i)},\dots, Y_{M_i}^{(i)}\right)$ for the ordered tuple of test labels.

On each of the datasets, $K$ different prediction strategies are fitted to the training set: these are formalized as random prediction functionals $\PredFunction_{i,k}$ t.v.in $[\calX^{(i)} \rightarrow \calY]$, where $i=1\dots D$ and $k=1\dots K$. We interpret $\PredFunction_{i,k}$ as the fitted prediction functional obtained from applying the $k$-th prediction strategy on the $i$-th dataset where it is fitted to the training set.

Statistically, we make mathematical assumptions to mirror the reasonable intuitive assumptions that there is no active information exchange between different strategies, a copies of a given strategy applied to different data sets: we assume that the random variable $\PredFunction_{i,k}$ may depend on the training set $\TrainingSet_i$, but is independent of all other data, i.e., the test set $\TestSet_i$ of the $i$-th dataset, and training and test sets of all the other datasets. It is further assumed that $\PredFunction_{i,k}$ is independent of all other fitted functionals $\PredFunction_{i',k'}$ where $i'\neq i$ and $k'$ is entirely arbitrary. It is also assumed that $\PredFunction_{i,k}$ is conditionally independent of all $\PredFunction_{i,k'},$ where $k'\neq k$, given $\TrainingSet_i$.

We further introduce notation for predictions $\PredictedLabel_{j,k}^{(i)} := \PredFunction_{i,k}(X_{j}^{(i)}),$ i.e., $\PredictedLabel_{j,k}^{(i)}$ is the prediction made by the fitted prediction functional $\PredFunction_{i,k}$ for the actually observed test label $Y_{j}^{(i)}$.

For convenience, the same notation is introduced for the generative random variables, i.e., $\PredictedLabel_{k}^{(i)} := \PredFunction_{i,k}(X^{(i)}).$
Similarly, we denote by $\PredictedLabel^{(i)}_{\star,k}:=(\PredictedLabel^{(i)}_{1,k},\dots, \PredictedLabel^{(i)}_{M_i,k})$ the random vectors of length $M_i$ whose entries are predictions for full test sample, made by method $k$.

\subsection{Performance - which performance?}
\label{sec:theory.performance}

%While most benchmarking exercises do report a ``performance'', it is crucial to distinguish the kind of performance that is computed. While the freedom to choose among different quantifiers (e.g., loss functions) is folklore, the kind of \emph{guarantee} the performance estimate gives is often less clear, i.e., which scenario the performance estimate should be considered a guarantee for.

Benchmarking experiments produce performance and comparison quantifiers for the competitor methods. It is important to recognise that these quantifiers are computed to create guarantees for the methods' use on putative \emph{future data}. These guarantees are obtained based on mathematical theorems such as the central limit theorem, applicable under empirically justified assumptions. It is crucial to note that mathematical theorems allow establishing performance guarantees on future data, despite the future data not being available to the experimenter at all. It is also important to note that the future data for which the guarantees are created are different from, and in general not identical to, the test data.

Contrary to occasional belief, performance on the test data in isolation is empirically not useful: without a guarantee it is unrelated to the argument of algorithmic effectivity the experimenter wishes to make.

%It should also be noted the freedom to choose among different performance quantifiers (e.g., loss functions) is folklore, what is less well understood is how the performance estimate relates to the kind of guarantee given, i.e., which scenario the performance estimate should be considered a guarantee for.

While a full argument usually \emph{does }involve computing performance on a statistically independent test set, the argumentative reason for this best practice is more subtle than being of interest by itself. It is a consequence of ``prediction'' performance on the training data not being be a fair proxy for performance on future data. Instead, ``prediction'' on an unseen (statistically independent) test set is a fair(er) proxy, as it allows for formation of performance guarantees on future data: the test set being unseen allows to leverage the central limit theorems for this purpose.

In benchmark evaluation, it is hence crucial to make precise the relation between the testing setting and the application case on future data - there are two key types of distinctions on the future data application case:
\begin{enumerate}
\itemsep-0.2em
\item[(i)] whether in the scenario, a fitted prediction function is to be re-used, or whether it is re-fitted on new data (potentially from a new data source).
\item[(ii)] whether in the scenario, the data source is identical with the source of one of the observed datasets, or whether the source is merely a source from the same population as the data sources observed.
\end{enumerate}

Being precise about these distinctions is, in fact, practically crucial: similar to the best practice of not testing on the training set, one needs to be careful about whether a \emph{data source}, or a \emph{fitted strategy} that will occur in the future test case has already been observed in the benchmarking experiment, or not.

We make the above mathematically precise (a reader interested only in an informal explanation may first like skip forward to the subsequent paragraph).

To formalize ``re-use'', distinction (i) translates to conditioning on the fitted prediction functionals $f_{i,k}$, or not. Conditioning corresponds to prior observation, hence having observed the outcome of the fitting process, therefore ``re-using'' $f_{i,k}$. Not doing so corresponds to sampling again from the random variable, hence ``re-fitting''.

To formalize the ``data source'' distinction, we will assume an i.i.d.~process $P$ (taking values in joint distributions over $\calX^{(i)} \times \calY$ also selected at random), generating distributions according to which population laws are distributed, i.e., $P_1,\dots, P_D \sim P$ is an i.i.d.~sample. The $i$-th element of this sample, $P_i$ is the (generating) data source for the $i$-th data set i.e., $(X^{(i)},Y^{(i)})\sim P_i$. We stress that $P_i$ takes values in distributions, i.e., $P_i$ is a distribution which is itself random\footnote{Thus, the symbol $\sim$ is used here in its common ``distribution'' and not ``distribution of random variable'' meaning which are usually confounded by abuse of notation} and from which data are generated.
In this mathematical setting, the distinction (ii) then states whether the guarantee applies for data sampled from $(X^{(i)},Y^{(i)})$ with a specific $i$, or instead data sampled from $(X^{*},Y^{*})\sim P.$ The former is ``data from the already observed $i$-th source, the latter is ``data from a source similar to, but not identical to, the observed source''.
If the latter is the case, the same generative principle is applied to yield a prediction functional $f_{*,k}$, drawn i.i.d.~from a hypothetical generating process which yielded the $f_{i,k}$ on the $i$-th dataset. We remain notationally consistent by defining $\PredictedLabel_k^{*}:=f_{*,k}(X^{*}).$

For {\bf intuitive} clarity, let us consider an {\bf example}: three supervised classification methods, a random forest, logistic regression, and the baseline ``predicting the majority class'' are benchmarked on 50 datasets, from 50 hospitals, one dataset corresponding to observations in exactly one hospital. Every dataset is a sample of patients (data frame rows) for which as variables (data frame rows) the outcome (= prediction target and data frame column) therapy success yes/no for a certain disease is recorded, plus a variety of demographic and clinical variables (data frame columns) - where what is recorded differs by hospital.

A benchmarking experiment may be asked to produce a performance quantifier for one of the following three distinct key future data scenarios:
\begin{enumerate}
\itemsep-0.2em
\item[(a)] re-using the trained classifiers (e.g., random forest), trained on the training data of hospital 42, to make predictions on future data observed in hospital 42.
\item[(b)] (re-)fitting a given classifier (e.g., random forest) to new data from hospital 42, to make predictions on further future data observed in hospital 42.
\item[(c)] obtaining future data from a new hospital 51, fitting the classifiers to that data, and using the so fitted classifiers to make predictions on further future data observed from hospital 51.
\end{enumerate}

It is crucial to note that both performances and guarantees may (and in general will) differ between these three scenarios. In hospital 42, a random forests may outperform logistic regression and the baseline, while in hospital 43 nothing outperforms the baseline. The behaviour and ranking of strategies may also be different, depending on whether classifiers are re-used, or re-fitted. This may happen in the same hospital, or when done in an average unseen hospital.
Furthermore, the same qualitative differences as for observed performances may hold for the precision of the statistical guarantees obtained from performances in a benchmarking experiment: the sample size of patients in a given hospital may be large enough or too small to observe a significant difference of performances in a given hospital, while the sample size of hospitals is the key determinant of how reliable statistical guarantees about performances and performance differences for unseen hospitals are.

In the subsequent, we introduce abbreviating {\bf terminology} for denoting the distinctions above: for (i), we will talk about \emph{re-used} (after training once) and \emph{re-trained} (on new data) prediction algorithm. For (ii), we will talk about \emph{seen} and \emph{unseen} data sources. Further, we will refer to the three future data scenarios abbreviatingly by the letters (a), (b), and (c). By terminology, in these scenarios the algorithm is: (a) re-used on seen sources, (b) re-trained on seen sources, and (c) re-trained on an unseen source (similar to but not identical to seen sources).

It should be noted that it is impossible to re-use an algorithm on an unseen source, by definition of the word ``unseen'', hence the hypothetical fourth combination of the two dichotomies re-used/re-trained and unseen/seen is logically impossible.

\subsection{Performance quantification}

Performance of the prediction strategy is measured by a variety of quantifiers which compare predictions for the test set with actual observations from the test set, the ``ground truth''. Three types of quantifiers are common:

\begin{enumerate}
\itemsep-0.2em
\item[(i)] Average loss based performance quantifiers, obtained from a comparison of one method's predictions and ground truth observations one-by-one. An example is the mean squared error on the test set, which is the average squared loss.
\item[(ii)] Aggregate performance quantifiers, obtained from a comparison of all of a given method's predictions with all of the ground truth observations. Examples are sensitivity or specifity.
\item[(iii)] Ranking based performance quantifiers, obtained from relative performance ranks of multiple methods, from a ranked comparison against each other. These are usually leveraged for comparative hypothesis tests, and may or may not involve computation of ranks based on average or aggregate performances as in (i) and (ii). Examples are the Friedman rank test to compare multiple strategies.
\end{enumerate}

The three kinds of performance quantifiers are discussed in more detail below.

\subsubsection{Average based performance quantification}
\label{subsection:loss_functions}
\label{sec:theory.performance.avg}
For this, the most widely used method is a loss (or score) function $\LossFunction: \calY\times\calY \rightarrow \RR$, which compares a single prediction (by convention the first argument) with a single observation (by convention the second argument).

Common examples for such loss/quantifier functions are listed below in Table~\ref{table:lossfunctions}.

\begin{table}[h]
  \centering
  \begin{tabular}{| c | c | c |}
    \myToprule
    task & name & loss/quantifier function \\
    \myMidrule
    \multirow{1}{*}{classification (det.)} & MMCE & $L(\widehat{y},y) = 1- \OOne [y = \widehat{y}]$ \\
    \myMidrule
    \multirow{4}{*}{regression} & squared loss & $L(\widehat{y},y) = (y-\widehat{y})^2$ \\
     & absolute loss & $L(\widehat{y},y) = |y-\widehat{y}|$ \\
     & Q-loss & $L(\widehat{y},y) = \alpha\cdot m (\widehat{y},y) + (1-\alpha)\cdot m(y,\widehat{y})$ \\
     & & where $m(x,z)=\min(x-z,0)$  \\

               %                     & sensitivity = TPR, TP & & ``yes''\\
                %                    & specifity = TNR, TN & & ``no''\\
%     \myMidrule
%    \multirow{2}{*}{classification (prob.)} & log-loss & $L(p,y_*) = -\log p(y_*)$  & the label proportions\\
%     & squared loss & $L(p,y_*) = \left(1- p(y_*)\right)^2$ & the label proportions\\
%     & Brier loss & $L(p,y_*) = \left(1- p(y_*)\right)^2+\sum_{y\neq y_*}p(y)^2$ & the label proportions\\
    \myBottomrule
  \end{tabular}
\caption{List of some popular loss functions to measure prediction goodness (2nd column) used in the most frequent supervised prediction scenarios (1st column). Above, $y$ and $\widehat{y}$ are elements of $\calY$. For classification, $\calY$ is discrete; for regression, $\calY=\RR$. The symbol $\OOne [A]$ evaluates to $1$ if the boolean expression $A$ is true, otherwise to $0$.}
\label{table:lossfunctions}
\end{table}
%	For a discussion of how this framework can be applied to the multi class supervised classification setting please refer to \cite{machart_confusion_2012}.

In direct alignment with the different future data scenarios discussed in Section~\ref{sec:theory.performance}, the distributions of three generative random variables are of interest:
\begin{enumerate}
\itemsep-0.2em
\item[(a)] The conditional random variable $L(f_{i,k}(X^{(i)}),Y^{(i)})|f_{i,k} = L(\PredictedLabel_k^{(i)},Y^{(i)})|f_{i,k}$, the loss when predicting on future data from the $i$-th data source, when re-using the already trained prediction functional $f_{i,k}$. Note that formally, through conditioning $f_{i,k}$ is implicitly considered constant (not random), therefore reflects re-use of an already trained functional.

\item[(b)] The random variable $L(f_{i,k}(X^{(i)}),Y^{(i)}) = L(\PredictedLabel_k^{(i)},Y^{(i)})$, the loss when re-training method $k$ on training data from the $i$-th data source, and predicting labels on future data from the $i$-th data source. Without conditioning, no re-use occurs, and this random variable reflects repeating the whole random experiment including re-training of $f_{i,k}$.

\item[(c)] The random variable $L(f_{*,k}(X^{*}),Y^{*}) = L(\PredictedLabel_k^{*},Y^{*})$, the loss when training method $k$ on a completely new data source, and predicting labels on future data from the same source as that dataset.
\end{enumerate}

The distributions of the above random variables are generative, hence unknown. In practice, the validation workflow estimates summary statistics of these. Of particular interest in the mlaut workflow are related expectations, i.e., (arithmetic) population average errors. We list them below, suppressing notational dependency on $L$ for ease of notation:

\begin{enumerate}
\itemsep-0.2em
\item[(a.1)] $\eta_{i,k} := \EE[L(\PredictedLabel_k^{(i)},Y^{(i)})|f_{i,k}]$, the (training set) \emph{conditional} expected generalization error of (a re-used) $f_{i,k}$, on data source $k$.
\item[(a.2)] $\overline{\eta}_{k} := \frac{1}{D}\sum_{i=1}^D\EE[L(\PredictedLabel_k^{(i)},Y^{(i)})]|(f_{i,k})_{i=1\dots D}$, the \emph{conditional} expected generalization error of the (re-used) $k$-th strategy, averaged over all \emph{seen} data sources.
\item[(b)] $\varepsilon_{i,k} := \EE[L(\PredictedLabel_k^{(i)},Y^{(i)})]$, the \emph{unconditional} expected generalization error of (a re-trained) $f_{i,k}$, on data source $k$.
\item[(c)] $\varepsilon^*_k := \EE[L(\PredictedLabel_k^{*},Y^{*})]$, the expected generalization error on a typical (unseen) data source.
\end{enumerate}

%
%\begin{enumerate}
%\itemsep-0.2em
%\item[(a.1)] $\eta_{i,k} := \EE[L(\PredictedLabel_k^{(i)},Y^{(i)})|f_{i,k}]$, the (training set) \emph{conditional} expected generalization error of $f_{i,k}$, on data source $k$.
%\item[(a.2)] $\overline{\eta}_{k} := \frac{1}{D}\sum_{i=1}^D\EE[L(\PredictedLabel_k^{(i)},Y^{(i)})]$, the \emph{conditional} expected generalization error of the $k$-th strategy, averaged over all \emph{seen} data sources.
%\item[(b.1)] $\varepsilon_{i,k} := \EE[L(\PredictedLabel_k^{(i)},Y^{(i)})]$, the \emph{unconditional} expected generalization error of $f_{i,k}$, on data source $k$.
%\item[(b.2)]
%$\overline{\varepsilon}_{k} := \frac{1}{D}\sum_{i=1}^D\EE[L(\PredictedLabel_k^{(i)},Y^{(i)})]$, the \emph{unconditional} expected generalization error of the $k$-th strategy, averaged over all \emph{seen} data sources.
%\item[(c)] $\varepsilon^*_k := \EE[L(\PredictedLabel_k^{*},Y^{*})]$, the expected generalization error on a typical data source.
%\end{enumerate}

%It should be noted that $\eta_{i,k}$ and $\overline{\eta}_{k}$ are random quantities, but conditionally constant once the respective $f_{i,k}$ are known (e.g., once $f_{i,k}$ has been trained). It holds that $\eta_{i,k} = \EE[\varepsilon_{i,k}],$ and $\overline{\varepsilon}_{k} = \EE[\overline{\eta}_{k}]$.\\
%It should also be noted that $\varepsilon^*_k$ and $\overline{\varepsilon}_{k}$ are, in general, different.\\

It should be noted that $\eta_{i,k}$ and $\overline{\eta}_{k}$ are random quantities, but conditionally constant once the respective $f_{i,k}$ are known (e.g., once $f_{i,k}$ has been trained). It further holds that $\eta_{i,k} = \EE[\varepsilon_{i,k}].$\\

The mlaut toolbox currently implements estimators for only two of the above three future data situations - namely, only for situations (a: re-used, seen) and (c: re-trained, unseen), i.e., estimators for all quantities with the exception of $\varepsilon_{i,k}$. The reason for this is that for situation (b: re-trained, seen), at the current state of literature it appears unclear how to obtain good estimates, that is, with provably favourable statistical properties independent of the data distribution or the algorithmic strategy. For situations (a) and (c), classical statistical theory may be leveraged, e.g., mean estimation and frequentist hypothesis testing.

It should also be noted that $\varepsilon_{i,k}$ is a \emph{single dataset} performance quantifier rather than a \emph{benchmark} performance quantifier, and therefore outside the scope of mlaut's core use case. While $\eta_{i,k}$ is also a single dataset quantifier, it is easy to estimate en passant while estimating the benchmark quantifier $\overline{\eta}_{k}$, hence included in discussion as well as in mlaut's functionality.

\subsubsection{Aggregate based performance quantification}
\label{sec:theory.performance.aggr}

A somewhat less frequently used alternative are aggregate loss/score functions $\LossFunction: (\calY\times\calY)^+ \rightarrow \RR$, which compare a tuple of predictions with a tuple of observations in a way that is not expressible as a mean loss such as in Section~\ref{subsection:loss_functions}. Here, by slight abuse of notation, $(\calY\times\calY)^+$ denotes tuples of $\calY$-pairs, of fixed length. The use of the symbol $\LossFunction$ is discordant with the previous section and assumes a case distinction on whether an average or an aggregate is used.

The most common uses of aggregate performance quantifiers are found in deterministic binary classification, as entries of the classification contingency table. These, and further common examples are listed below in Table~\ref{table:lossfunctionsaggr}.

\begin{table}[h]
  \centering
  \begin{tabular}{| c | c | c |}
    \myToprule
    task & name & loss/quantifier function \\
    \myMidrule
    \multirow{4}{*}{classification (det., binary)}
    & sensitivity, recall & $L(\widehat{y},y) = \langle y, \widehat{y}\rangle /\|y\|_1$ \\
    & specificity & $L(\widehat{y},y) = \langle \OOne-y, \OOne-\widehat{y}\rangle /\|1-y\|_1$ \\
    & precision, PPV & $L(\widehat{y},y) = \langle y, \widehat{y}\rangle /\|\widehat{y}\|_1$ \\
    & F1 score & $L(\widehat{y},y) =2 \langle y, \widehat{y}\rangle /\langle\OOne, \widehat{y}+ y\rangle$\\
    \myMidrule
    \multirow{1}{*}{regression} & root mean squared error & $L(y,\widehat{y}) = \|y-\widehat{y}\|_2$ \\

    \myBottomrule
  \end{tabular}
\caption{List of some popular aggregate performance measues of prediction goodness (2nd column) used in the most frequent supervised prediction scenarios (1st column). Overall, the test sample size is assumed to be $M$. Hence above, both $y$ and $\widehat{y}$ are elements of $\calY^M$. For binary classification, $\calY = \{0,1\}$ without loss of generality, $1$ being the ``positive'' class; for regression, $\calY =\RR$. By convention, $y$ denotes the true value, and $\widehat{y}$ denotes the prediction. We use vector notation for brevity: $\langle.,.\rangle$ denotes the vector/scalar/inner product, $\|.\|_p$ denotes the $p$-norm, and $\OOne$ the $M$-vector with entries all equal to the number $1$.}
\label{table:lossfunctionsaggr}
\end{table}

As before, for the different future data scenarios in Section~\ref{sec:theory.performance}, the distributions of three types of generative random variables are of interest. The main complication is that aggregate performance metrics take multiple test points and predictions as input, hence to specify a population performance one must specify a test set size. In what follows, we will fix a specific test set size, $M_i$, for the $i$-th dataset. Recall the notation $Y^{(i)}_{\star}$ for the full vector of test labels on data set $i$. In analogy, we abbreviatingly denote by $\PredictedLabel^{(i)}_{\star,k}$ random vectors of length $M_i$ whose entries are predictions for full test sample, made by method $k$, i.e., having as the $j$-th entry to predictions $\PredictedLabel^{(i)}_{j,k}$, as introduced in Section~\ref{sec:theory.setting}. Similarly, we denote by $Y^{*}_{\star}$ and $\PredictedLabel^{*}_{\star}$ vectors whose entries, are i.i.d.~from the data generating distribution of the new data source, and both of length $M^*$, which is by assumption the sampling distribution of the $M_i$.

The population performance quantities of interest can be formulated in terms of the above:

\begin{enumerate}
\itemsep-0.2em
\item[(a.1)] $\eta_{i,k} := \EE[L({\PredictedLabel}_{\star,k}^{(i)},Y^{(i)}_{\star})|f_{i,k}]$, the (training set) \emph{conditional} expected generalization error of (a re-used) $f_{i,k}$, on data source $k$.
\item[(a.2)] $\overline{\eta}_{k} := \frac{1}{D}\sum_{i=1}^D\EE[L({\PredictedLabel}_{\star,k}^{(i)},Y^{(i)}_{\star})]|(f_{i,k})_{i=1\dots D}$, the \emph{conditional} expected generalization error of the (re-used) $k$-th strategy, averaged over all \emph{seen} data sources.
\item[(b)] $\varepsilon_{i,k} := \EE[L({\PredictedLabel}_{\star,k}^{(i)},Y^{(i)}_{\star})]$, the \emph{unconditional} expected generalization error of (a re-trained) $f_{i,k}$, on data source $k$.
\item[(c)] $\varepsilon^*_k := \EE[L(\PredictedLabel_{\star,k}^{*},Y_{\star}^{*})]$, the expected generalization error on a typical (unseen) data source.
\end{enumerate}

As before, the future data situations are (a: re-used algorithm, seen sources), (b: re-trained, seen), and (c: re-trained, unseen). In the general setting, the expectations in (a) and (b) may or may not converge to sensible values as $M_i$ approaches infinity, depending on properties of $L$. General methods of estimating these depend on availability of test data, which due to the complexities arising and the currently limited state-of-art are outside the scope of mlaut. This unfortunately leaves benchmarking quantity $\overline{\eta}_{k}$ outside the scope for aggregate performance quantifiers. For (c), classical estimation theory of the mean applies.

\subsubsection{Ranking based performance quantification}
\label{sec:theory.performance.rank}

Ranking based approaches consider, on each dataset, a performance ranking of the competitor strategies with respect to a chosen raw performance statistic, e.g., an average or an aggregate performance such as RMSE or F1-score. Performance assessment is then based on the rankings - in the case of ranking, this is most often a comparison, usually in the form of a frequentist hypothesis test.
Due to the dependence of the ranking on a raw performance statistic, it should always be understood that ranking based comparisons are with respect to the chosen raw performance statistic, and may yield different results for different raw performance statistics.

Mathematically, we introduce the population performances in question. Denote $L_{k}^{(i)}:=L(\PredictedLabel_k^{(i)},Y_{j}^{(i)})$ in the case the raw statistic being an average, and denote $L_{k}^{(i)}:=L({\PredictedLabel}_{\star,k}^{(i)},Y^{(i)})_{\star})$ in case it is an aggregate (on the RHS using notation of the respective previous Sections~\ref{sec:theory.performance.avg} and~\ref{sec:theory.performance.aggr}). The distribution of $L_{k}^{(i)}$ models generalization performance of the $k$-th strategy on the $i$-th dataset.

We further define rankings $R_k^{(i)}$ as the order rank of $L_{k}^{(i)}$ within the tuple $(L_{1}^{(i)},\dots, L_{K}^{(i)})$, i.e., the ranking of the performance $L_{k}^{(i)}$ within all $K$ strategies' performances on the $i$-th dataset.

Of common interest in performance quantification and benchmark comparison are the average ranks, i.e., ranks of a strategy averaged over datasets. The population quantity of interest is the expected average rank on a typical dataset, i.e.,
$r_k := \EE[R_k^{(*)}],$ where $R_k^{(*)}$ is the population variable corresponding to sample variables $R_k^{(i)}$.
It should be noted that the average rank depends not only on what the $k$-th strategy is or does, but also on the presence of the other $(K-1)$ strategies in the benchmarking study - hence it is not an absolute performance quantifier for a single method, but a relative quantifier, to be seen in the context of the competitor field.

Common benchmarking methodology of the ranking kind quantifies relative performance on the data sets observed in the sense of future data scenario (b) or (c), where the performance is considered including (re-)fitting of the strategies. 

%% file: inputs/benchmarking.tex
\newpage

\section{Benchmarking supervised learning strategies on multiple datasets - methods}

\label{Sec:benchmarking}

We now describe the suite of performance and comparison quantification methods implemented in the mlaut package. It consists largely of state-of-art of model comparison strategies for the multiple datasets situation, supplemented by our own constructions based on standard statistical estimation theory where appropriate. References and prior work will be discussed in the respective sub-sections.
mlaut supports the following types of benchmark quantification methodology and post-hoc analyses:
\begin{enumerate}
\itemsep-0.2em
\item[(i)] loss-based performance quantifiers, such as mean squared error and mean absolute error, including confidence intervals.
\item[(ii)] aggregate performance quantifiers, such as contingency table quantities (sensitivity, specifity) in classification, including confidence intervals.
\item[(iii)] rank based performance quantifiers, such as average performance rank.
\item[(iv)] comparative hypothesis tests, for relative performance of methods against each other.
\end{enumerate}

The exposition uses notation and terminology previously introduced in Section~\ref{sec:theory}. Different kinds of quantifiers (loss and/or rank based), and different kinds of future performance guarantees (trained vs re-fitted prediction functional; seen vs unseen sources), as discussed in Section~\ref{sec:theory.performance}, may apply across all types of benchmarking analyses.

Which of these is the case, especially under which future data scenario the guarantee given is supposed to hold, will be said explicitly for each, and should be taken into account by any use of the respective quantities in scientific argumentation.

Practically, our recommendation is to consider which of the future data scenarios (a), (b), (c) a guarantee is sought for, and whether evidencing differences in rank, or differences in absolute performances, are of interest.

\subsection{Average based performance quantifiers and confidence intervals}

For average based performance quantifiers, performances and their confidence intervals are estimated from the sample of loss/score evaluates. We will denote the elements in this sample by $L_{j,k}^{(i)}:=L(\PredictedLabel_k^{(i)},Y_{j}^{(i)})$ (for notation on RHS see Section~\ref{sec:theory.performance.avg}
). Note that, differently from the population quantities, there are three (not two) indices: $k$ for the strategy, $i$ for the dataset, and $j$ for which test set point we are considering.

\begin{table}[h]
  \centering
  \begin{tabular}{|c | c c | c c |}
    \myToprule
    estimate & estimates & f.d.s. & standard error estimate & CLT in\\
    \myMidrule
    $\widehat{\eta}_{i,k} := \frac{1}{M_i}\sum_{j=1}^{M_i} L_{j,k}^{(i)}$ & $\eta_{i,k}$ & (a) & $\sqrt{\frac{\widehat{v}_{i,k}}{M_i}},\;\mbox{where}\;\widehat{v}_{i,k}:=\frac{\sum_{j=1}^{M_i} (L_{j,k}^{(i)}-\widehat{\eta}_{i,k})^2}{M_i-1}$& $M_i$ \\
    $\widehat{\eta}_{k} := \frac{1}{D}\sum_{i=1}^{D} \widehat{\eta}_{i,k}$ & $\overline{\eta}_{k}$ & (a) & $\frac{1}{D}\sqrt{\sum_{i=1}^D\frac{\widehat{v}_{i,k}}{M_i}}$ & $D,M_1,\dots, M_D$ \\
    $\widehat{\varepsilon}^*_{k}:= \widehat{\eta}_{k}$ & $\varepsilon^*_{k}$ & (c) & $\sqrt{\frac{\widehat{w}_{k}}{D}},\;\mbox{where}\;\widehat{w}_{k}:=\frac{\sum_{i=1}^{D} (\widehat{\eta}_{i,k} - \widehat{\varepsilon}^*_{k})^2}{D-1}$ & $D$ \\

    \myBottomrule
  \end{tabular}
\caption{Table of basic estimates of expected loss, with confidence intervals. First column = definition of the estimate. Second column = the quantity which is estimated by the estimate. Third column = which future data scenario (f.d.s.) estimate and confidence intervals give a guarantee for. Fourth column = standard error estimate (normal approximation) for the estimate in the first column, e.g., to construct frequentist confidence intervals. Fifth column = quantities governing central limit theorem (CLT) asymptotics for the confidence intervals.}
\label{table:perfest}
\end{table}

Table~\ref{table:perfest} presents a number of expected loss estimates with proposed standard error estimates. As all estimates are mean estimates of independent (or conditionally independent) quantities, normal approximated, two-sided confidence intervals may be obtained for any of the quantities in the standard way, i.e., at $\alpha$ confidence as the interval
$$[\widehat{\theta} + \Phi^{-1}(\alpha/2) \sterror, \widehat{\theta} - \Phi^{-1}(\alpha/2) \sterror]$$
where $\widehat{\theta}$ is the respective (mean) estimate and $\sterror$ is the corresponding standard error estimate.

Note that different estimates and confidence intervals arise through the different future data scenarios that the guarantee is meant to cover - see Sections~\ref{subsection:loss_functions} and~\ref{sec:theory.performance} for a detailed explanation how precisely the future data scenarios differ in terms of re-fitting/re-using the prediction functional, and obtaining performance guarantees for predictive use on an unseen/seen data source.
In particular, choosing a different future data scenario may affect the confidence intervals even though the midpoint estimate is the same: the midpoint estimates $\widehat{\varepsilon}^*_{k}$ and $\widehat{\varepsilon}_{k}$ coincide, but the confidence intervals for future data scenario (c), i.e., new data source and the strategy is re-fitted, are usually wider than the confidence intervals for the future data scenario (a), i.e., already seen data source and no re-fitting of the strategy.

Technically, all expected loss estimates proposed in Table~\ref{table:perfest} are (conditional) mean estimates. The confidence intervals for $\widehat{\eta}_{i,k}$ and $\widehat{\varepsilon}^*_{k}$ are obtained as standard confidence intervals for a (conditionally) independent sample mean: $\widehat{\varepsilon}^*_{k}$ is considered to be the mean of the independent samples $\widehat{\eta}_{i,k}$ (varying over $i$). $\eta_{i,k}$ is considered to be the mean of the conditionally independent samples $L_{j,k}^{(i)}$ (varying over $j$, and conditioned on $f_{i,k}$). Confidence intervals for $\widehat{\eta}_{k}$ are obtained averaging the estimated variances of independent summands $\widehat{\eta}_{i,k}$, which corresponds to the plug-in estimate obtained from the equality $\Var(\widehat{\eta}_{k}) = \frac{1}{D^2}\sum_{i=1}^{D} \Var(\widehat{\eta}_{i,k})$ (all variances conditional on the $f_{i,k}$).

\subsection{Aggregate based performance quantifiers and confidence intervals}

For aggregate based performance quantifiers, performances and their confidence intervals are estimated from the sample of loss/score evaluates. We will denote the elements in this sample by $L_{k}^{(i)}:=L(\PredictedLabel_{\star,k}^{(i)},Y_{\star}^{(i)})$ (for notation on RHS see Section~\ref{sec:theory.performance.aggr}). We note that unlike in the case of average based evaluation, there is no running index for the test set data point, only indices $i$ for the data set and $k$ for the prediction strategy.

\begin{table}[h]
  \centering
  \begin{tabular}{|c | c c | c c |}
    \myToprule
    estimate & estimates & f.d.s. & standard error estimate & CLT in\\
    \myMidrule
    $\widehat{\varepsilon}^*_{k}:= \frac{1}{D}\sum_{i=1}^D L_{k}^{(i)}$ & $\varepsilon^*_{k}$ & (c) & $\sqrt{\frac{\widehat{w}_{k}}{D}},\;\mbox{where}\;\widehat{w}_{k}:=\frac{\sum_{i=1}^{D} (L_{k}^{(i)} - \widehat{\varepsilon}_{k}^*)^2}{D-1}$ & $D$ \\
    \myBottomrule
  \end{tabular}
\caption{Table of basic estimates of expected loss, with confidence intervals. First column = definition of the estimate. Second column = the quantity which is estimated by the estimate. Third column = which future data scenario (f.d.s.) estimate and confidence intervals give a guarantee for. Fourth column = standard error estimate (normal approximation) for the estimate in the first column, e.g., to construct frequentist confidence intervals. Fifth column = quantities governing central limit theorem (CLT) asymptotics for the confidence intervals.}
\label{table:perfestaggr}
\end{table}

Table~\ref{table:perfestaggr} presents one estimate of expected loss estimates with proposed standard error estimate, for future data situation (c), i.e., generalization of performance to a new dataset. Even though there is only a single estimate, we present it in a table for concordance with Table~\ref{table:perfest}. An confidence interval at $\alpha$ confidence is obtained as
$$[\widehat{\varepsilon}_{k}^* + \Phi^{-1}(\alpha/2) \widehat{w}_{k}, \widehat{\varepsilon}_{k}^* - \Phi^{-1}(\alpha/2) \widehat{w}_{k}].$$

The mean and variance estimates are obtained from standard theory of mean estimation, by the same principle as $\widehat{\varepsilon}^*_{k}$ for average based estimates.
Estimates for situations (a) may be naively constructed from multiple test sets of the same size, or obtained from further assumptions on $L$ via re-sampling, though we abstain from developing such an estimate as it does not seem to be common - or available - at the state-of-art.

\subsection{Rank based performance quantifiers}
\label{Sec:benchmarking.rank}

mlaut has functionality to compute rankings based on any average or aggregate performance statistic, denoted $L$ below. I.e., for any choice of $L$, the following may be computed.

As in Section~\ref{sec:theory.performance.rank}, define $L_{k}^{(i)}:=L(\PredictedLabel_k^{(i)},Y_{j}^{(i)})$ in the case the raw statistic being an average, and $L_{k}^{(i)}:=L(\PredictedLabel_{\star,k}^{(i)},Y_{\star}^{(i)})$ in case it is an aggregate. Denote by $R_k^{(i)}$ the order rank of $L_{k}^{(i)}$ within the tuple $(L_{1}^{(i)},\dots, L_{K}^{(i)})$.

\begin{table}[h]
  \centering
  \begin{tabular}{|c | c c | c c |}
    \myToprule
    estimate & estimates & f.d.s. & standard error estimate & CLT in\\
    \myMidrule
    $\widehat{r}_{k}= \frac{1}{D}\sum_{i=1}^D R_{k}^{(i)}$ & $r_k$ & (c) & $\sqrt{\frac{\widehat{\nu}_{k}}{D}},\;\mbox{where}\;\widehat{\nu}_{k}:=\frac{\sum_{i=1}^{D} (L_{k}^{(i)} - \widehat{r}_{k})^2}{D-1}$ & $D$ \\
    $z_{k,k'} = \widehat{r}_{k} - \widehat{r}_{k'}$ & $r_k - r_{k'}$ & (c) &
    $\frac{z_{k,k'}\cdot \sqrt{6D}}{\sqrt{K(K+1)}}$ & $K,D$\\
    \myBottomrule
  \end{tabular}
\caption{Table of basic estimates of average rank, with confidence intervals. First column = definition of the estimate. Second column = the quantity which is estimated by the estimate. Third column = which future data scenario (f.d.s.) estimate and confidence intervals give a guarantee for. Fourth column = standard error estimate (normal approximation) for the estimate in the first column, e.g., to construct frequentist confidence intervals. Fifth column = quantities governing central limit theorem (CLT) asymptotics for the confidence intervals.}
\label{table:perfestrank}
\end{table}

Table~\ref{table:perfestrank} presents an average rank estimates and an average rank difference estimate, for future data situation (c), i.e., generalization of performance to a new dataset.\\
The average rank estimate and its standard error is based on the central limit theorem in the number of data sets. The average rank difference estimate is Nem\'enyi's critical difference as referred to in~\cite{demsar_statistical_2006} which is used in visualizations.

\subsection{Statistical tests for method comparison}\label{subsection:statistical_tests}
	
While the methods in previous sections compute performances with confidence bands, they do not by themselves allow to compare methods in the sense of ruling out that differences are due to randomness (with the usual statistical caveat that this can never be ruled out entirely, but the plausibility can be quantified).

mlaut implements significance tests for two classes of comparisons: absolute performance differences, and average rank differences, in future data scenario (c), i.e., with a guarantee for the case where the strategy is re-fitted to a new data source.

mlaut's selection follows closely, and our exposition below follows loosely, the work of~\cite{demsar_statistical_2006}. While the latter is mainly concerned with classifier comparison, there is no restriction-in-principle to leverage the same testing procedures for quantitative comparison with respect to arbitrary (average or aggregate) raw performance quantifiers.

\subsubsection{Performance difference quantification}

The first class of tests we consider quantifies, for a choice of aggregate or average loss $L$, the significance of average differences of expected generalization performances, between two strategies $k$ and $k'$.
The meanings of ``average'' and ``significant'' may differ, and so does the corresponding effect size - these are made precise below.

All the tests we describe are based on the paired differences of performances, where the pairing considered is the pairing through datasets. That is, on dataset $i$, there are performances of strategy $k$ and $k'$ which are considered as a pair of performances.
For the paired differences, we introduce abbreviating notation $\Delta_{k,k'}^{(i)}:= \widehat{\eta}_{i,k} - \widehat{\eta}_{i,k'}$ if the performance is an average loss/score, and $\Delta_{k,k'}^{(i)}:= L^{(i)}_{k}-L^{(i)}_{k'}$ if the loss is an aggregate loss/score. Non-parametric tests below will also consider the ranks of the paired differences, we will write $\Lambda_{k,k'}^{(i)}$ for the rank of $\Delta_{k,k'}^{(i)}$ within the sample $(\Delta_{k,k'}^{(1)},\dots, \Delta_{k,k'}^{(D)})$, i.e., taking values between $1$ and $D$.\\
We denote by $\Delta_{k,k'}^{(*)}$ and $\Lambda_{k,k'}^{(*)}$ the respective population versions, i.e., the performance difference on a random future dataset, as in scenario (c).

\begin{table}[h]
  \centering
  \begin{tabular}{|c | c | c c c |}
    \myToprule
    name & tests null & e.s.(raw) & e.s.(norm) & stat.\\
    \myMidrule
    paired t-test & $\EE [\Delta_{k,k'}^{(*)}] \overset{?}{=} 0$ &
    $\overline{\Delta}_{k,k'} := \frac{1}{D}\sum_{i=1}^D \Delta_{k,k'}^{(i)}$
    & $d_{k,k'}:=\frac{\overline{\Delta}_{k,k'}}{\sqrt{\widehat{v}_{k,k'}}},$ where
    & $t_{k,k'}:= \sqrt{D}\cdot d_{k,k'}$\\
     & & & $\widehat{v}_{k,k'} = \frac{\sum_{i=1}^D (\Delta_{k,k'}^{(i)} - \overline{\Delta}_{k,k'})^2}{D-1}$ & \\
     \myMidrule
     Wilcoxon & $\mbox{med} [\Lambda_{k,k'}^{(*)}] \overset{?}{=} 0$ &
     $w_{k,k'}:= $
     &   & \\
     signed-rank t.& & $\frac{1}{D}\sum_{i=1}^D \Lambda_{k,k'}^{(i)}\sgn (\Delta_{k,k'}^{(i)})$
     & $\rho_{k,k'}:=\frac{2 w_{k,k'}}{D+1} $  & $W_{k,k'}:= D\cdot w_{k,k'}$\\
    \myBottomrule
  \end{tabular}
\caption{Table of pairwise comparison tests for benchmark comparison. name = name of the testing procedure. tests null = the null hypothesis that is tested by the testing procedure. e.s.(raw) = the corresponding effect size, in raw units. e.s.(norm) = the corresponding effect size, normalized. stat. = the test statistic which is used in computation of significance. Symbols are defined as in the previous sections.}
\label{table:testsdiff}
\end{table}

Table~\ref{table:testsdiff} lists a number of common testing procedures. The significances may be seen as guarantees for future data situation (c). The normalized effect size for the paired t-test comparing the performance of strategies $k$ and $k'$, the quantity $d_{k,k'}$ in Table~\ref{table:testsdiff}, is called Cohen's d(-statistic) for paired samples (to avoid confusion in comparison with literature, it should be noted that Cohen's d-statistic also exists for unpaired versions of the t-test which we do not consider here in the context of performance comparison). The normalized effect size for the Wilcoxon signed-rank test, the quantity $\rho_{k,k'}$, is called biserial rank correlation, or rank-biserial correlation.

It should also be noted that the Wilcoxon signed-rank test, while making use of rank differences, is not a pairwise comparison of strategies' performance ranks - this is a common misunderstanding. While ``ranks'' appear in both concepts, the ranks in the Wilcoxon signed-rank tests are the ranks of the performance differences, pooled \emph{across} data sets, while in a rank based performance quantifier, the ranking of different methods' performances (not differences) \emph{within} a data sets (not across data sets) is considered.

The above tests are implemented for one-sided and two-sided alternatives. See~\cite{ross_introductory_2010}, \cite{demsar_statistical_2006}, or~\cite{wilcoxon_individual_1945} for details.

Portmanteau tests for the above may be based on parametric ANOVA, though~\cite{demsar_statistical_2006} %citet
recommends avoiding these due to the empirical asymmetry and non-normality of loss distributions. Hence for multiple comparisons, mlaut implements Bonferroni and Bonferroni-Holm significance correction based post-hoc testing.

	In order to compare the performance of the prediction functions $\PredFunction$ one needs to perform statistical tests on the output produced by $\LossFunction(\PredFunction(\NewX), \NewY)$. Below we enumerate the statistical tests that can be employed to assess the results produced by the loss functions $\LossFunction$ as described in \ref{subsection:loss_functions}.

\subsubsection{Performance rank difference quantification}

Performance rank based testing uses the observed performance ranks $R_k^{(i)}$ of the $k$-th strategy, on the $i$-th data set. These are defined as above in Section~\ref{Sec:benchmarking.rank}, of which we keep notation, including notation for the average rank estimate $\widehat{r}_{k}= \frac{1}{D}\sum_{i=1}^D R_{k}^{(i)}$. We further introduce abbreviating notation for rank differences, $S_{k,k'}^{(i)}:= R_k^{(i)} - R_{k'}^{(i)}$.

\begin{table}[h]
  \centering
  \begin{tabular}{|c | c | c c c |}
    \myToprule
    name & tests null & e.s.(raw) & e.s.(norm) & stat.\\
    \myMidrule
    sign test & $\EE[\sgn (R^{(*)}_k-R^{(*)}_{k'})] \overset{?}{=} 0$ &
    $S_{k,k'} := \sum_{i=1}^D S_{k,k'}^{(i)}$
    & $z_{k,k'}:= \frac{\sqrt{D}\cdot S_{k,k'}}{\sqrt{D^2-S_{k,k'}^2}}$
    & $p_{k,k'}:= \frac{D+S_{k,k'}}{2D}$\\
    \myMidrule
     Friedman & $r_k-r_{k'} \overset{?}{=} 0$ &
     $Q:= $
     &   & \\
     test& (for some $k,k'$) &
     $\frac{12D}{K(K+1)}\sum_{k=1}^K (\widehat{r}_{k} - \frac{K+1}{2})^2$
    & \multicolumn{2}{c|}{$F := \frac{(D-1)Q}{D(K-1) - Q} $}\\
   % & & \\
    \myBottomrule
  \end{tabular}
\caption{Table of pairwise comparison tests for benchmark comparison. name = name of the testing procedure. tests null = the null hypothesis that is tested by the testing procedure. e.s.(raw) = the corresponding effect size, in raw units. e.s.(norm) = the corresponding effect size, normalized. stat. = the test statistic which is used in computation of significance. Symbols are defined as in the previous sections.}
\label{table:testsdiffrank}
\end{table}

Table~\ref{table:testsdiffrank} describes common testing procedures which may both be seen as tests for a guarantee of expected rank difference $r_k - r_{k'}$ in future data scenario (c). The sign test is a binomial test regarding the proportion $p_{k,k'}$ being significantly different from $\frac{1}{2}$. In case of ties, a trinomial test is used.
The implemented version of the Friedman test uses the F-statistic (and not the Q-statistic aka chi-squared-statistic) as described in~\cite{demsar_statistical_2006}.

For post-hoc comparison and visualization of average rank differences, mlaut implements the combination of Bonferroni and studentized rannge multiple testing correction with Nem\'enyi's confidence intervals, as described in~\ref{Sec:benchmarking.rank}.

%% file: inputs/api_design_and_features.tex
\newpage

	\section{MLaut, API Design and Main Features}
	\label{sec:api}
	
MLaut~\cite{mlaut_2018} is a modelling and workflow toolbox that was written with the aim of simplifying the task of running machine learning benchmarking experiments. MLaut was created with the specific use-case of large-scale performance evaluation on a large number of real life datasets, such as the study of~\cite{fernandez-delgado_we_2014}. Another key goal was to provide a scalable and unified high-level interface to the most important machine learning toolboxes, in particular to include deep learning models in such a large-scale comparison.. 	
	
Below, we describe package design and functionality. A short usage handbook is included in Section \ref{subsec:code_examples}
	
MLaut may be obtained from pyPI via \texttt{pip install mlaut}, and is maintained on GitHub at \texttt{github.com/alan-turing-institute/mlaut}. A Docker container can also be obtained from Docker Hub via \texttt{docker pull kazakovv/mlaut}.

	\subsection{Applications and Use}
	
	MLaut main use case is the set-up and execution of supervised (classification and regression) benchmarking experiments. The package currently provides an high-level workflow interface to scikit-learn and keras models, but can easily be extended by the user to incorporate model interfaces from additional toolboxes into the benchmarking workflow.
	
	MLaut automatically creates begin-to-end pipeline for processing data, training machine learning experiments, making predictions and applying statistical quantification methodology to benchmark the performance of the different models.
	
More precisely, MLaut provides functionality to:
	
	\begin{itemize}
		
		\item Automate the entire workflow for large-scale machine learning experiments studies. This includes structuring and transforming the data, selecting the appropriate estimators for the task and data data at hand, tuning the estimators and finally comparing the results.
		
		\item Fit data and make predictions by using the prediction strategies as described in \ref{subsection:estimators} or by implementing new prediction strategies.
		
		\item Evaluate the results of the prediction strategies in a uniform and statistically sound manner.
	\end{itemize}

	\subsection{High-level Design Principles}
	
		We adhered to the high-level API design principles adopted for the scikit-learn project \cite{buitinck_api_2013}. These are:
	
	\begin{enumerate}
		\item Consistency.
		\item Inspection.
		\item Non-proliferation of classes.
		\item Composition.
		\item Sensible defaults.
	\end{enumerate}
	
We were also inspired by the Weka project~\cite{jagtap_census_2013}, a platform widely used for its data mining functionalities. In particular, we wanted to replicate the ease of use of Weka in a pythonic setting.

	\subsection{Design Requirements}

Specific requirements arise from the main use case of scalable benchmarking and the main design principles:
	
	\begin{enumerate}
		\item \textbf{Extensibility.} MLaut needs to provide a uniform and consistent interface to level 3 toolbox interfaces (as in Section~\ref{sec:intro.toolboxes}). It needs to be easily extensible, e.g., by a user wanting to add a new custom strategy to benchmark.
		\item \textbf{Data collection management.} Collections of data sets to benchmark on may be found on the internet or exist on a local computer. MLaut needs to provide abstract functionality for managing such data set collections.
		\item \textbf{Algorithm/model management.} In order to match algorithms with data sets, MLaut needs to have abstract functionality to do so. This needs to include sensible default settings and easy meta-data inspection of standard methodology.
		\item \textbf{Orchestration management.} MLaut needs to conduct the benchmarking experiment in a standardized way with minimal user input beyond its specification, with sensible defaults for the experimental set-up. The orchestration module needs to interact with, but be separate from the data and algorithm interface.
		\item \textbf{User Friendliness.} The package needs to be written in a pythonic way and should not have a steep learning curve. Experiments need to be easy to set-up, conduct, and summarize, from a python console or a jupyter notebook.
	\end{enumerate}
	
In our implementation of MLaut, we attempt to address the above requirements by creating a package which:
	
	\begin{itemize}
		\item	\textit{Has a nice and intuitive scripting interface}. One of our main requirements was to have a native Python scripting interface that integrates well with the rest of our code. Our design attempts to reduce user interaction to the minimally necessary interface points of experiment specification, running of experiments, and querying of results.
		
		\item \textit{Provides a high level of abstraction form underlying toolboxes}. Our second criteria was that MLaut provided high level of abstraction from underlying toolboxes. One of our main requirements was for MLaut to be completely model and toolbox agnostic. The scikit-learn interface was too light-weight for our purposes as its parameter and meta-data management is not interface explicit (or inspectable).
		
		\item \textit{Provides Scalable workflow automation}. This needed to be one of MLaut's cornerstone contributions. Its main logic is implemented in the \code{orchestrator} class that orchestrates the evaluation of all estimators on all datasets. The class manages resources for building the estimator models, saving/loading the data and the estimator models. It is also aware of the experiment's partial run state and can be used for easy resuming of an interrupted experiment.
		
		\item \textit{Allows for easy estimator construction and retrieval}. The end user of the package should be able to easily add new machine learning models to the suite of build in ones in order to expand its functionality. Besides a small number of required methods to implement, we have provided interfaces to two of the most used level 3 toolbox packages, sklearn and keras.
		
		\item \textit{Has a dedicated meta-data interface for sensible defaults of estimators}. We wanted to ensure that the estimators that are packaged in MLaut come with sensible defaults, i.e. pre-defined hyper-parameters and tuning strategies that should be applicable in most use cases. The robustness of these defaults has been tested and proven as part of the original large-scale classification study. As such, the user is not required to have a detailed understanding of the algorithms and how they need to be set up, in order to make full use them.
		
		\item \textit{Provides a framework for quantitative benchmark reporting}. Easily accessible evaluation methodology for the benchmarking experiments is one of the key features of the package. We also considered reproducibility of results as vital, reflected in a standardized set-up and interface for the experiments, as well as control throughout of pseudo-random seeds..
		
		\item \textit{Orchestrates the experiments and parallelizes the load over all available CPU cores}.  A large benchmarking study can be quite computationally expensive. Therefore, we needed to make sure that all available machine resources are fully utilized in the process of training the estimators. In order to achieve this we used the parallelization methods that are available as part of the \code{GridSearch} method and natively with some of the estimators. Furthermore, we also provide a Docker container for running MLaut which we recommend using as a default as it allows the package to run in the background at full load.
						
		\item \textit{Provides a uniform way of storing a retrieving data}. Results of benchmarking experiments needed to be saved in a uniform way and made available to users and reviewers of the code. At the current stage, we implemented back-end functionality for management via local HDF5 database files. In the future, we hope to support further data storage back-ends with the same orchestrator-sided facade interface.
	\end{itemize}

		\subsubsection{Estimator encapsulation}
		
		MLaut implements a logic of encapsulating the meta-data with the estimators that it pertains to. This is achieved by using a decorator class that is attached to each estimator class. By doing this, our extended interface is are able to bundle wide-ranging meta-data information with each estimator class. This includes:
		
		\begin{itemize}
			\item Basic estimator properties such as name, estimator family;
			\item Types of tasks that a particular estimator can be applied to;
			\item The type of data which the estimator expects or can handle;
			\item The model architecture (on level 3, as in Section~\ref{sec:intro.toolboxes}). This is particularly useful for more complex estimators such as deep neural networks. By applying the decorator structure the model architecture can be easily altered without changing the underlying estimator class.
		\end{itemize}
		
This extended design choice has significant benefits for a benchmarking workflow package. First of all, it allows fsearching for estimators based on some basic criteria such as task or estimator family. Second of all, it allows to inspect, query, and change default hyper-parameter settings used by the estimators. Thirdly, strategies with different internal model architectures can be deployed with relative ease.
		
	\subsubsection{Workflow design}
	
The workflow supported by MLaut consists of the following main steps:
	
	\begin{figure}[h!]
		\centering
		\includegraphics[scale=0.7]{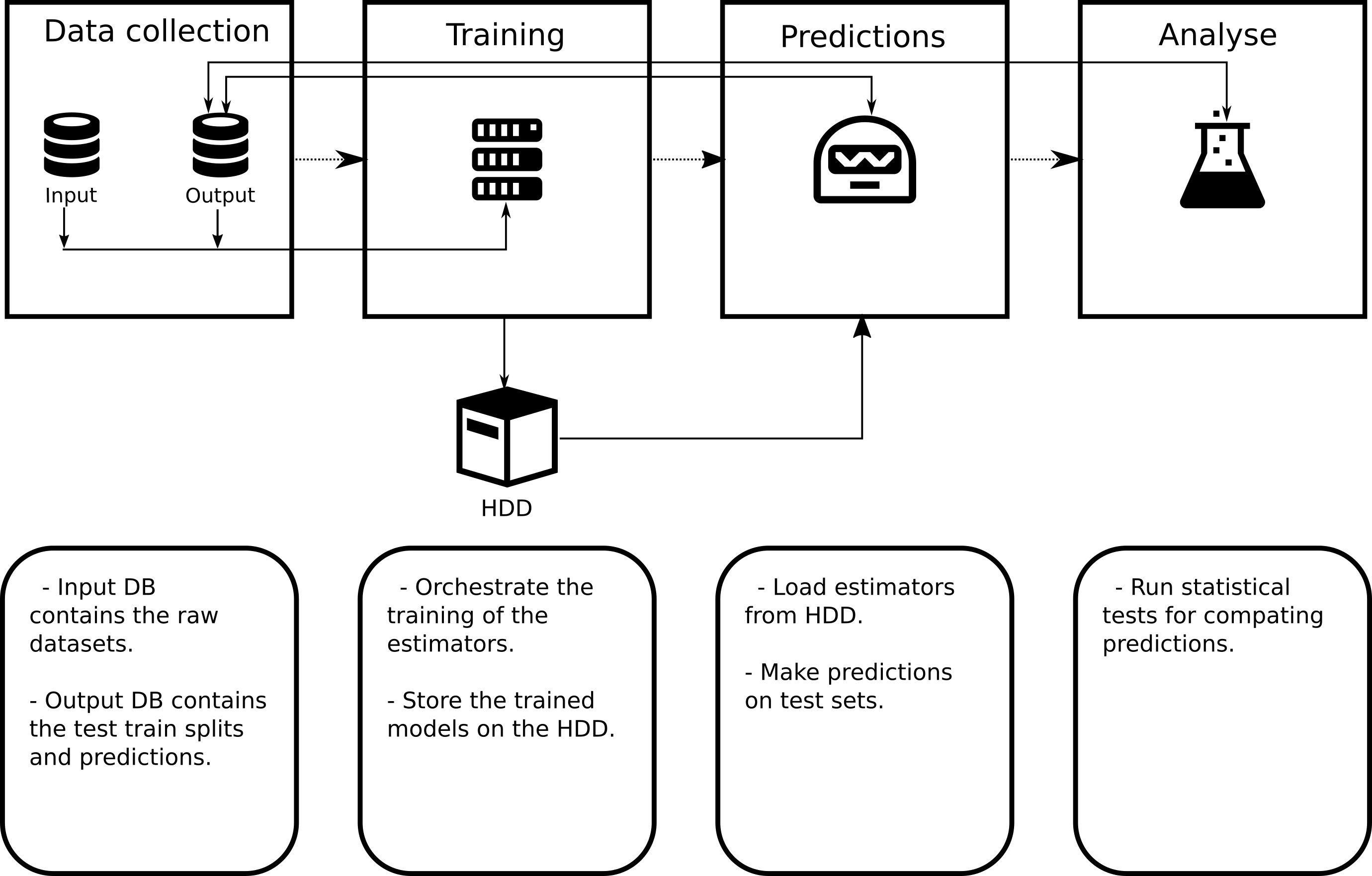}
		\caption{The orchestrated MLaut workflow}
	\end{figure}
	\begin{enumerate}

		\item \textbf{Data collection.} As a starting point the user needs to gather and organize the datasets of interest on which the experiments will be run. The raw datasets need to be saved in a HDF5 database. Metadata needs to be attached to each dataset which is later used in the training phase for example for distinguishing the target variables. MLaut provides an interface for manipulating the databases through its \code{Data} and \code{Files\_IO} classes. The logic of the toolbox is to provision two HDF5 databases one for storing the input data such as  the datasets and a second one to store the output of the machine learning experiments and processed data such as train/test index splits. This separation of input and output is not required but is recommended. The datasets also need to be split in a \textit{train} and \textit{test} set in advance of proceeding with the next phase in the pipeline. The indices of the train and test splits are stored separately from the actual datasets in the HDF5 database to ensure data integrity and reproducibility of the experiments. All estimators are trained and tuned on the training set only. At the end of this process the estimators are used on the test sets which guarantees that all predictions are made on unseen data.
		
		\item \textbf{Training phase.} After the datasets are stored in the HDF5 database by following the convention adopted by MLaut the user can proceed to training the estimators. The user needs to provide an array of machine learning estimators that will be used in the training process. MLaut provides a number of default estimators that can be instantiated. This can be done by the use of the \code{estimators} module. The package also provides the flexibility for the user to write its own estimator by inheriting from the \code{mlaut\_estimator} class. Furthermore, there is a \code{generic\_estimator} module which provides flexibility for the user to create new estimators with only a couple of lines of code.
		
		The task of training the experiments is performed by the \code{experiments. Orchestrator} class. This class manages the sequence of the training the the parallelization of the load. Before training each dataset is preprocessed according to metadata provided on the estimator level. This includes normalizing the features and target variables, conversion from categorical to numerical values.
		
		We recommend running the experiments inside a Docker container if they are very computationally intensive. This allows MLaut to run in the background on a server without shutting down unexpectedly due to loss of connection. We have provided a Docker image that makes this process easy.
		
		\item \textbf{Making predictions.} During training the fitted models are stored on the hard drive. At the end of the training phase the user can again use \code{experiments.Orchestrator} class to retrieve the trained models and make predictions on the test sets.
		
		\item \textbf{Analyse results.} \label{pipelineAnalyseResults} The last stage is analysing the output of the results of the machine learning experiments. In order to initiate the process the user needs to call the \code{analyze\_results.prediction\_errors} method which returns two dictionaries with the average errors per estimator on all datasets as well as the errors per estimator achieved on each dataset. These results can be used as inputs to the statistical tests that are also provided as part of the \code{analyze\_rezults} module which mostly follow the methodology proposed by \cite{demsar_statistical_2006}.
		
	\end{enumerate}
	
	\subsection{Software Interface and Main Toolbox Modules}	
	
	MLaut is built around the logic of the pipeline workflow described earlier. Our aim was to implement the programming logic for each step of the pipeline in a different module. The code that is logically used in more than one of the stages is implemented in a \code{Shared} module that is accessible by all other classes. The current design pattern is most closely represented by the \textit{fa\c{c}ade} and \textit{adaptor} patterns under which the user interacts with one common interface to access the underlying adaptors which represent the underlying machine learning and statistical toolboxes.
	
	\begin{figure}[h!]
			\centering
			\includegraphics[scale=0.7]{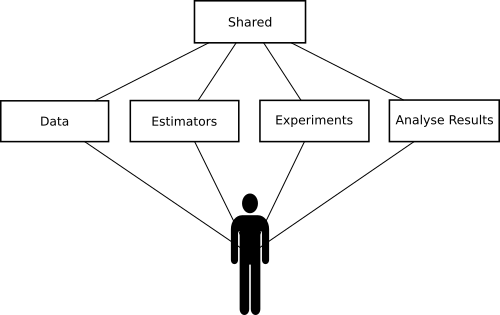}
			\caption{Interaction between the main MLaut modules}
	\end{figure}

		\subsubsection{Data Module}
		
		The \code{Data} module contains the high level methods for manipulating the raw datasets. It provides a second layer of interface to the lower level classes for accessing, storing and extracting data from HDF5 databases. This module uses heavily the functionality developed in the \code{Shared} module but provides a higher level of abstraction for the user.
		
		\subsubsection{Estimators Module} \label{subsection:estimators_module}
		
		This module encompasses all machine learning models that come with MLaut as well as methods for instantiating them based on criteria provided by the user. We created MLaut for the purpose of running supervised classification experiments but the toolbox also comes with estimators that can be used for supervised regression tasks.
		
		From a software design perspective the most notable method in this class is the \code{build} method which returns an instantiated estimator with the the appropriate hyper parameter search space and model architecture. In software design terms this approach resembles more closely the \textit{builder design pattern} which aims at separating the construction of and object from its representation. This design choice allows the base \code{mlaut\_estimator} class to create different representations of machine learning models.
		
		The \code{mlaut\_estimator} object includes methods that complete its set of functionalities. Some of the main ones are a \code{save} method that takes into account the most appropriate format to persist a trained estimator object. This could include the pickle format used by most \textit{scikit-learn} estimators or the \textit{HDF5} format used by keras. A \code{load} function is also available for restoring the saved estimators.
		
		The design of the package also relies on the estimators having a uniform \code{fit} and \code{predict} methods that takes the same input date and generate predictions in the same format. These methods are not implemented at the \code{mlaut\_estimator} level but instead we relied on the fact that these fundamental methods will be uniform across the underlying packages. However, there is a discrepancy in the behaviour of the \textit{scikit-learn} and \textit{keras} estimators. For classification tasks \textit{keras} requires the labels of the training data to be one hot encoded. Furthermore, the default behaviour of the keras \code{predict} method is equivalent to the \code{predict\_proba} in scikit-learn. We solved these discrepancies by overriding the \code{fit} and \code{predict} methods of the implemented keras estimators.
		
		Through the use of decorators and by implementing the \code{build} method we are able to fully customize the estimator object with minimal required programming. The decorator class allows to set the metadata associated with the estimator. This includes setting the name, estimator family, types of tasks and hyper parameters. This together with an implemented \code{build} method will give the user a fully specified machine learning  model. This approach also facilitates the application of the algorithms and the use of the software as we can ensure that each algorithm is matched to the correct datasets. Furthermore, this allows to easily retrieve the required algorithms by executing a simple command.
		
		\begin{figure}[h!]
			\centering
			\includegraphics[scale=0.7]{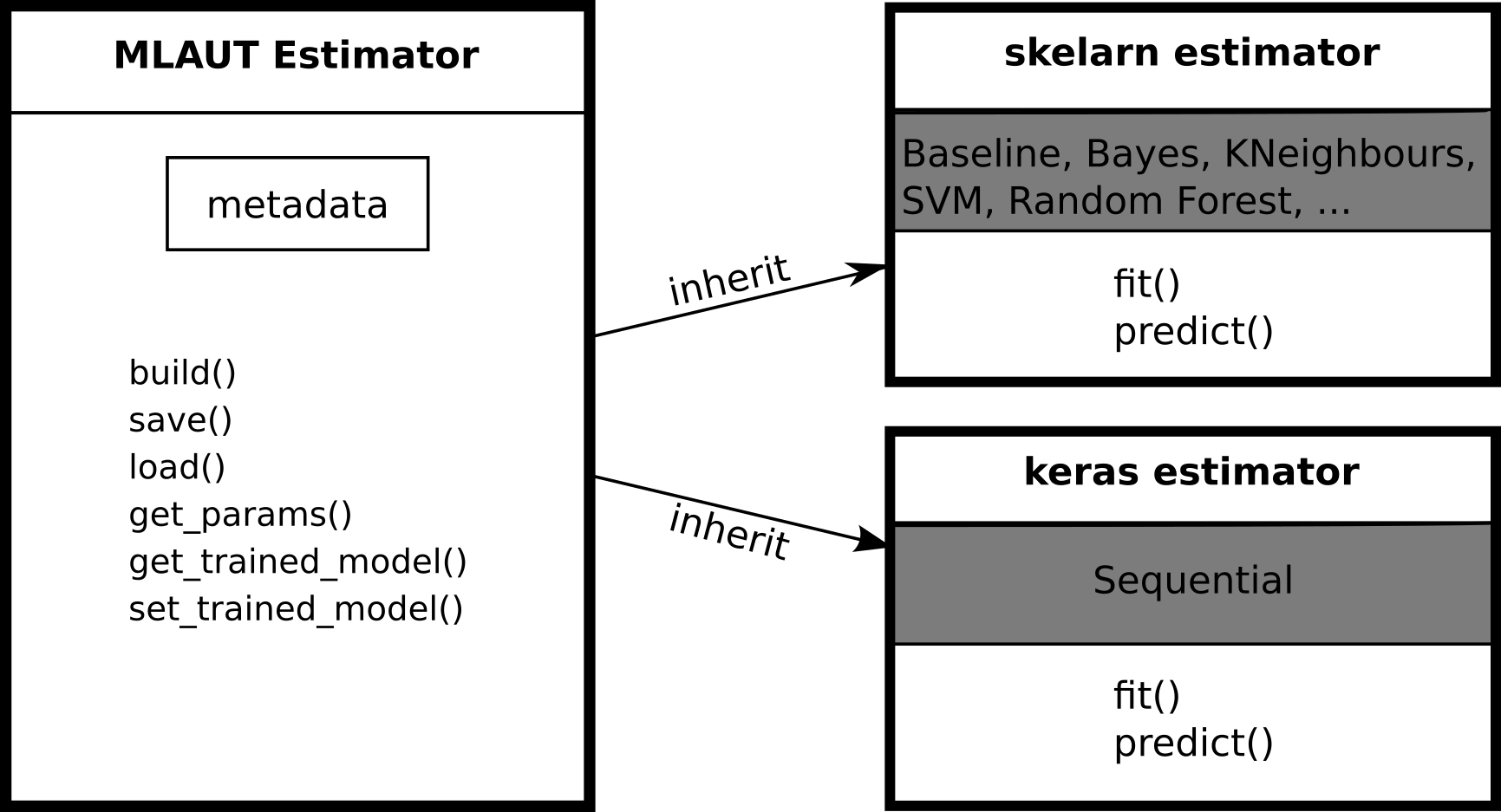}
			\caption{Interaction between MLaut Estimator class and third-party ml estimators}
		\end{figure}
	
Closely following terminology and taxonomy of~\cite{james_introduction_2013}, mlaut estimators are currently assigned to one of the following methodological families:
		\begin{enumerate}[label=\alph*)]
			\item \textbf{Baseline Estimators}. This family of models is also referred to as a dummy estimator and serves as a benchmark to compare other models to. It does not aim to learn any representation of the data but simply adopts a strategy of guessing.
			\item \textbf{Generalized Linear Model Estimators}. A family of models that assumes that a (generalized) linear relationship exists between the dependent and target values.
			\item \textbf{Naive Bayes Estimators}. This class of models applies the Bayes theorem my making the naive assumption that all features are independent.
			\item \textbf{Prototype Method Estimators}. Family of models that apply prototype matching techniques for fitting the data. The most prominent member of this family is the K-means algorithm.
			\item \textbf{Kernel Method Estimators}. Family of models using kernelization techniques, including support vector machine based estimazors.
			\item \textbf{Deep Learning and Neural Network Estimators}. This family of models provides implementation of neural network models, including deep neural networks.
			\item \textbf{Ensembles-of-Trees Estimators}. Family of methods that combines the predictions of several tree-based estimators in order to produce a more robust overall estimator. This family is further divided in:
				\begin{enumerate}
					\item \textit{averaging methods}. The models in this group average the predictions of several independent models in order to arrive at a combined estimator. An example is Breiman's random forest.
					\item \textit{boosting methods}. An ensembling approach of building models sequentially based on iterative weighted residual fitting. An example are stochastic gradient boosted tree models.
				\end{enumerate}
		\end{enumerate}
	
		In addition ot this the user also has the option to write their own estimator objects. In order to achieve this the new class needs to inherit from the \code{mlaut\_estimator} class. and implementing the abstract methods in each child class. The main abstract method that needs to be implemented is the \code{build} method which returns an wrapped instance of the estimator with a set of hyper-parameters that will be used in the tuning process. For further details about the implemented estimators refer to \ref{subsection:estimators}.
		
		\subsubsection{Experiments Module}
		
		This module contains the logic for orchestration of the machine learning experiments. The main parameters in this module are the  datasets and the estimator models that will be trained on the data. The main \code{run} method of the module then proceeds to training all estimators on all datasets, sequentially. The core of the method represent two embedded \code{for} loops the first of which iterates over the datasets and the second one over the estimators. Inside the inner loop the orchestrator class builds an estimator instance for each dataset. This allows to tailor the machine learning model for each dataset. For example, the architecture of a deep neural network can be altered to include the appropriate number of neurons based on the input dimensions of the data. This module is also responsible for saving the trained estimators and making predictions. It should be noted that the orchestrator module is not responsible for the parallelization of the experiments which is handled on an individual estimator level.
%
%
%		\begin{algorithm}[H]
%			\KwData{datasets, estimator instances}
%			\ForEach{dataset in datasets}{
%				\ForEach{estimator in estimators}{
%					estimator.build(num\_classes, input\_dim, num\_samples) \\
%					estimator.fit() \\
%					estimator.save()
%				}
%			}
%		\caption{Pseudo-code for MLaut experiments as carried out by orchestrator}
%		\end{algorithm}

\subsubsection{Result Analysis Module} \label{subsection:analyse_results}
		
This module includes the logic for performing the quantitative evaluation and comparison of the machine learning strategies' performance. The predictions of the trained estimators on the test sets for each dataset serve as input. First, performances and, if applicable, standard errors on the individual data sets are computed, for a given average or aggregate loss/performance quantifier. The samples of performances are then used as inputs for comparative quantification.
		
API-wise, the framework for assessing the performance of the machine learning estimators hinges on three main classes. The \code{anlyze\_results} class implements the calculation of the quantifiers. Through composition this class relies on the \code{losses} class that performs the actual calculation of the prediction performances over the individual test sets. The third main class that completes the framework design is the \code{scores} class. It defines the loss/quantifier function that is used for assessing the predictive power of the estimators. An instance of the \code{scores} class is passed as an argument to the \code{losses} class.
		
We believe that this design choice of using three classes is required to provide the necessary flexibility for the composite performance quantifiers as described in Section~\ref{Sec:benchmarking} - i.e., to allow to compute ranks for an arbitrarily chosen loss (e.g., mean rank with resect to mean absolute error), or to perform comparison testing using an arbitrarily chosen performance quantifier (e.g., Wilcoxon signed rank test comparing F1-scores). 

Our API also facilitates user custom extension, e.g., for users who wish to add a new score function, an efficient way to compute aggregate scores or standard errors, or a new comparison testing methodology.
For example, adding new score functions can be easily achieved by inheriting from the \code{MLautScore} abstract base class. On the other hand, the \code{losses} class completely encapsulates the logic for the calculation of the predictive performance of the estimators. This is particularly useful as the class internally implements a mini orchestrator procedure for calculating and presenting the loss achieved by all estimators supplied as inputs. Lastly, the suite of statistical tests available in MLaut can be easily expanded by adding the appropriate method to the \code{analyze\_results} class or a descendant.
		
		\begin{figure}[h!]
			\centering
			\includegraphics[scale=0.7]{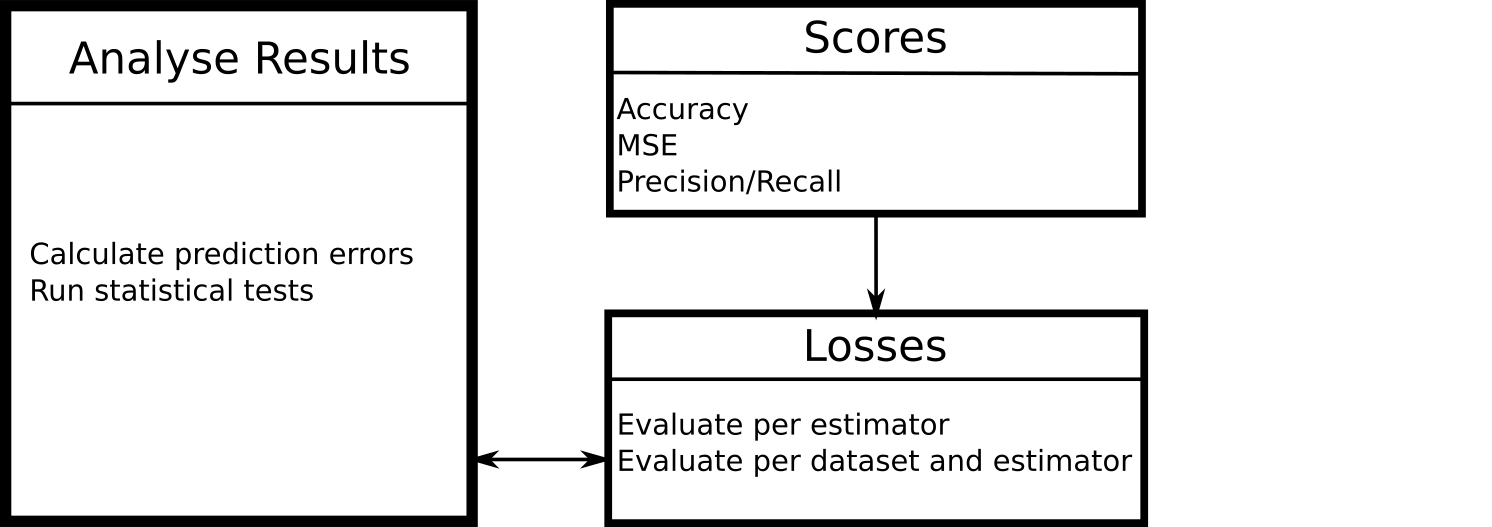}
			\caption{Interaction between main classes inside the \code{analyse\_results} module}
		\end{figure}
	
Mathematical details of the implemented quantification procedures implemented in MLaut were presented in Section~\ref{Sec:benchmarking}. Usage details
		
		In this implementation of MLaut we use third-party packages for performing the statistical tests. We rely mostly on the \code{scikit-learn} package. However, for post hoc tests we use the \code{scikit-posthocs} package \cite{terpilowski_scikit-posthocs:_nodate} and the \code{Orange} package \cite{demsar_orange:_2013} which we also used for creating critical distance graphs for comparing multiple classifiers.

\subsubsection{Shared Module}
		
This module includes classes and methods that are shared by the other modules in the package. The \code{Files\_IO class} comprises of all methods for manipulating files and datasets. This includes saving/loading of trained estimators from the HDD and manipulating the HDF5 databases. The \code{Shared} module also keeps all static variables that are used throughout the package.
	
\subsection{Workflow Example}
\label{subsec:code_examples}

We give a step-by-step overview over the most basic variant of the user workflow. Advanced examples with custom estimators and set-ups may be found in the MLaut tutorial~\cite{mlaut_2018}.

\subsubsection*{Step 0: setting up the data set collection}
The user should begin by setting up the data set collection via the \code{Files\_IO class}. Meta-data for each dataset needs to be provided that includes as a minimum the class column name/target attribute and name of dataset. This needs to be done {\emph once} for every dataset collection, and may not need to be done for a pre-existing or pre-deployed collection. Currently, only local HDF5 data bases are supported.

We have implemented back-end set-up routines which download specific data set collections and generate the meta-data automatically. Current support includes the UCI library data sets and OpenML.
Alternatively, the back-end may be populated directly by storing an in-memory pandas DataFrame via the \code{save\_pandas\_dataset} method, e.g., as part of custom loading scripts.
	
	\begin{lstlisting}[language=Python]
input_io.save_pandas_dataset(
	dataset=data, #in Pandas DataFrame format
	save_loc='/openml', #location in HDF5 database	
	metadata=metadata)
\end{lstlisting}
	
In this case, meta-data for the individual datasets needs to be provided in the following dictionary format:
	
	\begin{lstlisting}[language=Python]
metadata = {
	'class_name': ..., #label column name
	'source': , # source of the data
	'dataset_name': ... ,
	'dataset_id': id
	}
	\end{lstlisting}

\subsubsection*{Step 1: initializing data and output locations}
As the next step, the user should specify the back-end links to the data set collections (``input'') and to intermediate or analysis results (``output'').This is done via the \code{data} class. It is helpful for code readability to store these in code{input\_io} and \code{out\_io} variables.
	
	\begin{lstlisting}[language=Python]
input_io = data.open_hdf5('data/openml.h5', mode='r')
out_io = data.open_hdf5(
	'data/openml-classification.h5', mode='a')
	\end{lstlisting}
	
These may then be supplied as parameters to preparation and orchestration routines.
We then proceed to getting the paths to the raw datasets as well as the respective train/test splits  which is performed respectively though the use of \code{list\_datasets} and \code{split\_datasets} methods.
	
	\begin{lstlisting}[language=Python]
(dts_names_list,
dts_names_list_full_path) = data.list_datasets(
	hdf5_io=input_io,
	hdf5_group='openml/')
										
split_dts_list = data.split_datasets(hdf5_in=input_io,
	hdf5_out=out_io,
	dataset_paths=dts_names_list_full_path)
	\end{lstlisting}
	
\subsubsection*{Step 2: initializing estimators}
The next step is to instantiate the learning strategies, estimators in sklearn terminology, which we want to use in the benchmarking exercise. The most basic and fully automated variant is use of the \code{instantiate\_default\_estimators} method which loads a pre-defined set of defaults given specified criteria. Currently, only a simple string look-up via the \code{estimators} parameter is implemented, but we plan to extend the search/matching functionality. The string criterion may be used to fetch specific estimators by a list of names, entire families of models, estimators by task (e.g., classification), or simply all available estimators.
	
	\begin{lstlisting}[language=Python]
instantiated_models = instantiate_default_estimators(
	estimators=['Classification'],
	verbose=1,
	n_jobs=-1)
	\end{lstlisting}
	
\subsubsection*{\bf Step 3: orchestrating the experiment}
	
The final step is to run the experiment by passing references to data and estimators to the \code{orchestrator} class, then initating the training process by invoking its \code{run} method.
		
	\begin{lstlisting}[language=Python]
orchest = orchestrator(hdf5_input_io=input_io,
	hdf5_output_io=out_io,
	dts_names=dts_names_list,
	original_datasets_group_h5_path='openml/')
			
orchest.run(modelling_strategies=instantiated_models)
	\end{lstlisting}

%	At the end of the training phase we need to make predictions on the test sets. This is again done through the orchestrator class. The user needs to provide a list of MLaut estimator objects and the location where the trained estimators were saved.
%	
%	\begin{lstlisting}[language=Python]
%instantiated_models = instantiate_default_estimators(
%	estimators=['NeuralNetworkDeepClassifier'],
%	verbose=1)
%	
%orchest.predict_all(trained_models_dir='data/trained_models',
%	estimators=instantiated_models)
%	\end{lstlisting}
%	
\subsubsection*{Step 4: computing benchmark quantifiers}
After the estimators are trained and the predictions of the estimators are recorded we can proceed to obtaining quantitative benchmark results for the experiments.
	
For this, we need to instantiate the \code{AnalyseResults} class by supplying the folders where the raw datasets and predictions are stored. Its \code{prediction\_errors} method may be invoked to returns both the calculated prediction performance quantifiers, per estimator as well as the prediction performances per estimator and per dataset.
		
	\begin{lstlisting}[language=Python]
analyse = AnalyseResults(
	hdf5_output_io=out_io,
	hdf5_input_io=input_io,
	input_h5_original_datasets_group='openml/',
	output_h5_predictions_group='experiments/predictions/')

#Score function that will be used for the statistical tests
score_accuracy = ScoreAccuracy()

estimators = instantiate_default_estimators(['Classification'])

(errors_per_estimator,
errors_per_dataset_per_estimator,
errors_per_dataset_per_estimator_df) =
	analyse.prediction_errors(metric=score_accuracy, estimators=estimators)
	\end{lstlisting}

	The prediction errors per dataset per estimator can be directly examined by the user. On the other hand, the estimator performances may be used as further inputs for comparative quantification via hypothesis tests. For example, we can perform a paired t-test for pairwise comparison of methods by invoking the code below:
	
	\begin{lstlisting}[language=Python]
t_test, t_test_df = analyze.t_test(errors_per_estimator)
	\end{lstlisting}
	

%% file: inputs/ml_comparison_study.tex
% -----------------------------------------------------------------------------------------------------
% LOCAL MACROS
%table used for formatting the descriptions of estimators implemented by the package.
\newcommand{\desestim}[3]{
	\begin{tabular}{p{\dimexpr 0.25\linewidth-2\tabcolsep} p{\dimexpr 0.75\linewidth-2\tabcolsep} }

		\textbf{Estimator name} 	& \underline{\textbf{\code{#1}}} \\
		\textbf{Description} 		& #3 \\
		\textbf{Hyperparameters} 	& #2
	\end{tabular}
}

\newcommand{\linkA}{
	\href{http://scikit-learn.org/stable/auto_examples/svm/plot_rbf_parameters.html}
	{http://scikit-learn.org/stable/auto\_examples/svm/plot\_rbf\_parameters.html}
}
\newcommand{\linkB}{ 		
	\href{http://scikit-learn.org/stable/auto_examples/model_selection/plot_grid_search_digits.html}
	{http://scikit-learn.org/stable/auto\_examples/model\_selection/plot\_grid\_search\_digits.html}
}

\newcommand{\linkC}{
	\href{https://scikit-learn.org/stable/auto_examples/model_selection/plot_grid_search_digits.html\#sphx-glr-auto-examples-model-selection-plot-grid-search-digits-py}
	{https://scikit-learn.org/stable/auto\_examples/model\_selection/plot\_grid\_search\_digits.html\#sphx-glr-auto-examples-model-selection-plot-grid-search-digits-py}
}

% END OF LOCAL MACROS
% -----------------------------------------------------------------------------------------------------

\section{Using MLaut to Compare the Performance of Classification Algorithms}

As an major test use case for MLaut, we conducted a large-scale benchmark experiment comparing a selection of off-shelf classifiers on datasets from the UCI Machine Learning Repository. Our study had four main aims:
\begin{enumerate}
\itemsep-0.2em
\item[(1)] stress testing the MLaut framework on scale, and observing the user interaction workflow in a major test case.
\item[(2)] replicating the key points of the experimental set-up by~\cite{fernandez-delgado_we_2014}, while avoiding their severe mistake of tuning on the test set.
\item[(3)] including deep learning methodology to the experiment.
\end{enumerate}

Given the above, the below benchmarking study is, to the best of our knowledge, the first large-scale supervised classification study which\footnote{in the disjunctive sense: i.e., to the best of our knowledge, the first large-scale benchmarking study which does \emph{any} of the above rather than being only the first study to do \emph{all} of the above.}:
\begin{enumerate}
\itemsep-0.2em
\item[(a)] is correctly conducted via out-of-sample evaluation and comparison. This is since~\cite{fernandez-delgado_we_2014} commit the mistake of tuning on the test set, as it is even acknowledged in their own Section 3 Results and Discussion.
\item[(b)] includes contemporary deep neural network classification approaches, and is conducted on a broad selection of classification data sets which is not specific to a special domain such as image classification (the UCI dataset collection).
\end{enumerate}

We intend to extend the experiment in the future by including further dataset collections and learning strategies.

Full code for our experiments, including random seeds, can be found as a jupyter notebook in MLaut's documentation~\cite{mlaut_2018}.

\subsection{Hardware and software set-up}

The benchmark experiment was conducted on a Microsoft Azure VM with 16 CPU cores and 32 GB of RAM, by our Docker virtualized implementation of MLaut. The experiments ran for about 8 days. MLaut requires Python 3.6 and should be installed in a dedicated virtual environment in order to avoid conflicts or the Docker implementation should be used. The full code for running the experiments and the code for generating the results in results Appendix~\ref{section:statistical_appendix} can be found in the examples directory in the \href{https://github.com/alan-turing-institute/mlaut/blob/master/examples/mlaut_study.zip}{GitHub repository} of the project.
		
\subsection{Experimental set-up}
\label{subsection:expsetup}

\subsubsection{Data set collection}
	
The benchmarking study uses the same dataset collection as employed by~\cite{fernandez-delgado_we_2014}. This collection consists of 121 tabular datasets for supervised classification, taken directly from the UCI machine learning repository. Prior to the experiment, each dataset was standardized, such that each individual feature variable has a mean of 0 and a standard deviation of 1. 

The dataset collection of~\cite{fernandez-delgado_we_2014} intends to be representative of a wide scope of basic real-world classification problems. It should be noted that this representative cross-section of simple classification tasks \emph{excludes} more specialized tasks such as image, audio, or text/document classification which are usually regarded to by typical applications of deep learning, and for which deep learning is also the contemporary state-of-art.
For a detail description of the~\citet{fernandez-delgado_we_2014} data collection, see section~2.1 there.

%	In order to conduct our experiments we used [530] datasets that were downloaded from the OpenML website. We used OpenML's Python interface to filter for datasets that meet the following criteria:

\subsubsection{Re-sampling for evaluation}

Each dataset is in split into exactly one pair of training and test set. The training sets are selected, for each data set, uniformly at random\footnote{independently for each dataset in the collection} as (a rounded) $\frac{2}{3}$ of the available data sample; the remaining $\frac{1}{3}$ in the dataset form the test set on which the strategies are asked to make predictions. Random seeds and the indices of the exact splits were saved to ensure reproducibility and post-hoc scrutiny of the experiments.

The training set may (or may not) be further split by the contender methods for tuning - as stated previously in Section~\ref{sec:theory}, this is not enforced as part of the experimental set-up\footnote{Unlike in the set-up of~\cite{fernandez-delgado_we_2014} which, on top of doing so, is also faulty.}, but is left to each learning strategy to deal with internally, and will be discussed in the next section. In particular, none of the strategies have access to the test set for tuning or training.

\subsection{Evaluation and comparison}
We largely followed the procedure suggested by \cite{demsar_statistical_2006} for the analysis of the performance of the trained estimators.
For all classification strategies, the following performance quantifiers are computed per dataset:
\begin{enumerate}
\itemsep-0.2em
\item misclassification loss
\item rank of misclassification loss
\item runtime
\end{enumerate}
Averages of these are computed, with standard errors for future data situation (c: re-trained, on unseen dataset). In addition, for the misclassification loss on each data set, standard errors for future data situation (a: re-used, same dataset) are computed.

The following pairwise comparisons between samples of performances by dataset are computed:
\begin{enumerate}
\itemsep-0.2em
\item paired t-test on misclassification losses, with Bonferroni correction
\item (paired) Wilcoxon signed rank on misclassification losses, with Bonferroni correction
\item Friedman test on ranks, with Nem\'enyi's significant rank differences and post-hoc significances
\end{enumerate}

Detail descriptions of these may be found in Section~\ref{Sec:benchmarking}.

\subsection{Benchmarked machine learning strategies}
\label{subsection:estimators}
Our choice of classification strategies is not exhaustive, but is meant to be representative of off-shelf choices in the \textit{scikit-learn} and \textit{keras} packages. We intend to extend the selection in future iterations of this study.

From \textit{scikit-learn}, the suite of standard off-shelf approaches includes linear models, Naive Bayes, SVM, ensemble methods, and prototype methods.

We used \textit{keras} to construct a number of neural network architectures representative of the state-of-art. This proved a challenging task due to the lack of explicitly recommended architectures for simple supervised classification to be found in literature.

\subsubsection{Tuning of estimators}

It is important to note that the off-shelf choices and their default parameter settings are often not considered good or state-of-art: hyper-parameters in \textit{scikit-learn} are by default not tuned, and there are no default \textit{keras}  that come with the package.

For \textit{scikit-learn} classifiers, we tune parameters using \textit{scikit-learn}'s \texttt{GridSearchCV} wrapper-compositor (which never looks at the test set by construction).

In all cases of tuned methods, parameter selection in the inner tuning loop is done via grid tuning by 5-fold cross-validation, with respect to the default \textit{score} function implemented at the estimator level. For classifiers as in our study, the default tuning score is mean accuracy (averaged over all 5 tuning test folds in the inner cross-validation tuning loop), which is equivalent to tuning by mean misclassification loss.

The tuning grids will be specified in Section~\ref{subsection:estimators.sklearn} below.

For \textit{keras} classifiers, we built architectures by interpolating general best practice recommendations in scientific literature~\cite{ojha_metaheuristic_2017}, as well as based on concrete designs found in software documentation or unpublished case studies circulating on the web. We further followed the sensible default choices of \textit{keras} whenever possible.

The specific choices for neural network architecture and hyper-parameters are specified in Section~\ref{subsection:estimators.keras} below.
	
\subsubsection{Off-shelf scikit-learn supervised strategies}

\label{subsection:estimators.sklearn}

	\begin{enumerate}[label=\roman*)]
		
		\item Algorithms that do not have any tunable hyperparameters
		
			\desestim{sklearn.dummy.DummyClassifier}
			{None}
			{This classifier is a naive/uninformed baseline and always predicts the most frequent class in the training set (``majority class''). This corresponds to the choice of the \textit{most\_frequent} parameter.}
			
			\desestim{sklearn.naive\_bayes.BernoulliNB}{None}{Naive Bayes classifier for multivariate Bernoulli models. This classifier assumes that all features are binary, if not they are converted to binary. For reference please see \cite{bishop_pattern_2006} Chapter 2.}
			
			\desestim{sklearn.naive\_bayes.GaussianNB}
			{None}
			{Standard implementation of the Naive Bayes algorithm with the assumption that the features are Gaussian. For reference please see \cite{bishop_pattern_2006} Chapter 2. }

		\item Linear models
			
			\desestim{sklearn.linear\_model.PassiveAggressiveClassifier}		
			{\code{C}: array of 13 equally spaced numbers on a log scale in the range  $[10^{-2};10^{10}]$ \code{scikit-learn} default: 1}
			{Part of the online learning family of models based on the \textit{hinge loss} function. This algorithm observes feature-value pairs $z$ in sequential manner. After each observation the algorithm makes a prediction, checks the correct value and calibrates the weights. For further reference see \cite{crammer_online_2006}.}
				
		\item Clustering Algorithms
		
				\desestim{sklearn.neighbors.KNeighborsClassifier}
				{\code{n\_neighbors=[1;30]}, \code{scikit-learn} default: 5
				
				 \code{p=[1,2], \code{scikit-learn} default:2}
				}
				{The algorithm uses a majority vote of the nearest neighbours of each data point to make a classification decision. For reference see \cite{cover_nearest_1967} and \cite{bishop_pattern_2006}, Chapter 2.}
		
		\item Kernel Methods
			
				\desestim{sklearn.svm.SVC}{
				C: array of 13 equally spaced numbers on a log scale in the range $[2^{-5};2^{15}]$, \code{scikit-learn} default: 1. %$[2^{-5};2^{15}]$.
				
				gamma: array of 13 equally spaced numbers on a log scale in the range $[2^{-15};2^{3}]$, \code{scikit-learn} default: auto %$[2^{-15};2^{3}]$.
				
%				We used the RBF kernel. These defaults were suggested by other empirical studies \cite{hsu_practical_2016}.
				}
				{This estimator is part of the Support Vector family of algorithms. In this study, we use the Gaussian kernel only. For reference see \cite{cortes_support-vector_1995} and \cite{bishop_pattern_2006}, Chapter 7. The performance of support vector machine is very sensitive with respect to tuning parameters:
			\begin{itemize}
				\item \textbf{C}, the regularization parameter. There does not seem to be a consensus in the community regarding the space for the C hyper-parameter search. In an example\footnote{At the time of writing this paper the example was available on this link: \linkA}
				the \code{scikit-learn} documentation refers to an initial hyper-parameter search space for C in the range $[10^{-2};10^{10}]$ \cite{scikit-learn_model_2018}. However, a different example\footnote{At the time of writing this paper the example was available on this link: \linkB} suggests $[1,10,100,1000]$. A third \code{scikit-learn} example\footnote{At the time of writing this paper the example was available on this link: \linkC} suggests testing for both the linear and rbf kernels and broad values for the C and $\gamma$ parameters. Other researches \cite{hsu_practical_2016} suggest to use apply a search for C in the range $[2^{-5}; 2^{15}]$ which we used in our study as it provides a good compromise between reasonable running time and comprehensiveness of the search space.

				\item \textbf{$\gamma$}, the inverse kernel bandwith. The \code{scikit-learn} example\footnote{At the time of writing this paper the example was available on this link: \linkA} \cite{scikit-learn_model_2018} suggests hyper-parameter search space for $\gamma$ in the range $[10^{-9};10^3]$. However, a second \code{scikit-learn} example\footnote{At the time of writing this paper the example was available on this link: \linkB} suggest to search only in $[0.0001, 0.001]$. On the other hand, \cite{hsu_practical_2016} suggest searching for $\gamma$ in the range $[2^{-15}; 2^{3}]$ which again we found to be the middle ground and applied in our study.
				
			\end{itemize}

}
			
		\item Ensemble Methods
				
				The three main models that we used in this study that are part of this family are the RandomForest, Bagging and Boosting. The three models are built around the logic of using the predictions of a large number of weak estimators, such as decision trees. As such they share a lot of the same hyperparameters. Namely, some of the main parameters for this family of models are the number of estimators, max number of features and the maximum tree depth, default values for each estimator which are suggested in the \code{scikit-learn} package. Recent research \cite{probst_tunability:_2018} and informal consensus in the community suggest that the performance gains from deviating from the default parameters are rewarded for the Boosting algorithm but tend to have limited improvements for the RandomForest algorithm. As such, for the purposes of this study we will focus our efforts to tune the Boosting and Bagging algorithms but will use a relatively small parameter search space for tuning RandomForest.
				
				\desestim{sklearn.ensemble.GradientBoostingClassifier}
				{number of estimators: $[10, 50, 100]$,  \code{scikit-learn} default: 100.
					
				max depth: integers in the range $[1;10]$, \code{scikit-learn} default: 3.
				
				}
				{Part of the ensemble meta-estimators family of models. We used the default sklearn \textit{deviance} loss. The algorithm fits a series of decision trees on the data and predictions are made based on a majority vote. At each iteration the data is modified by applying weights to it and predictions are made again. At each iteration the weights of the incorrectly $x,y$ pairs are increased (boosted) and decreased for the correctly predicted pairs.
					
				As per the \code{scikit-learn} documentation \cite{scikit-learn_model_2018} This estimator is not recommended for datasets with more than two classes as it requires the introduction of regression tress at each iteration. The suggested approach is to use the RandomForest algorithm instead. A lot of the datasets used in this study are multiclass supervised learning problems. However, for the purposes of this study we will use the Gradient Boosting algorithm in order to see how it performs when benchmarked to the suggested approach. For reference see \cite{friedman_greedy_2001} and \cite{efron_computer_2016}, Chapter 17.}
				
				\desestim{sklearn.ensemble.RandomForestClassifier}
				{For this study we used the following hyperparameter grid:
					
				number of estimators: $[10, 50, 100]$, \code{scikit-learn} default: 10
					
				max features: [auto, sqrt, log2, None], \code{scikit-learn} default: auto
					
				max depth: $[10, 100, \text{ None}]$, \code{scikit-learn} default: None
				}
				{Part of the ensemble meta-estimators family of models. The algorithm fits decision trees on sub-samples of the dataset. The average voting rule is used for making predictions. For reference see \cite{breiman_random_2001} and \cite{efron_computer_2016}, Chapter 17.}
				
				\desestim{sklearn.ensemble.BaggingClassifier}
				{number of estimators: $[10, 50, 100]$, \code{scikit-learn} default: 10}
				{Part of the ensemble meta-estimators family of models. The algorithm draws with replacement feature-label pairs $(x,y)$, trains decision base tree estimators and makes predictions based on voting or averaging rules. For reference see \cite{breiman_bagging_1996} and \cite{efron_computer_2016}, Chapter 17.}
	\end{enumerate}

	\subsubsection{Keras neural network architectures including deep neural networks}
\label{subsection:estimators.keras}

We briefly summarize our choices for hyper-parameters and architecture.

	\begin{itemize}
		\item \textbf{Architecture}. Efforts have been made to make the choice of architectures less arbitrary by suggesting algorithms for finding the optimal neural network architecture \cite{gupta_optimizing_2018}. Other researches have suggested good starting points and best practices that one should adhere to when devising a network architecture \cite{hasanpour_lets_2016, sansone_training_2017}. We also followed the guidelines of \cite{goodfellow_deep_2016}, in particular Chapter 6.4. The authors conducted an empirical study and showed that the accuracy of networks increases as the number of layers grow but the gains are diminishing rapidly beyond 5 layers. These findings are also confirmed by other studies \cite{ba_deep_2013} that question to need to use very deep feed-forward networks. In general, the consensus in the community seems to be that 2-4 hidden layers are sufficient for most feed-forward network architectures. 
One notable exception to this rule seem to be convolutional network architecture which have been showed to perform best when several sequential layers are stacked one after the other. However, this study does \emph{not} make use of convolutional neural networks, as our data is not suitable for these models, in particular because there is no well-specified way to transform samples into a multidimensional array form.
The architectures are given below as their keras code specification.
		
		\item \textbf{Activation function}.	We used the rectified linear unit (ReLu) as our default choice of activation function as has been found to accelerate convergence and is relatively inexpensive to perform \cite{krizhevsky_imagenet_2017}.
		
		\item \textbf{Regularization}. We employ the current state-of-art in neural network regularization: dropout. In the absence of clear rules when and where dropout should be applied, we include two versions of each neural network in the study: one version \emph{not} using dropout, and one using dropout. Dropout regularization is as described by~\citet{hinton_improving_2012, srivastava_dropout:_2014} where its potential for improving the generalization accuracy of neural networks is shown. We used a dropout rate of 0.5 as suggested by the authors.

		\item \textbf{Hyper-parameter tuning}. We did not perform grid search to find the optimal hyper parameters for the network. The reason for this is two-fold. We interfaced the neural network models from \textit{keras}. The \textit{keras} interface is not fully compatible with \textit{scikit learn's GridSearch}, nor does it provide easy off-shelf tuning facilities (see subsection ~\ref{subsection:estimators_module} for details).
Furthermore, using grid search tuning does not seem to be considered common practice by the community, and it is even actively recommended to avoid by some researchers~\cite{bergstra_random_2012}, hence might not be considered a fair representation of the state-of-art. Instead, the prevalent practice seems to be manual tuning of hyper-parameters based on learning curves.
Following the latter in the absence of off-shelf automation, we manually tuned learning rate, batch size, and number of epochs by manual inspection of learning curves and performances on the full \emph{training sets} (see below).

		\item \textbf{Learning Rate}. The learning rate is one of the crucial hyper-parameter choices when training neural networks. The generally accepted rule to find the optimal rate is to start with a large rate and if the training process does not diverge decrease the learning rate by a factor of 3~\cite{bengio_practical_2012}. This approach is confirmed by~\cite{srivastava_dropout:_2014} who also affirm that a larger learning rate can be used in conjunction with dropout without risking that the weights of the model blow out.
		
		\item \textbf{Batch Size}. The datasets used in the study were relatively small and could fit in the memory of the machine that we used for training the algorithms. As a result we set the batch size to equal the entire dataset which is equivalent to full gradient descent.
		
		\item \textbf{Number of epochs}. We performed manual hyper-parameter selection by inspection of individual learning curves for all combinations of learning rate and architecture. For this, learning curves on individual data sets' training samples were inspected visually for the ``plateau range'' (range of minimal training error). For all architectures, and most data sets, the plateau  was already reached for \emph{one single epoch}, and training error usually tended to increase in the range of 50-500 epochs. The remaining, small number of datasets (most of which were of 4-or-above-digit sample size) plateaued in the 1-digit range.\\
While this is a \emph{very surprising finding} as it corresponds to a single gradient descent step, \emph{it is what we found}, while following what we consider the standard manual tuning steps for neural networks. We further discuss this in Section~\ref{subsection:exp.disc} and acknowledge that \emph{this surprising finding warrants further investigation}, e.g., through checking for mistakes, or including neural networks tuned by automated schemes.\\
Thus, all neural networks architectures were trained for \emph{one single epoch} - since choosing a larger (and more intuitive number of epochs) would have been somewhat arbitrary, and not in concordance with the common manual tuning protocol.
\end{itemize}
	
For the keras models, we adopted six neural network architectures with varying depths and widths. Our literature review revealed that there is no consistent body of knowledge or concrete rules pertaining to constructing neural network models for simple supervised classification (as opposed to image recognition etc). Therefore, we extrapolated from general best practice guidelines as applicable to our study, and also included (shallow) network architectures that were previously used in benchmark studies. The full \textit{keras} architecture of the neural networks used are listed below.\\
	
	\desestim{keras.models.Sequential}
	{batch size: None, learning rate: $[1, 0.01, 0.001]$, loss: mean squared error, optimizer: Adam, metrics: accuracy.}
	{Own architecture of Deep Neural Network model applying the principles highlighted above. For this experiment we made used of the empirical evidence that networks of 3-4 layers were sufficient to learn any function discussed in \cite{ba_deep_2013}. However, we opted for a slightly narrower network in order to investigate whether wider nets tend to perform better than narrow ones.
				
\begin{lstlisting}[language=Python]
model = OverwrittenSequentialClassifier()
model.add(Dense(288, input_dim=input_dim, activation='relu'))
model.add(Dense(144, activation='relu'))
model.add(Dropout(0.5))
model.add(Dense(12, activation='relu'))
model.add(Dense(num_classes, activation='softmax'))

model_optimizer = optimizers.Adam(lr=lr)
model.compile(loss='mean_squared_error', optimizer=model_optimizer, metrics=['accuracy'])
\end{lstlisting}}

	\desestim{keras.models.Sequential}
	{batch size: None, learning rate: $[1, 0.01, 0.001]$, loss: mean squared error, optimizer: Adam, metrics: accuracy.}
	{In this architecture we experimented with the idea that wider networks perform better than narrower ones. No dropout was performed in order to test the idea that regularization is necessary for all deep neural network models.
	
\begin{lstlisting}[language=Python]
nn_deep_model = OverwrittenSequentialClassifier()
nn_deep_model.add(Dense(2500, input_dim=input_dim, activation='relu'))
nn_deep_model.add(Dense(2000, activation='relu'))
nn_deep_model.add(Dense(1500, activation='relu'))
nn_deep_model.add(Dense(num_classes, activation='softmax'))

model_optimizer = optimizers.Adam(lr=lr)
nn_deep_model.compile(loss='mean_squared_error', optimizer=model_optimizer, metrics=['accuracy'])

\end{lstlisting}}

	\desestim{keras.models.Sequential}
	{batch size: None, learning $[1, 0.01, 0.001]$, loss: mean squared error, optimizer: Adam, metrics: accuracy.}
	{We tested the same architecture as above but applying dropout after the first two layers.
		
\begin{lstlisting}[language=Python]
nn_deep_model = OverwrittenSequentialClassifier()
nn_deep_model.add(Dense(2500, input_dim=input_dim, activation='relu'))
nn_deep_model.add(Dense(2000, activation='relu'))
nn_deep_model.add(Dropout(0.5))
nn_deep_model.add(Dense(1500, activation='relu'))
nn_deep_model.add(Dense(num_classes, activation='softmax'))

model_optimizer = optimizers.Adam(lr=lr)
nn_deep_model.compile(loss='mean_squared_error', optimizer=model_optimizer, metrics=['accuracy'])
\end{lstlisting}}

	\desestim{keras.models.Sequential}
	{batch size: None, learning rate: $[1, 0.01, 0.001]$, loss: mean squared error, optimizer: Adam, metrics: accuracy.}
	{Deep Neural Network model inspired from architecture suggested by \cite{sansone_training_2017}:

\begin{lstlisting}[language=Python]
nn_deep_model = OverwrittenSequentialClassifier()
nn_deep_model.add(Dense(5000, input_dim=input_dim, activation='relu'))
nn_deep_model.add(Dense(4500, activation='relu'))
nn_deep_model.add(Dense(4000, activation='relu'))
nn_deep_model.add(Dropout(0.5))

nn_deep_model.add(Dense(3500, activation='relu'))
nn_deep_model.add(Dense(3000, activation='relu'))
nn_deep_model.add(Dense(2500, activation='relu'))
nn_deep_model.add(Dropout(0.5))

nn_deep_model.add(Dense(2000, activation='relu'))
nn_deep_model.add(Dense(1500, activation='relu'))
nn_deep_model.add(Dense(1000, activation='relu'))
nn_deep_model.add(Dropout(0.5))

nn_deep_model.add(Dense(500, activation='relu'))
nn_deep_model.add(Dense(250, activation='relu'))
nn_deep_model.add(Dense(num_classes, activation='softmax'))

model_optimizer = optimizers.Adam(lr=lr)
nn_deep_model.compile(loss='mean_squared_error', optimizer=model_optimizer, metrics=['accuracy'])
\end{lstlisting}}

	\desestim{keras.models.Sequential}
	{batch size: None, learning rate: $[1,0.01, 0.001]$, loss: mean squared error, optimizer: Adam, metrics: accuracy.}
	{Deep Neural Network model suggested in \cite{hinton_improving_2012} with the following architecture:
\begin{lstlisting}[language=Python]
nn_deep_model = OverwrittenSequentialClassifier()
nn_deep_model.add(Dense(2000, input_dim=input_dim, activation='relu'))
nn_deep_model.add(Dropout(0.5))
nn_deep_model.add(Dense(1000, activation='relu'))
nn_deep_model.add(Dropout(0.5))
nn_deep_model.add(Dense(1000, activation='relu'))
nn_deep_model.add(Dropout(0.5))
nn_deep_model.add(Dense(50, activation='relu'))
nn_deep_model.add(Dropout(0.5))

nn_deep_model.add(Dense(num_classes, activation='softmax'))

model_optimizer = optimizers.Adam(lr=lr)
nn_deep_model.compile(loss='mean_squared_error', optimizer=model_optimizer, metrics=['accuracy'])

\end{lstlisting}}

	\desestim{keras.models.Sequential}
	{batch size: None, learning rate: $[1,0.01, 0.001]$, loss: mean squared error, optimizer: Adam, metrics: accuracy.}
	{Deep Neural Network model suggested in \cite{srivastava_dropout:_2014} with the following architecture:
\begin{lstlisting}[language=Python]
nn_deep_model = OverwrittenSequentialClassifier()
nn_deep_model.add(Dropout(0.7, input_shape=(input_dim,)))
nn_deep_model.add(Dense(1024, activation='relu'))
nn_deep_model.add(Dropout(0.5))
nn_deep_model.add(Dense(num_classes, activation='softmax'))

model_optimizer = optimizers.Adam(lr=lr)
nn_deep_model.compile(loss='mean_squared_error', optimizer=model_optimizer, metrics=['accuracy'])
\end{lstlisting}}

\subsection{Results}

Table~\ref{table:average_performance_estimators} shows an summary overview of results.

	\begin{table}[h!]
		\centering
		\input{tables/avg_metrics.tex}
		\caption{Columns are: average rank (lower is better), classification accuracy, standard error of average score (version c) and training time of the prediction strategy. Performances are estimated as described in Section~\ref{subsection:expsetup}. Rows correspond to prediction strategies, increasingly ordered by their average rank. Naming for sklearn estimators is as in Section~\ref{subsection:estimators.sklearn}. Naming of keras estimators is as in Section~\ref{subsection:estimators.keras}, followed by a string \texttt{dropout} or \texttt{no\_dropout} indicating whether dropout was used, and by a string \texttt{lr} and some number indicating the choice of learning rate.} \label{table:average_performance_estimators}
	\end{table}

\begin{figure}[h!]
		\centering
		\includegraphics[scale=0.7]{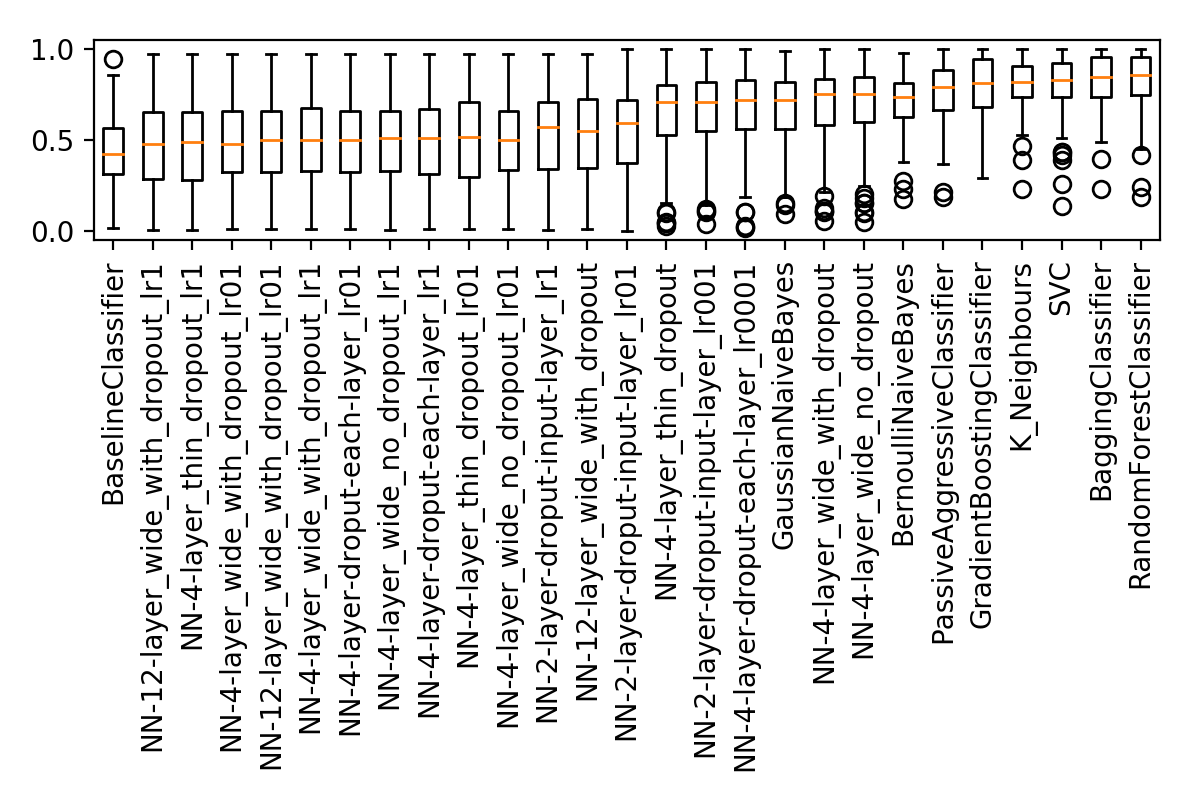}
		\caption{Box-and-whiskers plot of samples of classification accuracy performances classification accuracy by method, ranging over data sets, averaged over the test sample within each dataset. y-axis is classification accuracy. x-axis correspond to prediction strategies, ordered by mean classification accuracy. Naming for sklearn estimators is as in Section~\ref{subsection:estimators.sklearn}. Naming of keras estimators is as in Section~\ref{subsection:estimators.keras}, followed by a string \texttt{dropout} or \texttt{no\_dropout} indicating whether dropout was used, and by a string \texttt{lr} and some number indicating the choice of learning rate. Whisker length is limited at 1.5 times interquartile range.} \label{fig:boxchart_results}
\end{figure}

Figure~\ref{fig:boxchart_results} summarizes the samples of performances in terms of classification accuracy. The sample is performance by method, ranging over data sets, averaged over the test sample within each dataset - i.e., the size of the sample of performance equals the number of data sets in the collection.
	
\begin{figure}[h!]
		\centering
		\includegraphics[scale=0.7]{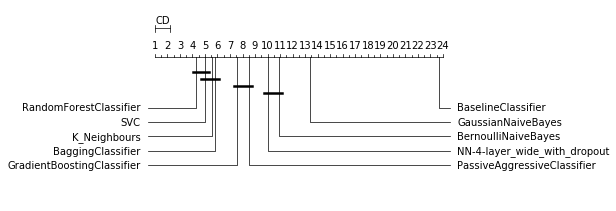}
		\caption{Nem\'enyi post-hoc critical differences comparison diagram after~\cite{demsar_statistical_2006}. CD = critical average rank difference range. x-axis displays average rank (lower is better). Indicated x-axis location = average ranks of prediction strategy, with strategies of below-critical average rank difference connected by a bar. Naming for sklearn estimators is as in Section~\ref{subsection:estimators.sklearn}. Naming of keras estimators is as in Section~\ref{subsection:estimators.keras}, followed by a string \texttt{dropout} or \texttt{no\_dropout} indicating whether dropout was used, and by a string \texttt{lr} and some number indicating the choice of learning rate. Note that for the sake of readability, the worst performing neural networks were removed from the plot. } \label{fig:nemeniy_test_critical_distance}
	\end{figure}

The Friedman test was significant at level p=2e-16. Figure~\ref{fig:nemeniy_test_critical_distance} displays effect sizes, i.e., average ranks with Nem\'enyi's post-hoc critical differences. 
	
From all the above, the top five algorithms among the contenders were the Random Forest, SVC, Bagging, K Neighbours and Gradient Boosting classifiers.

Further benchmarking results may be found in the automatically generated Appendix~\ref{section:statistical_appendix}. These include results of paired t-tests and Wilcoxon signed rank tests. Briefly summarizing these: Neither t-test (Appendix~\ref{table:t_test}), nor the Wilcoxon signed rank test (Appendix~\ref{table:wilcoxon}), with Bonferroni correction (adjacent strategies and all vs baseline), in isolation, are able to reject the null hypothesis of a performance difference between any two of the top five performers.

\subsection{Discussion}
\label{subsection:exp.disc}

We discuss our findings below, including a comparison with the benchmarking study by~\citet{fernandez-delgado_we_2014}.

\subsubsection{Key findings}
In summary, the key findings of the benchmarking study are:
\begin{enumerate}
\itemsep-0.2em
\item[(i)] MLaut is capable of carrying out large-scale benchmarking experiments across a representative selection of off-shelf supervised learning strategies, including state-of-art deep learning models, and a selection of small-to-moderate-sized basic supervised learning benchmark data sets.
\item[(ii)] On the selection of benchmark data sets representative for basic (non-specialized) supervised learning, the best performing algorithms are ensembles of trees and kernel-based algorithms. Neural networks (deep or not) perform poorly in comparison.
\item[(iii)] Of the algorithms benchmarked, grid-tuned support vector classifiers are the most demanding of computation time. Neural networks (deep or not) and the other algorithms benchmarked require computation time in a comparable orders of magnitude.
\end{enumerate}

\subsubsection{Limitations}
The main limitations of our study are:
\begin{enumerate}
\itemsep-0.2em
\item[(i)] restriction to the Delgado data set collection. Our study is at most as representative for the methods' performance as the Delgado data set collection is for basic supervised learning.
\item[(ii)] training the neural networks for one epoch only. As described in~\ref{subsection:estimators.keras} we believe we arrived at this choice following standard tuning protocol, but it requires further investigation, especially to rule out a mistake - or to corroborate evidence of a potential general issue of neural networks with basic supervised learning (i.e., not on image, audio, text data etc).
\item[(iii)] A relative small set of prediction strategies. While our study is an initial proof-of-concept for MLaut on commonly used algorithms, it did not include composite strategies (e.g., full pipelines), or the full selection available in state-of-art packages.
\end{enumerate}

\subsubsection{Comparison to the study of Delgado et al}

In comparison to the benchmarking study of~\citet{fernandez-delgado_we_2014}, for most algorithms we find comparable performances which are within 95\% confidence bands (of ours). A notable major departure is performance of the neural networks, which we find to be substantially worse. The latter finding may be plausibly explained by at least one of the following:
\begin{enumerate}
\itemsep-0.2em
\item[(i)] an issue-in-principle with how we tuned the neural networks - e.g., a mistake; or a difference to how the neural networks were tuned by~\citet{fernandez-delgado_we_2014}. However, it appears that~\citet{fernandez-delgado_we_2014} used default settings.
\item[(ii)] an overly optimistic bias of~\citet{fernandez-delgado_we_2014}, through their mistake of tuning on the test set. This bias would be expected to be most severe for the models with most degrees of freedom to tune - i.e., the neural networks.
\end{enumerate}

In additional comparison, the general rankings (when disregarding the neural networks) are similar. Though, since a replication of rankings is dependent on conducting the study on exactly the same set of strategies, we are only able to state this qualitatively. Conversely, our confidence intervals indicate that rankings in general are very unstable on the data set collection, as roughly a half of the 179 classifiers which~\citet{fernandez-delgado_we_2014} benchmarked seem to be within 95\% confidence ranges of each other. 

This seems to highlight the crucial necessity of reporting not only performances but also confidence bands, if reasoning is to be conducted about which algorithmic strategies are the ``best'' performing ones.

\subsubsection{Conclusions}
\label{subsection:conclusions}

Our findings corroborate most of the findings of the major existing benchmarking study of~\citet{fernandez-delgado_we_2014}. In addition, we validate the usefulness of MLaut to easily conduct such a study.
	
As a notable exception to this confirmation of results, we find that neural networks do not perform well on ``basic'' supervised classification data sets. While it may be explained by a bias that~\citet{fernandez-delgado_we_2014} introduced into their study by the mistake of tuning on the test set, it is still under the strong caveat that further investigation needs to be carried out, in particular with respect to the tuning behaviour of said networks, and our experiment not containing other mistakes.

However, if further investigation confirms our findings, it would be consistent with the findings of one of the original dropout papers~\cite{srivastava_dropout:_2014}, in which the authors also conclude that the improvements are more noticeable on image datasets and less so on other types of data such as text. For example, the authors found that the performance improvements achieved on the Reuters RCV1 corpus were not significant in comparison with architectures that did not use dropout. 
Furthermore, at least in our study we found no evidence to suggest that deep architectures performed better than shallow ones. In fact the 12 layer deep neural network architecture ranked just slightly better than our baseline classifier. Our findings also may suggest that wide architectures tend to perform better than thin ones on our training data.
It should also be pointed out that the datasets we used in this experiment were relatively small in size. Therefore, it could be argued that deep neural networks can easily overfit such data, the default parameter choices and standard procedures are not appropriate - especially since such common practice may arguably be strongly adapted to image/audio/text data.
	
In terms of training time, the SVC algorithm proved to be the most expensive, taking on average almost 30 min to train in our set-up. However, it should be noted that this is due to the relatively large hyper-parameter search space that we used. On the other hand, among the top five algorithms the Bagging Classifier was one of the least expensive ones to train taking an average of only 5 seconds. Our top performer, the Random Forest Classifier, was also relatively inexpensive to train taking an average of only 14 seconds.

As our main finding, however, we consider the ease with which a user may generate the above results, using MLaut. The reader may (hopefully) convince themselves of this by inspecting the code and jupyter notebooks in the repository \href{https://github.com/alan-turing-institute/mlaut/blob/master/examples/mlaut_study.zip}.
We are also very appreciative of any criticism, or suggestions for improvement, made (say, by an unconvinced reader) through the project's issue tracker. 

%% file: tables/avg_metrics.tex
\begin{tabular}{lrrrr}
	\toprule
	{} &  avg\_rank &  avg\_score &  std\_error &  avg training time (in sec) \\
	\midrule
	RandomForestClassifier              &       4.3 &      0.831 &      0.013 &                      14.277 \\
	SVC                                 &       5.0 &      0.818 &      0.014 &                    1742.466 \\
	K\_Neighbours                        &       5.6 &      0.805 &      0.014 &                     107.796 \\
	BaggingClassifier                   &       5.8 &      0.820 &      0.014 &                       5.231 \\
	GradientBoostingClassifier          &       7.6 &      0.790 &      0.016 &                      49.509 \\
	PassiveAggressiveClassifier         &       8.5 &      0.758 &      0.016 &                      19.352 \\
	NN-4-layer\_wide\_with\_dropout\_lr001  &      10.0 &      0.692 &      0.021 &                      14.617 \\
	NN-4-layer\_wide\_no\_dropout\_lr001    &      10.5 &      0.694 &      0.021 &                      14.609 \\
	BernoulliNaiveBayes                 &      10.9 &      0.707 &      0.015 &                       0.005 \\
	NN-4-layer-droput-each-layer\_lr0001 &      11.2 &      0.662 &      0.022 &                       6.786 \\
	NN-4-layer\_thin\_dropout\_lr001       &      11.6 &      0.652 &      0.022 &                       2.869 \\
	NN-2-layer-droput-input-layer\_lr001 &      11.7 &      0.655 &      0.021 &                       5.420 \\
	GaussianNaiveBayes                  &      13.4 &      0.674 &      0.019 &                       0.004 \\
	NN-12-layer\_wide\_with\_dropout\_lr001 &      16.3 &      0.535 &      0.023 &                      40.003 \\
	NN-2-layer-droput-input-layer\_lr01  &      17.3 &      0.543 &      0.023 &                       5.413 \\
	NN-2-layer-droput-input-layer\_lr1   &      17.9 &      0.509 &      0.023 &                       5.437 \\
	NN-4-layer\_thin\_dropout\_lr01        &      18.0 &      0.494 &      0.024 &                       5.559 \\
	NN-4-layer\_wide\_no\_dropout\_lr01     &      18.4 &      0.494 &      0.022 &                      10.530 \\
	NN-4-layer-droput-each-layer\_lr1    &      18.5 &      0.488 &      0.022 &                       6.901 \\
	NN-4-layer\_wide\_with\_dropout\_lr1    &      18.5 &      0.483 &      0.022 &                      10.738 \\
	NN-4-layer\_wide\_no\_dropout\_lr1      &      18.6 &      0.490 &      0.022 &                      10.561 \\
	NN-4-layer\_wide\_with\_dropout\_lr01   &      18.7 &      0.478 &      0.022 &                      10.696 \\
	NN-4-layer-droput-each-layer\_lr01   &      18.8 &      0.482 &      0.022 &                       6.818 \\
	NN-12-layer\_wide\_with\_dropout\_lr01  &      18.8 &      0.479 &      0.022 &                      70.574 \\
	NN-12-layer\_wide\_with\_dropout\_lr1   &      19.0 &      0.458 &      0.023 &                      68.505 \\
	NN-4-layer\_thin\_dropout\_lr1         &      19.4 &      0.462 &      0.023 &                       4.299 \\
	BaselineClassifier                  &      23.7 &      0.419 &      0.019 &                       0.001 \\
	\bottomrule
\end{tabular}

%% file: inputs/appendix.tex
\begin{landscape}

\section{Further benchmarking results} \label{section:statistical_appendix}

\subsection{paired t-test, without multiple testing correction} \label{table:t_test}
\begin{adjustbox}{max width=\linewidth}
\input{tables/t_test1.tex}
\end{adjustbox}
\clearpage
\begin{adjustbox}{max width=\linewidth}
\input{tables/t_test2.tex}
\end{adjustbox}
\begin{adjustbox}{max width=\linewidth}
	\input{tables/t_test3.tex}
\end{adjustbox}
\begin{adjustbox}{max width=\linewidth}
	\input{tables/t_test4.tex}
\end{adjustbox}
\begin{adjustbox}{max width=\linewidth}
	\input{tables/t_test5.tex}
\end{adjustbox}
\begin{adjustbox}{max width=\linewidth}
	\input{tables/t_test6.tex}
\end{adjustbox}
\begin{adjustbox}{max width=\linewidth}
	\input{tables/t_test7.tex}
\end{adjustbox}

%
%\subsection{paired t-test, with Bonferroni correction} \label{table:t_test_bonferroni}
%\begin{adjustbox}{max width=\linewidth}
%	\input{tables/t_test_bonferroni1.tex}
%\end{adjustbox}
%\clearpage
%\begin{adjustbox}{max width=\linewidth}
%	\input{tables/t_test_bonferroni2.tex}
%\end{adjustbox}
%\begin{adjustbox}{max width=\linewidth}
%	\input{tables/t_test_bonferroni3.tex}
%\end{adjustbox}
%\begin{adjustbox}{max width=\linewidth}
%	\input{tables/t_test_bonferroni4.tex}
%\end{adjustbox}
%\begin{adjustbox}{max width=\linewidth}
%	\input{tables/t_test_bonferroni5.tex}
%\end{adjustbox}
%\begin{adjustbox}{max width=\linewidth}
%	\input{tables/t_test_bonferroni6.tex}
%\end{adjustbox}
%\begin{adjustbox}{max width=\linewidth}
%	\input{tables/t_test_bonferroni7.tex}
%\end{adjustbox}
%
%\subsection{Sign test} \label{table:sign_test}
%\begin{adjustbox}{max width=\linewidth}
%	\input{tables/sign_test1.tex}
%\end{adjustbox}
%\clearpage
%\begin{adjustbox}{max width=\linewidth}
%	\input{tables/sign_test2.tex}
%\end{adjustbox}
%\begin{adjustbox}{max width=\linewidth}
%	\input{tables/sign_test3.tex}
%\end{adjustbox}
%\begin{adjustbox}{max width=\linewidth}
%	\input{tables/sign_test4.tex}
%\end{adjustbox}
%\begin{adjustbox}{max width=\linewidth}
%	\input{tables/sign_test5.tex}
%\end{adjustbox}

\subsection{Wilcoxon signed-rank test, without Bonferroni correction} \label{table:wilcoxon}
\begin{adjustbox}{max width=\linewidth}
	\input{tables/wilcoxon_test1.tex}
\end{adjustbox}
\clearpage
\begin{adjustbox}{max width=\linewidth}
	\input{tables/wilcoxon_test2.tex}
\end{adjustbox}
\begin{adjustbox}{max width=\linewidth}
	\input{tables/wilcoxon_test3.tex}
\end{adjustbox}
\begin{adjustbox}{max width=\linewidth}
	\input{tables/wilcoxon_test4.tex}
\end{adjustbox}
\begin{adjustbox}{max width=\linewidth}
	\input{tables/wilcoxon_test5.tex}
\end{adjustbox}
\begin{adjustbox}{max width=\linewidth}
	\input{tables/wilcoxon_test6.tex}
\end{adjustbox}
\begin{adjustbox}{max width=\linewidth}
	\input{tables/wilcoxon_test7.tex}
\end{adjustbox}

%
%\subsection{Ranksum test} \label{table:ranksum}
%\begin{adjustbox}{max width=\linewidth}
%	\input{tables/ranksum_test1.tex}
%\end{adjustbox}
%\clearpage
%\begin{adjustbox}{max width=\linewidth}
%	\input{tables/ranksum_test2.tex}
%\end{adjustbox}
%\clearpage
%\begin{adjustbox}{max width=\linewidth}
%	\input{tables/ranksum_test3.tex}
%\end{adjustbox}
%\clearpage
%\begin{adjustbox}{max width=\linewidth}
%	\input{tables/ranksum_test4.tex}
%\end{adjustbox}
%\clearpage
%\begin{adjustbox}{max width=\linewidth}
%	\input{tables/ranksum_test5.tex}
%\end{adjustbox}
%\clearpage
%\begin{adjustbox}{max width=\linewidth}
%	\input{tables/ranksum_test6.tex}
%\end{adjustbox}
%\clearpage
%\begin{adjustbox}{max width=\linewidth}
%	\input{tables/ranksum_test7.tex}
%\end{adjustbox}
%
\end{landscape}

%\subsection{Friedman test}
%\begin{table}[h!]
%	\centering
%	\input{tables/friedman_test.tex}
%	\label{table:friedman}
%\end{table}

%\clearpage

\begin{landscape}
\subsection{Nem\'enyi post-hoc significance} \label{table:nemeniy}
\begin{adjustbox}{max width=\linewidth}
	\input{tables/nemeniy_test1.tex}
\end{adjustbox}
\clearpage
\begin{adjustbox}{max width=\linewidth}
	\input{tables/nemeniy_test2.tex}
\end{adjustbox}
\begin{adjustbox}{max width=\linewidth}
	\input{tables/nemeniy_test3.tex}
\end{adjustbox}
\begin{adjustbox}{max width=\linewidth}
	\input{tables/nemeniy_test4.tex}
\end{adjustbox}
\begin{adjustbox}{max width=\linewidth}
	\input{tables/nemeniy_test5.tex}
\end{adjustbox}
\begin{adjustbox}{max width=\linewidth}
	\input{tables/nemeniy_test6.tex}
\end{adjustbox}
\begin{adjustbox}{max width=\linewidth}
	\input{tables/nemeniy_test7.tex}
\end{adjustbox}

\end{landscape}

\subsection{Accuracy by data set, type (a) standard errors} \label{table:avg_performance_per_estimator}
\input{tables/errors_per_dataset_per_estimator.tex}
\label{table:performance_per_dataset_and_per_estimator}

%% file: tables/t_test1.tex
\begin{tabular}{lrrrrrrrr}
\toprule
{} & \multicolumn{2}{l}{BaggingClassifier} & \multicolumn{2}{l}{BaselineClassifier} & \multicolumn{2}{l}{BernoulliNaiveBayes} & \multicolumn{2}{l}{GaussianNaiveBayes} \\
{} &            t\_stat &  p\_val &             t\_stat &  p\_val &              t\_stat &  p\_val &             t\_stat &  p\_val \\
\midrule
BaggingClassifier                   &             0.000 &  1.000 &             16.779 &  0.000 &               5.456 &  0.000 &              6.137 &  0.000 \\
BaselineClassifier                  &           -16.779 &  0.000 &              0.000 &  1.000 &             -11.764 &  0.000 &             -9.373 &  0.000 \\
BernoulliNaiveBayes                 &            -5.456 &  0.000 &             11.764 &  0.000 &               0.000 &  1.000 &              1.366 &  0.173 \\
GaussianNaiveBayes                  &            -6.137 &  0.000 &              9.373 &  0.000 &              -1.366 &  0.173 &              0.000 &  1.000 \\
GradientBoostingClassifier          &            -1.407 &  0.161 &             14.703 &  0.000 &               3.719 &  0.000 &              4.610 &  0.000 \\
K\_Neighbours                        &            -0.734 &  0.463 &             16.373 &  0.000 &               4.837 &  0.000 &              5.600 &  0.000 \\
NN-12-layer\_wide\_with\_dropout       &           -10.631 &  0.000 &              3.963 &  0.000 &              -6.273 &  0.000 &             -4.629 &  0.000 \\
NN-12-layer\_wide\_with\_dropout\_lr01  &           -13.058 &  0.000 &              2.045 &  0.042 &              -8.562 &  0.000 &             -6.687 &  0.000 \\
NN-12-layer\_wide\_with\_dropout\_lr1   &           -13.606 &  0.000 &              1.309 &  0.192 &              -9.183 &  0.000 &             -7.296 &  0.000 \\
NN-2-layer-droput-input-layer\_lr001 &            -6.453 &  0.000 &              8.206 &  0.000 &              -2.001 &  0.047 &             -0.660 &  0.510 \\
NN-2-layer-droput-input-layer\_lr01  &           -10.186 &  0.000 &              4.101 &  0.000 &              -5.927 &  0.000 &             -4.347 &  0.000 \\
NN-2-layer-droput-input-layer\_lr1   &           -11.574 &  0.000 &              3.020 &  0.003 &              -7.230 &  0.000 &             -5.521 &  0.000 \\
NN-4-layer-droput-each-layer\_lr0001 &            -5.912 &  0.000 &              8.451 &  0.000 &              -1.557 &  0.121 &             -0.278 &  0.781 \\
NN-4-layer-droput-each-layer\_lr01   &           -12.754 &  0.000 &              2.109 &  0.036 &              -8.336 &  0.000 &             -6.514 &  0.000 \\
NN-4-layer-droput-each-layer\_lr1    &           -12.640 &  0.000 &              2.350 &  0.020 &              -8.177 &  0.000 &             -6.346 &  0.000 \\
NN-4-layer\_thin\_dropout             &            -6.405 &  0.000 &              8.003 &  0.000 &              -2.042 &  0.042 &             -0.723 &  0.471 \\
NN-4-layer\_thin\_dropout\_lr01        &           -12.112 &  0.000 &              2.299 &  0.022 &              -7.839 &  0.000 &             -6.117 &  0.000 \\
NN-4-layer\_thin\_dropout\_lr1         &           -13.293 &  0.000 &              1.411 &  0.159 &              -8.937 &  0.000 &             -7.100 &  0.000 \\
NN-4-layer\_wide\_no\_dropout          &            -4.958 &  0.000 &              9.704 &  0.000 &              -0.477 &  0.634 &              0.742 &  0.459 \\
NN-4-layer\_wide\_no\_dropout\_lr01     &           -12.554 &  0.000 &              2.500 &  0.013 &              -8.067 &  0.000 &             -6.234 &  0.000 \\
NN-4-layer\_wide\_no\_dropout\_lr1      &           -12.618 &  0.000 &              2.316 &  0.021 &              -8.174 &  0.000 &             -6.351 &  0.000 \\
NN-4-layer\_wide\_with\_dropout        &            -5.043 &  0.000 &              9.661 &  0.000 &              -0.548 &  0.584 &              0.680 &  0.497 \\
NN-4-layer\_wide\_with\_dropout\_lr01   &           -13.170 &  0.000 &              1.892 &  0.060 &              -8.690 &  0.000 &             -6.813 &  0.000 \\
NN-4-layer\_wide\_with\_dropout\_lr1    &           -12.877 &  0.000 &              2.118 &  0.035 &              -8.416 &  0.000 &             -6.568 &  0.000 \\
PassiveAggressiveClassifier         &            -2.876 &  0.004 &             13.497 &  0.000 &               2.313 &  0.022 &              3.371 &  0.001 \\
RandomForestClassifier              &             0.253 &  0.800 &             16.752 &  0.000 &               5.597 &  0.000 &              6.262 &  0.000 \\
SVC                                 &            -0.607 &  0.544 &             15.781 &  0.000 &               4.660 &  0.000 &              5.443 &  0.000 \\
\bottomrule
\end{tabular}

%% file: tables/t_test2.tex
\begin{tabular}{lrrrrrrrr}
\toprule
{} & \multicolumn{2}{l}{GradientBoostingClassifier} & \multicolumn{2}{l}{K\_Neighbours} & \multicolumn{2}{l}{NN-12-layer\_wide\_with\_dropout} & \multicolumn{2}{l}{NN-12-layer\_wide\_with\_dropout\_lr01} \\
{} &                     t\_stat &  p\_val &       t\_stat &  p\_val &                        t\_stat &  p\_val &                             t\_stat &  p\_val \\
\midrule
BaggingClassifier                   &                      1.407 &  0.161 &        0.734 &  0.463 &                        10.631 &  0.000 &                             13.058 &  0.000 \\
BaselineClassifier                  &                    -14.703 &  0.000 &      -16.373 &  0.000 &                        -3.963 &  0.000 &                             -2.045 &  0.042 \\
BernoulliNaiveBayes                 &                     -3.719 &  0.000 &       -4.837 &  0.000 &                         6.273 &  0.000 &                              8.562 &  0.000 \\
GaussianNaiveBayes                  &                     -4.610 &  0.000 &       -5.600 &  0.000 &                         4.629 &  0.000 &                              6.687 &  0.000 \\
GradientBoostingClassifier          &                      0.000 &  1.000 &       -0.749 &  0.454 &                         9.091 &  0.000 &                             11.374 &  0.000 \\
K\_Neighbours                        &                      0.749 &  0.454 &        0.000 &  1.000 &                        10.190 &  0.000 &                             12.634 &  0.000 \\
NN-12-layer\_wide\_with\_dropout       &                     -9.091 &  0.000 &      -10.190 &  0.000 &                         0.000 &  1.000 &                              1.837 &  0.068 \\
NN-12-layer\_wide\_with\_dropout\_lr01  &                    -11.374 &  0.000 &      -12.634 &  0.000 &                        -1.837 &  0.068 &                              0.000 &  1.000 \\
NN-12-layer\_wide\_with\_dropout\_lr1   &                    -11.936 &  0.000 &      -13.193 &  0.000 &                        -2.470 &  0.014 &                             -0.665 &  0.507 \\
NN-2-layer-droput-input-layer\_lr001 &                     -5.027 &  0.000 &       -5.953 &  0.000 &                         3.805 &  0.000 &                              5.751 &  0.000 \\
NN-2-layer-droput-input-layer\_lr01  &                     -8.702 &  0.000 &       -9.748 &  0.000 &                         0.190 &  0.850 &                              2.002 &  0.046 \\
NN-2-layer-droput-input-layer\_lr1   &                    -10.008 &  0.000 &      -11.144 &  0.000 &                        -0.855 &  0.394 &                              0.961 &  0.338 \\
NN-4-layer-droput-each-layer\_lr0001 &                     -4.541 &  0.000 &       -5.415 &  0.000 &                         4.099 &  0.000 &                              6.021 &  0.000 \\
NN-4-layer-droput-each-layer\_lr01   &                    -11.115 &  0.000 &      -12.332 &  0.000 &                        -1.734 &  0.084 &                              0.084 &  0.933 \\
NN-4-layer-droput-each-layer\_lr1    &                    -10.985 &  0.000 &      -12.214 &  0.000 &                        -1.540 &  0.125 &                              0.295 &  0.768 \\
NN-4-layer\_thin\_dropout             &                     -5.013 &  0.000 &       -5.914 &  0.000 &                         3.686 &  0.000 &                              5.602 &  0.000 \\
NN-4-layer\_thin\_dropout\_lr01        &                    -10.559 &  0.000 &      -11.693 &  0.000 &                        -1.474 &  0.142 &                              0.310 &  0.757 \\
NN-4-layer\_thin\_dropout\_lr1         &                    -11.665 &  0.000 &      -12.881 &  0.000 &                        -2.338 &  0.020 &                             -0.549 &  0.584 \\
NN-4-layer\_wide\_no\_dropout          &                     -3.581 &  0.000 &       -4.436 &  0.000 &                         5.141 &  0.000 &                              7.126 &  0.000 \\
NN-4-layer\_wide\_no\_dropout\_lr01     &                    -10.891 &  0.000 &      -12.125 &  0.000 &                        -1.416 &  0.158 &                              0.427 &  0.670 \\
NN-4-layer\_wide\_no\_dropout\_lr1      &                    -10.972 &  0.000 &      -12.193 &  0.000 &                        -1.560 &  0.120 &                              0.269 &  0.788 \\
NN-4-layer\_wide\_with\_dropout        &                     -3.658 &  0.000 &       -4.520 &  0.000 &                         5.091 &  0.000 &                              7.079 &  0.000 \\
NN-4-layer\_wide\_with\_dropout\_lr01   &                    -11.489 &  0.000 &      -12.748 &  0.000 &                        -1.968 &  0.050 &                             -0.138 &  0.891 \\
NN-4-layer\_wide\_with\_dropout\_lr1    &                    -11.215 &  0.000 &      -12.454 &  0.000 &                        -1.751 &  0.081 &                              0.079 &  0.937 \\
PassiveAggressiveClassifier         &                     -1.370 &  0.172 &       -2.238 &  0.026 &                         7.982 &  0.000 &                             10.251 &  0.000 \\
RandomForestClassifier              &                      1.618 &  0.107 &        0.976 &  0.330 &                        10.700 &  0.000 &                             13.096 &  0.000 \\
SVC                                 &                      0.791 &  0.430 &        0.086 &  0.931 &                         9.919 &  0.000 &                             12.269 &  0.000 \\
\bottomrule
\end{tabular}

%% file: tables/t_test3.tex
\begin{tabular}{lrrrrrrrr}
\toprule
{} & \multicolumn{2}{l}{NN-12-layer\_wide\_with\_dropout\_lr1} & \multicolumn{2}{l}{NN-2-layer-droput-input-layer\_lr001} & \multicolumn{2}{l}{NN-2-layer-droput-input-layer\_lr01} & \multicolumn{2}{l}{NN-2-layer-droput-input-layer\_lr1} \\
{} &                            t\_stat &  p\_val &                              t\_stat &  p\_val &                             t\_stat &  p\_val &                            t\_stat &  p\_val \\
\midrule
BaggingClassifier                   &                            13.606 &  0.000 &                               6.453 &  0.000 &                             10.186 &  0.000 &                            11.574 &  0.000 \\
BaselineClassifier                  &                            -1.309 &  0.192 &                              -8.206 &  0.000 &                             -4.101 &  0.000 &                            -3.020 &  0.003 \\
BernoulliNaiveBayes                 &                             9.183 &  0.000 &                               2.001 &  0.047 &                              5.927 &  0.000 &                             7.230 &  0.000 \\
GaussianNaiveBayes                  &                             7.296 &  0.000 &                               0.660 &  0.510 &                              4.347 &  0.000 &                             5.521 &  0.000 \\
GradientBoostingClassifier          &                            11.936 &  0.000 &                               5.027 &  0.000 &                              8.702 &  0.000 &                            10.008 &  0.000 \\
K\_Neighbours                        &                            13.193 &  0.000 &                               5.953 &  0.000 &                              9.748 &  0.000 &                            11.144 &  0.000 \\
NN-12-layer\_wide\_with\_dropout       &                             2.470 &  0.014 &                              -3.805 &  0.000 &                             -0.190 &  0.850 &                             0.855 &  0.394 \\
NN-12-layer\_wide\_with\_dropout\_lr01  &                             0.665 &  0.507 &                              -5.751 &  0.000 &                             -2.002 &  0.046 &                            -0.961 &  0.338 \\
NN-12-layer\_wide\_with\_dropout\_lr1   &                             0.000 &  1.000 &                              -6.349 &  0.000 &                             -2.624 &  0.009 &                            -1.602 &  0.111 \\
NN-2-layer-droput-input-layer\_lr001 &                             6.349 &  0.000 &                               0.000 &  1.000 &                              3.552 &  0.000 &                             4.663 &  0.000 \\
NN-2-layer-droput-input-layer\_lr01  &                             2.624 &  0.009 &                              -3.552 &  0.000 &                              0.000 &  1.000 &                             1.031 &  0.304 \\
NN-2-layer-droput-input-layer\_lr1   &                             1.602 &  0.111 &                              -4.663 &  0.000 &                             -1.031 &  0.304 &                             0.000 &  1.000 \\
NN-4-layer-droput-each-layer\_lr0001 &                             6.609 &  0.000 &                               0.354 &  0.724 &                              3.846 &  0.000 &                             4.944 &  0.000 \\
NN-4-layer-droput-each-layer\_lr01   &                             0.741 &  0.460 &                              -5.600 &  0.000 &                             -1.899 &  0.059 &                            -0.868 &  0.386 \\
NN-4-layer-droput-each-layer\_lr1    &                             0.954 &  0.341 &                              -5.430 &  0.000 &                             -1.708 &  0.089 &                            -0.668 &  0.505 \\
NN-4-layer\_thin\_dropout             &                             6.194 &  0.000 &                              -0.070 &  0.945 &                              3.439 &  0.001 &                             4.533 &  0.000 \\
NN-4-layer\_thin\_dropout\_lr01        &                             0.950 &  0.343 &                              -5.247 &  0.000 &                             -1.640 &  0.102 &                            -0.627 &  0.531 \\
NN-4-layer\_thin\_dropout\_lr1         &                             0.109 &  0.914 &                              -6.174 &  0.000 &                             -2.493 &  0.013 &                            -1.479 &  0.141 \\
NN-4-layer\_wide\_no\_dropout          &                             7.710 &  0.000 &                               1.339 &  0.182 &                              4.864 &  0.000 &                             5.999 &  0.000 \\
NN-4-layer\_wide\_no\_dropout\_lr01     &                             1.087 &  0.278 &                              -5.319 &  0.000 &                             -1.587 &  0.114 &                            -0.542 &  0.588 \\
NN-4-layer\_wide\_no\_dropout\_lr1      &                             0.927 &  0.355 &                              -5.439 &  0.000 &                             -1.728 &  0.085 &                            -0.691 &  0.491 \\
NN-4-layer\_wide\_with\_dropout        &                             7.664 &  0.000 &                               1.281 &  0.201 &                              4.815 &  0.000 &                             5.951 &  0.000 \\
NN-4-layer\_wide\_with\_dropout\_lr01   &                             0.528 &  0.598 &                              -5.874 &  0.000 &                             -2.130 &  0.034 &                            -1.093 &  0.275 \\
NN-4-layer\_wide\_with\_dropout\_lr1    &                             0.740 &  0.460 &                              -5.643 &  0.000 &                             -1.916 &  0.057 &                            -0.879 &  0.380 \\
PassiveAggressiveClassifier         &                            10.834 &  0.000 &                               3.867 &  0.000 &                              7.614 &  0.000 &                             8.909 &  0.000 \\
RandomForestClassifier              &                            13.640 &  0.000 &                               6.571 &  0.000 &                             10.261 &  0.000 &                            11.632 &  0.000 \\
SVC                                 &                            12.823 &  0.000 &                               5.808 &  0.000 &                              9.503 &  0.000 &                            10.848 &  0.000 \\
\bottomrule
\end{tabular}

%% file: tables/t_test4.tex
\begin{tabular}{lrrrrrrrr}
\toprule
{} & \multicolumn{2}{l}{NN-4-layer-droput-each-layer\_lr0001} & \multicolumn{2}{l}{NN-4-layer-droput-each-layer\_lr01} & \multicolumn{2}{l}{NN-4-layer-droput-each-layer\_lr1} & \multicolumn{2}{l}{NN-4-layer\_thin\_dropout} \\
{} &                              t\_stat &  p\_val &                            t\_stat &  p\_val &                           t\_stat &  p\_val &                  t\_stat &  p\_val \\
\midrule
BaggingClassifier                   &                               5.912 &  0.000 &                            12.754 &  0.000 &                           12.640 &  0.000 &                   6.405 &  0.000 \\
BaselineClassifier                  &                              -8.451 &  0.000 &                            -2.109 &  0.036 &                           -2.350 &  0.020 &                  -8.003 &  0.000 \\
BernoulliNaiveBayes                 &                               1.557 &  0.121 &                             8.336 &  0.000 &                            8.177 &  0.000 &                   2.042 &  0.042 \\
GaussianNaiveBayes                  &                               0.278 &  0.781 &                             6.514 &  0.000 &                            6.346 &  0.000 &                   0.723 &  0.471 \\
GradientBoostingClassifier          &                               4.541 &  0.000 &                            11.115 &  0.000 &                           10.985 &  0.000 &                   5.013 &  0.000 \\
K\_Neighbours                        &                               5.415 &  0.000 &                            12.332 &  0.000 &                           12.214 &  0.000 &                   5.914 &  0.000 \\
NN-12-layer\_wide\_with\_dropout       &                              -4.099 &  0.000 &                             1.734 &  0.084 &                            1.540 &  0.125 &                  -3.686 &  0.000 \\
NN-12-layer\_wide\_with\_dropout\_lr01  &                              -6.021 &  0.000 &                            -0.084 &  0.933 &                           -0.295 &  0.768 &                  -5.602 &  0.000 \\
NN-12-layer\_wide\_with\_dropout\_lr1   &                              -6.609 &  0.000 &                            -0.741 &  0.460 &                           -0.954 &  0.341 &                  -6.194 &  0.000 \\
NN-2-layer-droput-input-layer\_lr001 &                              -0.354 &  0.724 &                             5.600 &  0.000 &                            5.430 &  0.000 &                   0.070 &  0.945 \\
NN-2-layer-droput-input-layer\_lr01  &                              -3.846 &  0.000 &                             1.899 &  0.059 &                            1.708 &  0.089 &                  -3.439 &  0.001 \\
NN-2-layer-droput-input-layer\_lr1   &                              -4.944 &  0.000 &                             0.868 &  0.386 &                            0.668 &  0.505 &                  -4.533 &  0.000 \\
NN-4-layer-droput-each-layer\_lr0001 &                               0.000 &  1.000 &                             5.870 &  0.000 &                            5.704 &  0.000 &                   0.418 &  0.677 \\
NN-4-layer-droput-each-layer\_lr01   &                              -5.870 &  0.000 &                             0.000 &  1.000 &                           -0.208 &  0.835 &                  -5.455 &  0.000 \\
NN-4-layer-droput-each-layer\_lr1    &                              -5.704 &  0.000 &                             0.208 &  0.835 &                            0.000 &  1.000 &                  -5.286 &  0.000 \\
NN-4-layer\_thin\_dropout             &                              -0.418 &  0.677 &                             5.455 &  0.000 &                            5.286 &  0.000 &                   0.000 &  1.000 \\
NN-4-layer\_thin\_dropout\_lr01        &                              -5.518 &  0.000 &                             0.225 &  0.822 &                            0.023 &  0.982 &                  -5.112 &  0.000 \\
NN-4-layer\_thin\_dropout\_lr1         &                              -6.435 &  0.000 &                            -0.625 &  0.533 &                           -0.835 &  0.404 &                  -6.024 &  0.000 \\
NN-4-layer\_wide\_no\_dropout          &                               0.963 &  0.337 &                             6.956 &  0.000 &                            6.796 &  0.000 &                   1.390 &  0.166 \\
NN-4-layer\_wide\_no\_dropout\_lr01     &                              -5.595 &  0.000 &                             0.338 &  0.735 &                            0.130 &  0.896 &                  -5.175 &  0.000 \\
NN-4-layer\_wide\_no\_dropout\_lr1      &                              -5.712 &  0.000 &                             0.183 &  0.855 &                           -0.025 &  0.980 &                  -5.296 &  0.000 \\
NN-4-layer\_wide\_with\_dropout        &                               0.905 &  0.367 &                             6.910 &  0.000 &                            6.749 &  0.000 &                   1.333 &  0.184 \\
NN-4-layer\_wide\_with\_dropout\_lr01   &                              -6.142 &  0.000 &                            -0.220 &  0.826 &                           -0.431 &  0.667 &                  -5.724 &  0.000 \\
NN-4-layer\_wide\_with\_dropout\_lr1    &                              -5.914 &  0.000 &                            -0.006 &  0.996 &                           -0.215 &  0.830 &                  -5.496 &  0.000 \\
PassiveAggressiveClassifier         &                               3.400 &  0.001 &                            10.008 &  0.000 &                            9.867 &  0.000 &                   3.874 &  0.000 \\
RandomForestClassifier              &                               6.036 &  0.000 &                            12.797 &  0.000 &                           12.684 &  0.000 &                   6.524 &  0.000 \\
SVC                                 &                               5.296 &  0.000 &                            11.988 &  0.000 &                           11.867 &  0.000 &                   5.777 &  0.000 \\
\bottomrule
\end{tabular}

%% file: tables/t_test5.tex
\begin{tabular}{lrrrrrrrr}
\toprule
{} & \multicolumn{2}{l}{NN-4-layer\_thin\_dropout\_lr01} & \multicolumn{2}{l}{NN-4-layer\_thin\_dropout\_lr1} & \multicolumn{2}{l}{NN-4-layer\_wide\_no\_dropout} & \multicolumn{2}{l}{NN-4-layer\_wide\_no\_dropout\_lr01} \\
{} &                       t\_stat &  p\_val &                      t\_stat &  p\_val &                     t\_stat &  p\_val &                          t\_stat &  p\_val \\
\midrule
BaggingClassifier                   &                       12.112 &  0.000 &                      13.293 &  0.000 &                      4.958 &  0.000 &                          12.554 &  0.000 \\
BaselineClassifier                  &                       -2.299 &  0.022 &                      -1.411 &  0.159 &                     -9.704 &  0.000 &                          -2.500 &  0.013 \\
BernoulliNaiveBayes                 &                        7.839 &  0.000 &                       8.937 &  0.000 &                      0.477 &  0.634 &                           8.067 &  0.000 \\
GaussianNaiveBayes                  &                        6.117 &  0.000 &                       7.100 &  0.000 &                     -0.742 &  0.459 &                           6.234 &  0.000 \\
GradientBoostingClassifier          &                       10.559 &  0.000 &                      11.665 &  0.000 &                      3.581 &  0.000 &                          10.891 &  0.000 \\
K\_Neighbours                        &                       11.693 &  0.000 &                      12.881 &  0.000 &                      4.436 &  0.000 &                          12.125 &  0.000 \\
NN-12-layer\_wide\_with\_dropout       &                        1.474 &  0.142 &                       2.338 &  0.020 &                     -5.141 &  0.000 &                           1.416 &  0.158 \\
NN-12-layer\_wide\_with\_dropout\_lr01  &                       -0.310 &  0.757 &                       0.549 &  0.584 &                     -7.126 &  0.000 &                          -0.427 &  0.670 \\
NN-12-layer\_wide\_with\_dropout\_lr1   &                       -0.950 &  0.343 &                      -0.109 &  0.914 &                     -7.710 &  0.000 &                          -1.087 &  0.278 \\
NN-2-layer-droput-input-layer\_lr001 &                        5.247 &  0.000 &                       6.174 &  0.000 &                     -1.339 &  0.182 &                           5.319 &  0.000 \\
NN-2-layer-droput-input-layer\_lr01  &                        1.640 &  0.102 &                       2.493 &  0.013 &                     -4.864 &  0.000 &                           1.587 &  0.114 \\
NN-2-layer-droput-input-layer\_lr1   &                        0.627 &  0.531 &                       1.479 &  0.141 &                     -5.999 &  0.000 &                           0.542 &  0.588 \\
NN-4-layer-droput-each-layer\_lr0001 &                        5.518 &  0.000 &                       6.435 &  0.000 &                     -0.963 &  0.337 &                           5.595 &  0.000 \\
NN-4-layer-droput-each-layer\_lr01   &                       -0.225 &  0.822 &                       0.625 &  0.533 &                     -6.956 &  0.000 &                          -0.338 &  0.735 \\
NN-4-layer-droput-each-layer\_lr1    &                       -0.023 &  0.982 &                       0.835 &  0.404 &                     -6.796 &  0.000 &                          -0.130 &  0.896 \\
NN-4-layer\_thin\_dropout             &                        5.112 &  0.000 &                       6.024 &  0.000 &                     -1.390 &  0.166 &                           5.175 &  0.000 \\
NN-4-layer\_thin\_dropout\_lr01        &                        0.000 &  1.000 &                       0.835 &  0.405 &                     -6.569 &  0.000 &                          -0.104 &  0.917 \\
NN-4-layer\_thin\_dropout\_lr1         &                       -0.835 &  0.405 &                       0.000 &  1.000 &                     -7.519 &  0.000 &                          -0.966 &  0.335 \\
NN-4-layer\_wide\_no\_dropout          &                        6.569 &  0.000 &                       7.519 &  0.000 &                      0.000 &  1.000 &                           6.690 &  0.000 \\
NN-4-layer\_wide\_no\_dropout\_lr01     &                        0.104 &  0.917 &                       0.966 &  0.335 &                     -6.690 &  0.000 &                           0.000 &  1.000 \\
NN-4-layer\_wide\_no\_dropout\_lr1      &                       -0.047 &  0.963 &                       0.809 &  0.420 &                     -6.801 &  0.000 &                          -0.155 &  0.877 \\
NN-4-layer\_wide\_with\_dropout        &                        6.523 &  0.000 &                       7.474 &  0.000 &                     -0.061 &  0.951 &                           6.642 &  0.000 \\
NN-4-layer\_wide\_with\_dropout\_lr01   &                       -0.442 &  0.659 &                       0.413 &  0.680 &                     -7.246 &  0.000 &                          -0.564 &  0.574 \\
NN-4-layer\_wide\_with\_dropout\_lr1    &                       -0.232 &  0.817 &                       0.623 &  0.534 &                     -7.010 &  0.000 &                          -0.346 &  0.729 \\
PassiveAggressiveClassifier         &                        9.480 &  0.000 &                      10.574 &  0.000 &                      2.402 &  0.017 &                           9.767 &  0.000 \\
RandomForestClassifier              &                       12.165 &  0.000 &                      13.332 &  0.000 &                      5.096 &  0.000 &                          12.598 &  0.000 \\
SVC                                 &                       11.391 &  0.000 &                      12.532 &  0.000 &                      4.342 &  0.000 &                          11.777 &  0.000 \\
\bottomrule
\end{tabular}

%% file: tables/t_test6.tex
\begin{tabular}{lrrrrrrrr}
\toprule
{} & \multicolumn{2}{l}{NN-4-layer\_wide\_no\_dropout\_lr1} & \multicolumn{2}{l}{NN-4-layer\_wide\_with\_dropout} & \multicolumn{2}{l}{NN-4-layer\_wide\_with\_dropout\_lr01} & \multicolumn{2}{l}{NN-4-layer\_wide\_with\_dropout\_lr1} \\
{} &                         t\_stat &  p\_val &                       t\_stat &  p\_val &                            t\_stat &  p\_val &                           t\_stat &  p\_val \\
\midrule
BaggingClassifier                   &                         12.618 &  0.000 &                        5.043 &  0.000 &                            13.170 &  0.000 &                           12.877 &  0.000 \\
BaselineClassifier                  &                         -2.316 &  0.021 &                       -9.661 &  0.000 &                            -1.892 &  0.060 &                           -2.118 &  0.035 \\
BernoulliNaiveBayes                 &                          8.174 &  0.000 &                        0.548 &  0.584 &                             8.690 &  0.000 &                            8.416 &  0.000 \\
GaussianNaiveBayes                  &                          6.351 &  0.000 &                       -0.680 &  0.497 &                             6.813 &  0.000 &                            6.568 &  0.000 \\
GradientBoostingClassifier          &                         10.972 &  0.000 &                        3.658 &  0.000 &                            11.489 &  0.000 &                           11.215 &  0.000 \\
K\_Neighbours                        &                         12.193 &  0.000 &                        4.520 &  0.000 &                            12.748 &  0.000 &                           12.454 &  0.000 \\
NN-12-layer\_wide\_with\_dropout       &                          1.560 &  0.120 &                       -5.091 &  0.000 &                             1.968 &  0.050 &                            1.751 &  0.081 \\
NN-12-layer\_wide\_with\_dropout\_lr01  &                         -0.269 &  0.788 &                       -7.079 &  0.000 &                             0.138 &  0.891 &                           -0.079 &  0.937 \\
NN-12-layer\_wide\_with\_dropout\_lr1   &                         -0.927 &  0.355 &                       -7.664 &  0.000 &                            -0.528 &  0.598 &                           -0.740 &  0.460 \\
NN-2-layer-droput-input-layer\_lr001 &                          5.439 &  0.000 &                       -1.281 &  0.201 &                             5.874 &  0.000 &                            5.643 &  0.000 \\
NN-2-layer-droput-input-layer\_lr01  &                          1.728 &  0.085 &                       -4.815 &  0.000 &                             2.130 &  0.034 &                            1.916 &  0.057 \\
NN-2-layer-droput-input-layer\_lr1   &                          0.691 &  0.491 &                       -5.951 &  0.000 &                             1.093 &  0.275 &                            0.879 &  0.380 \\
NN-4-layer-droput-each-layer\_lr0001 &                          5.712 &  0.000 &                       -0.905 &  0.367 &                             6.142 &  0.000 &                            5.914 &  0.000 \\
NN-4-layer-droput-each-layer\_lr01   &                         -0.183 &  0.855 &                       -6.910 &  0.000 &                             0.220 &  0.826 &                            0.006 &  0.996 \\
NN-4-layer-droput-each-layer\_lr1    &                          0.025 &  0.980 &                       -6.749 &  0.000 &                             0.431 &  0.667 &                            0.215 &  0.830 \\
NN-4-layer\_thin\_dropout             &                          5.296 &  0.000 &                       -1.333 &  0.184 &                             5.724 &  0.000 &                            5.496 &  0.000 \\
NN-4-layer\_thin\_dropout\_lr01        &                          0.047 &  0.963 &                       -6.523 &  0.000 &                             0.442 &  0.659 &                            0.232 &  0.817 \\
NN-4-layer\_thin\_dropout\_lr1         &                         -0.809 &  0.420 &                       -7.474 &  0.000 &                            -0.413 &  0.680 &                           -0.623 &  0.534 \\
NN-4-layer\_wide\_no\_dropout          &                          6.801 &  0.000 &                        0.061 &  0.951 &                             7.246 &  0.000 &                            7.010 &  0.000 \\
NN-4-layer\_wide\_no\_dropout\_lr01     &                          0.155 &  0.877 &                       -6.642 &  0.000 &                             0.564 &  0.574 &                            0.346 &  0.729 \\
NN-4-layer\_wide\_no\_dropout\_lr1      &                          0.000 &  1.000 &                       -6.754 &  0.000 &                             0.405 &  0.686 &                            0.190 &  0.850 \\
NN-4-layer\_wide\_with\_dropout        &                          6.754 &  0.000 &                        0.000 &  1.000 &                             7.199 &  0.000 &                            6.963 &  0.000 \\
NN-4-layer\_wide\_with\_dropout\_lr01   &                         -0.405 &  0.686 &                       -7.199 &  0.000 &                             0.000 &  1.000 &                           -0.216 &  0.829 \\
NN-4-layer\_wide\_with\_dropout\_lr1    &                         -0.190 &  0.850 &                       -6.963 &  0.000 &                             0.216 &  0.829 &                            0.000 &  1.000 \\
PassiveAggressiveClassifier         &                          9.858 &  0.000 &                        2.476 &  0.014 &                            10.371 &  0.000 &                           10.098 &  0.000 \\
RandomForestClassifier              &                         12.663 &  0.000 &                        5.180 &  0.000 &                            13.207 &  0.000 &                           12.918 &  0.000 \\
SVC                                 &                         11.849 &  0.000 &                        4.422 &  0.000 &                            12.382 &  0.000 &                           12.099 &  0.000 \\
\bottomrule
\end{tabular}

%% file: tables/t_test7.tex
\begin{tabular}{lrrrrrr}
\toprule
{} & \multicolumn{2}{l}{PassiveAggressiveClassifier} & \multicolumn{2}{l}{RandomForestClassifier} & \multicolumn{2}{l}{SVC} \\
{} &                      t\_stat &  p\_val &                 t\_stat &  p\_val &  t\_stat &  p\_val \\
\midrule
BaggingClassifier                   &                       2.876 &  0.004 &                 -0.253 &  0.800 &   0.607 &  0.544 \\
BaselineClassifier                  &                     -13.497 &  0.000 &                -16.752 &  0.000 & -15.781 &  0.000 \\
BernoulliNaiveBayes                 &                      -2.313 &  0.022 &                 -5.597 &  0.000 &  -4.660 &  0.000 \\
GaussianNaiveBayes                  &                      -3.371 &  0.001 &                 -6.262 &  0.000 &  -5.443 &  0.000 \\
GradientBoostingClassifier          &                       1.370 &  0.172 &                 -1.618 &  0.107 &  -0.791 &  0.430 \\
K\_Neighbours                        &                       2.238 &  0.026 &                 -0.976 &  0.330 &  -0.086 &  0.931 \\
NN-12-layer\_wide\_with\_dropout       &                      -7.982 &  0.000 &                -10.700 &  0.000 &  -9.919 &  0.000 \\
NN-12-layer\_wide\_with\_dropout\_lr01  &                     -10.251 &  0.000 &                -13.096 &  0.000 & -12.269 &  0.000 \\
NN-12-layer\_wide\_with\_dropout\_lr1   &                     -10.834 &  0.000 &                -13.640 &  0.000 & -12.823 &  0.000 \\
NN-2-layer-droput-input-layer\_lr001 &                      -3.867 &  0.000 &                 -6.571 &  0.000 &  -5.808 &  0.000 \\
NN-2-layer-droput-input-layer\_lr01  &                      -7.614 &  0.000 &                -10.261 &  0.000 &  -9.503 &  0.000 \\
NN-2-layer-droput-input-layer\_lr1   &                      -8.909 &  0.000 &                -11.632 &  0.000 & -10.848 &  0.000 \\
NN-4-layer-droput-each-layer\_lr0001 &                      -3.400 &  0.001 &                 -6.036 &  0.000 &  -5.296 &  0.000 \\
NN-4-layer-droput-each-layer\_lr01   &                     -10.008 &  0.000 &                -12.797 &  0.000 & -11.988 &  0.000 \\
NN-4-layer-droput-each-layer\_lr1    &                      -9.867 &  0.000 &                -12.684 &  0.000 & -11.867 &  0.000 \\
NN-4-layer\_thin\_dropout             &                      -3.874 &  0.000 &                 -6.524 &  0.000 &  -5.777 &  0.000 \\
NN-4-layer\_thin\_dropout\_lr01        &                      -9.480 &  0.000 &                -12.165 &  0.000 & -11.391 &  0.000 \\
NN-4-layer\_thin\_dropout\_lr1         &                     -10.574 &  0.000 &                -13.332 &  0.000 & -12.532 &  0.000 \\
NN-4-layer\_wide\_no\_dropout          &                      -2.402 &  0.017 &                 -5.096 &  0.000 &  -4.342 &  0.000 \\
NN-4-layer\_wide\_no\_dropout\_lr01     &                      -9.767 &  0.000 &                -12.598 &  0.000 & -11.777 &  0.000 \\
NN-4-layer\_wide\_no\_dropout\_lr1      &                      -9.858 &  0.000 &                -12.663 &  0.000 & -11.849 &  0.000 \\
NN-4-layer\_wide\_with\_dropout        &                      -2.476 &  0.014 &                 -5.180 &  0.000 &  -4.422 &  0.000 \\
NN-4-layer\_wide\_with\_dropout\_lr01   &                     -10.371 &  0.000 &                -13.207 &  0.000 & -12.382 &  0.000 \\
NN-4-layer\_wide\_with\_dropout\_lr1    &                     -10.098 &  0.000 &                -12.918 &  0.000 & -12.099 &  0.000 \\
PassiveAggressiveClassifier         &                       0.000 &  1.000 &                 -3.062 &  0.002 &  -2.205 &  0.028 \\
RandomForestClassifier              &                       3.062 &  0.002 &                  0.000 &  1.000 &   0.839 &  0.402 \\
SVC                                 &                       2.205 &  0.028 &                 -0.839 &  0.402 &   0.000 &  1.000 \\
\bottomrule
\end{tabular}

%% file: tables/wilcoxon_test1.tex
\begin{tabular}{lrrrrrrrr}
\toprule
{} & \multicolumn{2}{l}{BaggingClassifier} & \multicolumn{2}{l}{BaselineClassifier} & \multicolumn{2}{l}{BernoulliNaiveBayes} & \multicolumn{2}{l}{GaussianNaiveBayes} \\
{} &         statistic &  p\_val &          statistic & p\_val &           statistic &  p\_val &          statistic &  p\_val \\
\midrule
BaggingClassifier                   &               0.0 &    NaN &                6.0 &   0.0 &               443.5 &  0.000 &              411.5 &  0.000 \\
BaselineClassifier                  &               6.0 &  0.000 &                0.0 &   NaN &                49.0 &  0.000 &              400.0 &  0.000 \\
BernoulliNaiveBayes                 &             443.5 &  0.000 &               49.0 &   0.0 &                 0.0 &    NaN &             2228.0 &  0.056 \\
GaussianNaiveBayes                  &             411.5 &  0.000 &              400.0 &   0.0 &              2228.0 &  0.056 &                0.0 &    NaN \\
GradientBoostingClassifier          &             984.0 &  0.000 &               23.5 &   0.0 &              1073.5 &  0.000 &             1053.0 &  0.000 \\
K\_Neighbours                        &            1929.0 &  0.020 &                1.0 &   0.0 &               306.0 &  0.000 &              344.0 &  0.000 \\
NN-12-layer\_wide\_with\_dropout       &             218.0 &  0.000 &              574.0 &   0.0 &               822.5 &  0.000 &             1587.0 &  0.000 \\
NN-12-layer\_wide\_with\_dropout\_lr01  &             148.0 &  0.000 &             1061.0 &   0.0 &               362.0 &  0.000 &              984.0 &  0.000 \\
NN-12-layer\_wide\_with\_dropout\_lr1   &             119.0 &  0.000 &             1393.0 &   0.0 &               331.0 &  0.000 &              755.0 &  0.000 \\
NN-2-layer-droput-input-layer\_lr001 &             475.0 &  0.000 &               52.0 &   0.0 &              2260.5 &  0.036 &             2944.5 &  0.873 \\
NN-2-layer-droput-input-layer\_lr01  &             142.0 &  0.000 &              239.0 &   0.0 &               410.5 &  0.000 &             1259.0 &  0.000 \\
NN-2-layer-droput-input-layer\_lr1   &             174.0 &  0.000 &              649.0 &   0.0 &               417.5 &  0.000 &             1191.0 &  0.000 \\
NN-4-layer-droput-each-layer\_lr0001 &             633.0 &  0.000 &               53.0 &   0.0 &              2665.5 &  0.316 &             2394.0 &  0.214 \\
NN-4-layer-droput-each-layer\_lr01   &             122.0 &  0.000 &             1023.0 &   0.0 &               349.0 &  0.000 &              992.0 &  0.000 \\
NN-4-layer-droput-each-layer\_lr1    &             139.0 &  0.000 &              732.0 &   0.0 &               363.0 &  0.000 &             1034.0 &  0.000 \\
NN-4-layer\_thin\_dropout             &             507.0 &  0.000 &               60.0 &   0.0 &              2357.0 &  0.053 &             2788.5 &  0.755 \\
NN-4-layer\_thin\_dropout\_lr01        &             118.0 &  0.000 &             1118.0 &   0.0 &               338.5 &  0.000 &              944.5 &  0.000 \\
NN-4-layer\_thin\_dropout\_lr1         &             103.0 &  0.000 &             1555.0 &   0.0 &               325.0 &  0.000 &              947.0 &  0.000 \\
NN-4-layer\_wide\_no\_dropout          &             702.5 &  0.000 &               27.0 &   0.0 &              2550.5 &  0.369 &             1891.5 &  0.007 \\
NN-4-layer\_wide\_no\_dropout\_lr01     &             106.0 &  0.000 &              816.0 &   0.0 &               337.0 &  0.000 &             1002.0 &  0.000 \\
NN-4-layer\_wide\_no\_dropout\_lr1      &             112.0 &  0.000 &              899.0 &   0.0 &               348.0 &  0.000 &             1007.0 &  0.000 \\
NN-4-layer\_wide\_with\_dropout        &             698.5 &  0.000 &               40.0 &   0.0 &              2860.0 &  0.466 &             2144.0 &  0.014 \\
NN-4-layer\_wide\_with\_dropout\_lr01   &             123.0 &  0.000 &             1212.5 &   0.0 &               333.0 &  0.000 &              984.0 &  0.000 \\
NN-4-layer\_wide\_with\_dropout\_lr1    &             114.0 &  0.000 &              999.5 &   0.0 &               302.0 &  0.000 &              988.0 &  0.000 \\
PassiveAggressiveClassifier         &             980.0 &  0.000 &               11.5 &   0.0 &              1330.0 &  0.000 &              958.5 &  0.000 \\
RandomForestClassifier              &            1459.0 &  0.005 &                0.0 &   0.0 &               259.5 &  0.000 &              252.5 &  0.000 \\
SVC                                 &            2571.0 &  0.606 &                0.0 &   0.0 &               387.5 &  0.000 &              249.0 &  0.000 \\
\bottomrule
\end{tabular}

%% file: tables/wilcoxon_test2.tex
\begin{tabular}{lrrrrrrrr}
\toprule
{} & \multicolumn{2}{l}{GradientBoostingClassifier} & \multicolumn{2}{l}{K\_Neighbours} & \multicolumn{2}{l}{NN-12-layer\_wide\_with\_dropout} & \multicolumn{2}{l}{NN-12-layer\_wide\_with\_dropout\_lr01} \\
{} &                  statistic &  p\_val &    statistic &  p\_val &                     statistic &  p\_val &                          statistic &  p\_val \\
\midrule
BaggingClassifier                   &                      984.0 &  0.000 &       1929.0 &  0.020 &                         218.0 &  0.000 &                              148.0 &  0.000 \\
BaselineClassifier                  &                       23.5 &  0.000 &          1.0 &  0.000 &                         574.0 &  0.000 &                             1061.0 &  0.000 \\
BernoulliNaiveBayes                 &                     1073.5 &  0.000 &        306.0 &  0.000 &                         822.5 &  0.000 &                              362.0 &  0.000 \\
GaussianNaiveBayes                  &                     1053.0 &  0.000 &        344.0 &  0.000 &                        1587.0 &  0.000 &                              984.0 &  0.000 \\
GradientBoostingClassifier          &                        0.0 &    NaN &       2202.5 &  0.118 &                         320.0 &  0.000 &                              241.0 &  0.000 \\
K\_Neighbours                        &                     2202.5 &  0.118 &          0.0 &    NaN &                         121.0 &  0.000 &                               51.0 &  0.000 \\
NN-12-layer\_wide\_with\_dropout       &                      320.0 &  0.000 &        121.0 &  0.000 &                           0.0 &    NaN &                              274.0 &  0.000 \\
NN-12-layer\_wide\_with\_dropout\_lr01  &                      241.0 &  0.000 &         51.0 &  0.000 &                         274.0 &  0.000 &                                0.0 &    NaN \\
NN-12-layer\_wide\_with\_dropout\_lr1   &                      221.0 &  0.000 &         23.0 &  0.000 &                         214.0 &  0.000 &                              541.0 &  0.808 \\
NN-2-layer-droput-input-layer\_lr001 &                      846.5 &  0.000 &        230.0 &  0.000 &                         744.0 &  0.000 &                              301.0 &  0.000 \\
NN-2-layer-droput-input-layer\_lr01  &                      312.0 &  0.000 &         35.0 &  0.000 &                        1269.5 &  0.656 &                              463.5 &  0.000 \\
NN-2-layer-droput-input-layer\_lr1   &                      285.0 &  0.000 &         66.0 &  0.000 &                         681.0 &  0.025 &                              451.0 &  0.005 \\
NN-4-layer-droput-each-layer\_lr0001 &                      946.5 &  0.000 &        380.0 &  0.000 &                         330.0 &  0.000 &                               87.0 &  0.000 \\
NN-4-layer-droput-each-layer\_lr01   &                      229.0 &  0.000 &         24.0 &  0.000 &                         269.5 &  0.000 &                              398.0 &  0.872 \\
NN-4-layer-droput-each-layer\_lr1    &                      230.0 &  0.000 &         39.0 &  0.000 &                         294.0 &  0.000 &                              232.0 &  0.386 \\
NN-4-layer\_thin\_dropout             &                      756.0 &  0.000 &        361.0 &  0.000 &                         637.0 &  0.000 &                              310.0 &  0.000 \\
NN-4-layer\_thin\_dropout\_lr01        &                      205.0 &  0.000 &         24.0 &  0.000 &                         415.0 &  0.001 &                              657.0 &  0.250 \\
NN-4-layer\_thin\_dropout\_lr1         &                      222.0 &  0.000 &         25.0 &  0.000 &                         194.5 &  0.000 &                              394.0 &  0.239 \\
NN-4-layer\_wide\_no\_dropout          &                     1107.5 &  0.000 &        562.0 &  0.000 &                         380.5 &  0.000 &                              172.5 &  0.000 \\
NN-4-layer\_wide\_no\_dropout\_lr01     &                      222.0 &  0.000 &         23.0 &  0.000 &                         370.0 &  0.001 &                              353.5 &  0.149 \\
NN-4-layer\_wide\_no\_dropout\_lr1      &                      220.0 &  0.000 &         26.0 &  0.000 &                         420.5 &  0.000 &                              326.0 &  0.372 \\
NN-4-layer\_wide\_with\_dropout        &                     1109.0 &  0.000 &        529.0 &  0.000 &                         361.5 &  0.000 &                              169.0 &  0.000 \\
NN-4-layer\_wide\_with\_dropout\_lr01   &                      211.0 &  0.000 &         41.0 &  0.000 &                         294.5 &  0.000 &                              389.0 &  0.778 \\
NN-4-layer\_wide\_with\_dropout\_lr1    &                      223.0 &  0.000 &         26.0 &  0.000 &                         290.0 &  0.000 &                              375.0 &  0.472 \\
PassiveAggressiveClassifier         &                     2379.5 &  0.062 &       1178.0 &  0.000 &                         435.0 &  0.000 &                              229.0 &  0.000 \\
RandomForestClassifier              &                      732.0 &  0.000 &       1187.0 &  0.000 &                         109.0 &  0.000 &                               82.0 &  0.000 \\
SVC                                 &                     1797.5 &  0.001 &       1867.5 &  0.067 &                          58.0 &  0.000 &                               39.0 &  0.000 \\
\bottomrule
\end{tabular}

%% file: tables/wilcoxon_test3.tex
\begin{tabular}{lrrrrrrrr}
\toprule
{} & \multicolumn{2}{l}{NN-12-layer\_wide\_with\_dropout\_lr1} & \multicolumn{2}{l}{NN-2-layer-droput-input-layer\_lr001} & \multicolumn{2}{l}{NN-2-layer-droput-input-layer\_lr01} & \multicolumn{2}{l}{NN-2-layer-droput-input-layer\_lr1} \\
{} &                         statistic &  p\_val &                           statistic &  p\_val &                          statistic &  p\_val &                         statistic &  p\_val \\
\midrule
BaggingClassifier                   &                             119.0 &  0.000 &                               475.0 &  0.000 &                              142.0 &  0.000 &                             174.0 &  0.000 \\
BaselineClassifier                  &                            1393.0 &  0.000 &                                52.0 &  0.000 &                              239.0 &  0.000 &                             649.0 &  0.000 \\
BernoulliNaiveBayes                 &                             331.0 &  0.000 &                              2260.5 &  0.036 &                              410.5 &  0.000 &                             417.5 &  0.000 \\
GaussianNaiveBayes                  &                             755.0 &  0.000 &                              2944.5 &  0.873 &                             1259.0 &  0.000 &                            1191.0 &  0.000 \\
GradientBoostingClassifier          &                             221.0 &  0.000 &                               846.5 &  0.000 &                              312.0 &  0.000 &                             285.0 &  0.000 \\
K\_Neighbours                        &                              23.0 &  0.000 &                               230.0 &  0.000 &                               35.0 &  0.000 &                              66.0 &  0.000 \\
NN-12-layer\_wide\_with\_dropout       &                             214.0 &  0.000 &                               744.0 &  0.000 &                             1269.5 &  0.656 &                             681.0 &  0.025 \\
NN-12-layer\_wide\_with\_dropout\_lr01  &                             541.0 &  0.808 &                               301.0 &  0.000 &                              463.5 &  0.000 &                             451.0 &  0.005 \\
NN-12-layer\_wide\_with\_dropout\_lr1   &                               0.0 &    NaN &                               203.0 &  0.000 &                              399.0 &  0.000 &                             468.0 &  0.002 \\
NN-2-layer-droput-input-layer\_lr001 &                             203.0 &  0.000 &                                 0.0 &    NaN &                              442.0 &  0.000 &                             365.0 &  0.000 \\
NN-2-layer-droput-input-layer\_lr01  &                             399.0 &  0.000 &                               442.0 &  0.000 &                                0.0 &    NaN &                             703.0 &  0.010 \\
NN-2-layer-droput-input-layer\_lr1   &                             468.0 &  0.002 &                               365.0 &  0.000 &                              703.0 &  0.010 &                               0.0 &    NaN \\
NN-4-layer-droput-each-layer\_lr0001 &                              91.0 &  0.000 &                              1645.5 &  0.106 &                              170.0 &  0.000 &                             138.0 &  0.000 \\
NN-4-layer-droput-each-layer\_lr01   &                             536.0 &  0.767 &                               238.0 &  0.000 &                              360.0 &  0.000 &                             436.0 &  0.003 \\
NN-4-layer-droput-each-layer\_lr1    &                             332.0 &  0.202 &                               364.0 &  0.000 &                              522.5 &  0.000 &                             485.0 &  0.017 \\
NN-4-layer\_thin\_dropout             &                             310.0 &  0.000 &                              2456.5 &  0.814 &                              517.0 &  0.000 &                             434.0 &  0.000 \\
NN-4-layer\_thin\_dropout\_lr01        &                             545.5 &  0.060 &                               382.5 &  0.000 &                              626.5 &  0.001 &                             739.0 &  0.138 \\
NN-4-layer\_thin\_dropout\_lr1         &                             489.5 &  0.430 &                               227.0 &  0.000 &                              383.0 &  0.000 &                             478.5 &  0.001 \\
NN-4-layer\_wide\_no\_dropout          &                             165.0 &  0.000 &                              1309.0 &  0.000 &                              160.0 &  0.000 &                             206.5 &  0.000 \\
NN-4-layer\_wide\_no\_dropout\_lr01     &                             331.0 &  0.056 &                               253.0 &  0.000 &                              575.5 &  0.001 &                             675.5 &  0.163 \\
NN-4-layer\_wide\_no\_dropout\_lr1      &                             403.5 &  0.198 &                               289.0 &  0.000 &                              544.0 &  0.001 &                             462.5 &  0.025 \\
NN-4-layer\_wide\_with\_dropout        &                             189.0 &  0.000 &                              1468.5 &  0.000 &                              238.0 &  0.000 &                             223.5 &  0.000 \\
NN-4-layer\_wide\_with\_dropout\_lr01   &                             490.0 &  0.756 &                               264.0 &  0.000 &                              424.0 &  0.000 &                             397.0 &  0.002 \\
NN-4-layer\_wide\_with\_dropout\_lr1    &                             506.0 &  0.289 &                               225.0 &  0.000 &                              451.5 &  0.000 &                             610.0 &  0.010 \\
PassiveAggressiveClassifier         &                             173.0 &  0.000 &                               776.0 &  0.000 &                              200.0 &  0.000 &                             288.0 &  0.000 \\
RandomForestClassifier              &                              61.0 &  0.000 &                               210.0 &  0.000 &                               73.0 &  0.000 &                             104.0 &  0.000 \\
SVC                                 &                              19.0 &  0.000 &                               254.0 &  0.000 &                               40.0 &  0.000 &                              68.0 &  0.000 \\
\bottomrule
\end{tabular}

%% file: tables/wilcoxon_test4.tex
\begin{tabular}{lrrrrrrrr}
\toprule
{} & \multicolumn{2}{l}{NN-4-layer-droput-each-layer\_lr0001} & \multicolumn{2}{l}{NN-4-layer-droput-each-layer\_lr01} & \multicolumn{2}{l}{NN-4-layer-droput-each-layer\_lr1} & \multicolumn{2}{l}{NN-4-layer\_thin\_dropout} \\
{} &                           statistic &  p\_val &                         statistic &  p\_val &                        statistic &  p\_val &               statistic &  p\_val \\
\midrule
BaggingClassifier                   &                               633.0 &  0.000 &                             122.0 &  0.000 &                            139.0 &  0.000 &                   507.0 &  0.000 \\
BaselineClassifier                  &                                53.0 &  0.000 &                            1023.0 &  0.000 &                            732.0 &  0.000 &                    60.0 &  0.000 \\
BernoulliNaiveBayes                 &                              2665.5 &  0.316 &                             349.0 &  0.000 &                            363.0 &  0.000 &                  2357.0 &  0.053 \\
GaussianNaiveBayes                  &                              2394.0 &  0.214 &                             992.0 &  0.000 &                           1034.0 &  0.000 &                  2788.5 &  0.755 \\
GradientBoostingClassifier          &                               946.5 &  0.000 &                             229.0 &  0.000 &                            230.0 &  0.000 &                   756.0 &  0.000 \\
K\_Neighbours                        &                               380.0 &  0.000 &                              24.0 &  0.000 &                             39.0 &  0.000 &                   361.0 &  0.000 \\
NN-12-layer\_wide\_with\_dropout       &                               330.0 &  0.000 &                             269.5 &  0.000 &                            294.0 &  0.000 &                   637.0 &  0.000 \\
NN-12-layer\_wide\_with\_dropout\_lr01  &                                87.0 &  0.000 &                             398.0 &  0.872 &                            232.0 &  0.386 &                   310.0 &  0.000 \\
NN-12-layer\_wide\_with\_dropout\_lr1   &                                91.0 &  0.000 &                             536.0 &  0.767 &                            332.0 &  0.202 &                   310.0 &  0.000 \\
NN-2-layer-droput-input-layer\_lr001 &                              1645.5 &  0.106 &                             238.0 &  0.000 &                            364.0 &  0.000 &                  2456.5 &  0.814 \\
NN-2-layer-droput-input-layer\_lr01  &                               170.0 &  0.000 &                             360.0 &  0.000 &                            522.5 &  0.000 &                   517.0 &  0.000 \\
NN-2-layer-droput-input-layer\_lr1   &                               138.0 &  0.000 &                             436.0 &  0.003 &                            485.0 &  0.017 &                   434.0 &  0.000 \\
NN-4-layer-droput-each-layer\_lr0001 &                                 0.0 &    NaN &                              89.0 &  0.000 &                            135.0 &  0.000 &                  1816.0 &  0.085 \\
NN-4-layer-droput-each-layer\_lr01   &                                89.0 &  0.000 &                               0.0 &    NaN &                            411.0 &  0.613 &                   316.5 &  0.000 \\
NN-4-layer-droput-each-layer\_lr1    &                               135.0 &  0.000 &                             411.0 &  0.613 &                              0.0 &    NaN &                   366.0 &  0.000 \\
NN-4-layer\_thin\_dropout             &                              1816.0 &  0.085 &                             316.5 &  0.000 &                            366.0 &  0.000 &                     0.0 &    NaN \\
NN-4-layer\_thin\_dropout\_lr01        &                               200.0 &  0.000 &                             510.0 &  0.218 &                            703.0 &  0.734 &                   425.5 &  0.000 \\
NN-4-layer\_thin\_dropout\_lr1         &                                96.0 &  0.000 &                             338.0 &  0.103 &                            250.0 &  0.019 &                   301.5 &  0.000 \\
NN-4-layer\_wide\_no\_dropout          &                              1216.5 &  0.000 &                             152.0 &  0.000 &                            183.0 &  0.000 &                  1345.5 &  0.000 \\
NN-4-layer\_wide\_no\_dropout\_lr01     &                                78.0 &  0.000 &                             305.0 &  0.158 &                            395.0 &  0.346 &                   320.0 &  0.000 \\
NN-4-layer\_wide\_no\_dropout\_lr1      &                               130.0 &  0.000 &                             331.0 &  0.567 &                            379.5 &  0.682 &                   362.5 &  0.000 \\
NN-4-layer\_wide\_with\_dropout        &                              1222.0 &  0.000 &                             180.0 &  0.000 &                            170.0 &  0.000 &                  1367.0 &  0.000 \\
NN-4-layer\_wide\_with\_dropout\_lr01   &                               103.0 &  0.000 &                             306.5 &  0.497 &                            305.0 &  0.236 &                   298.0 &  0.000 \\
NN-4-layer\_wide\_with\_dropout\_lr1    &                                77.0 &  0.000 &                             331.0 &  0.757 &                            358.0 &  0.856 &                   237.0 &  0.000 \\
PassiveAggressiveClassifier         &                              1205.0 &  0.000 &                             207.0 &  0.000 &                            208.0 &  0.000 &                  1111.0 &  0.000 \\
RandomForestClassifier              &                               296.5 &  0.000 &                              64.0 &  0.000 &                             73.0 &  0.000 &                   198.0 &  0.000 \\
SVC                                 &                               321.5 &  0.000 &                              23.0 &  0.000 &                             29.0 &  0.000 &                   197.0 &  0.000 \\
\bottomrule
\end{tabular}

%% file: tables/wilcoxon_test5.tex
\begin{tabular}{lrrrrrrrr}
\toprule
{} & \multicolumn{2}{l}{NN-4-layer\_thin\_dropout\_lr01} & \multicolumn{2}{l}{NN-4-layer\_thin\_dropout\_lr1} & \multicolumn{2}{l}{NN-4-layer\_wide\_no\_dropout} & \multicolumn{2}{l}{NN-4-layer\_wide\_no\_dropout\_lr01} \\
{} &                    statistic &  p\_val &                   statistic &  p\_val &                  statistic &  p\_val &                       statistic &  p\_val \\
\midrule
BaggingClassifier                   &                        118.0 &  0.000 &                       103.0 &  0.000 &                      702.5 &  0.000 &                           106.0 &  0.000 \\
BaselineClassifier                  &                       1118.0 &  0.000 &                      1555.0 &  0.000 &                       27.0 &  0.000 &                           816.0 &  0.000 \\
BernoulliNaiveBayes                 &                        338.5 &  0.000 &                       325.0 &  0.000 &                     2550.5 &  0.369 &                           337.0 &  0.000 \\
GaussianNaiveBayes                  &                        944.5 &  0.000 &                       947.0 &  0.000 &                     1891.5 &  0.007 &                          1002.0 &  0.000 \\
GradientBoostingClassifier          &                        205.0 &  0.000 &                       222.0 &  0.000 &                     1107.5 &  0.000 &                           222.0 &  0.000 \\
K\_Neighbours                        &                         24.0 &  0.000 &                        25.0 &  0.000 &                      562.0 &  0.000 &                            23.0 &  0.000 \\
NN-12-layer\_wide\_with\_dropout       &                        415.0 &  0.001 &                       194.5 &  0.000 &                      380.5 &  0.000 &                           370.0 &  0.001 \\
NN-12-layer\_wide\_with\_dropout\_lr01  &                        657.0 &  0.250 &                       394.0 &  0.239 &                      172.5 &  0.000 &                           353.5 &  0.149 \\
NN-12-layer\_wide\_with\_dropout\_lr1   &                        545.5 &  0.060 &                       489.5 &  0.430 &                      165.0 &  0.000 &                           331.0 &  0.056 \\
NN-2-layer-droput-input-layer\_lr001 &                        382.5 &  0.000 &                       227.0 &  0.000 &                     1309.0 &  0.000 &                           253.0 &  0.000 \\
NN-2-layer-droput-input-layer\_lr01  &                        626.5 &  0.001 &                       383.0 &  0.000 &                      160.0 &  0.000 &                           575.5 &  0.001 \\
NN-2-layer-droput-input-layer\_lr1   &                        739.0 &  0.138 &                       478.5 &  0.001 &                      206.5 &  0.000 &                           675.5 &  0.163 \\
NN-4-layer-droput-each-layer\_lr0001 &                        200.0 &  0.000 &                        96.0 &  0.000 &                     1216.5 &  0.000 &                            78.0 &  0.000 \\
NN-4-layer-droput-each-layer\_lr01   &                        510.0 &  0.218 &                       338.0 &  0.103 &                      152.0 &  0.000 &                           305.0 &  0.158 \\
NN-4-layer-droput-each-layer\_lr1    &                        703.0 &  0.734 &                       250.0 &  0.019 &                      183.0 &  0.000 &                           395.0 &  0.346 \\
NN-4-layer\_thin\_dropout             &                        425.5 &  0.000 &                       301.5 &  0.000 &                     1345.5 &  0.000 &                           320.0 &  0.000 \\
NN-4-layer\_thin\_dropout\_lr01        &                          0.0 &    NaN &                       468.0 &  0.028 &                      193.0 &  0.000 &                           824.5 &  0.987 \\
NN-4-layer\_thin\_dropout\_lr1         &                        468.0 &  0.028 &                         0.0 &    NaN &                      129.0 &  0.000 &                           226.0 &  0.005 \\
NN-4-layer\_wide\_no\_dropout          &                        193.0 &  0.000 &                       129.0 &  0.000 &                        0.0 &    NaN &                           154.0 &  0.000 \\
NN-4-layer\_wide\_no\_dropout\_lr01     &                        824.5 &  0.987 &                       226.0 &  0.005 &                      154.0 &  0.000 &                             0.0 &    NaN \\
NN-4-layer\_wide\_no\_dropout\_lr1      &                        658.0 &  0.611 &                       253.0 &  0.056 &                      154.0 &  0.000 &                           442.5 &  0.540 \\
NN-4-layer\_wide\_with\_dropout        &                        221.5 &  0.000 &                       115.0 &  0.000 &                     2055.0 &  0.503 &                           141.0 &  0.000 \\
NN-4-layer\_wide\_with\_dropout\_lr01   &                        559.0 &  0.166 &                       464.0 &  0.403 &                      165.0 &  0.000 &                           366.0 &  0.132 \\
NN-4-layer\_wide\_with\_dropout\_lr1    &                        645.5 &  0.297 &                       526.0 &  0.093 &                      135.0 &  0.000 &                           530.0 &  0.552 \\
PassiveAggressiveClassifier         &                        191.0 &  0.000 &                       160.0 &  0.000 &                     1667.0 &  0.000 &                           170.0 &  0.000 \\
RandomForestClassifier              &                         72.0 &  0.000 &                        59.0 &  0.000 &                      359.0 &  0.000 &                            59.0 &  0.000 \\
SVC                                 &                         24.0 &  0.000 &                        23.0 &  0.000 &                      422.0 &  0.000 &                            22.0 &  0.000 \\
\bottomrule
\end{tabular}

%% file: tables/wilcoxon_test6.tex
\begin{tabular}{lrrrrrrrr}
\toprule
{} & \multicolumn{2}{l}{NN-4-layer\_wide\_no\_dropout\_lr1} & \multicolumn{2}{l}{NN-4-layer\_wide\_with\_dropout} & \multicolumn{2}{l}{NN-4-layer\_wide\_with\_dropout\_lr01} & \multicolumn{2}{l}{NN-4-layer\_wide\_with\_dropout\_lr1} \\
{} &                      statistic &  p\_val &                    statistic &  p\_val &                         statistic &  p\_val &                        statistic &  p\_val \\
\midrule
BaggingClassifier                   &                          112.0 &  0.000 &                        698.5 &  0.000 &                             123.0 &  0.000 &                            114.0 &  0.000 \\
BaselineClassifier                  &                          899.0 &  0.000 &                         40.0 &  0.000 &                            1212.5 &  0.000 &                            999.5 &  0.000 \\
BernoulliNaiveBayes                 &                          348.0 &  0.000 &                       2860.0 &  0.466 &                             333.0 &  0.000 &                            302.0 &  0.000 \\
GaussianNaiveBayes                  &                         1007.0 &  0.000 &                       2144.0 &  0.014 &                             984.0 &  0.000 &                            988.0 &  0.000 \\
GradientBoostingClassifier          &                          220.0 &  0.000 &                       1109.0 &  0.000 &                             211.0 &  0.000 &                            223.0 &  0.000 \\
K\_Neighbours                        &                           26.0 &  0.000 &                        529.0 &  0.000 &                              41.0 &  0.000 &                             26.0 &  0.000 \\
NN-12-layer\_wide\_with\_dropout       &                          420.5 &  0.000 &                        361.5 &  0.000 &                             294.5 &  0.000 &                            290.0 &  0.000 \\
NN-12-layer\_wide\_with\_dropout\_lr01  &                          326.0 &  0.372 &                        169.0 &  0.000 &                             389.0 &  0.778 &                            375.0 &  0.472 \\
NN-12-layer\_wide\_with\_dropout\_lr1   &                          403.5 &  0.198 &                        189.0 &  0.000 &                             490.0 &  0.756 &                            506.0 &  0.289 \\
NN-2-layer-droput-input-layer\_lr001 &                          289.0 &  0.000 &                       1468.5 &  0.000 &                             264.0 &  0.000 &                            225.0 &  0.000 \\
NN-2-layer-droput-input-layer\_lr01  &                          544.0 &  0.001 &                        238.0 &  0.000 &                             424.0 &  0.000 &                            451.5 &  0.000 \\
NN-2-layer-droput-input-layer\_lr1   &                          462.5 &  0.025 &                        223.5 &  0.000 &                             397.0 &  0.002 &                            610.0 &  0.010 \\
NN-4-layer-droput-each-layer\_lr0001 &                          130.0 &  0.000 &                       1222.0 &  0.000 &                             103.0 &  0.000 &                             77.0 &  0.000 \\
NN-4-layer-droput-each-layer\_lr01   &                          331.0 &  0.567 &                        180.0 &  0.000 &                             306.5 &  0.497 &                            331.0 &  0.757 \\
NN-4-layer-droput-each-layer\_lr1    &                          379.5 &  0.682 &                        170.0 &  0.000 &                             305.0 &  0.236 &                            358.0 &  0.856 \\
NN-4-layer\_thin\_dropout             &                          362.5 &  0.000 &                       1367.0 &  0.000 &                             298.0 &  0.000 &                            237.0 &  0.000 \\
NN-4-layer\_thin\_dropout\_lr01        &                          658.0 &  0.611 &                        221.5 &  0.000 &                             559.0 &  0.166 &                            645.5 &  0.297 \\
NN-4-layer\_thin\_dropout\_lr1         &                          253.0 &  0.056 &                        115.0 &  0.000 &                             464.0 &  0.403 &                            526.0 &  0.093 \\
NN-4-layer\_wide\_no\_dropout          &                          154.0 &  0.000 &                       2055.0 &  0.503 &                             165.0 &  0.000 &                            135.0 &  0.000 \\
NN-4-layer\_wide\_no\_dropout\_lr01     &                          442.5 &  0.540 &                        141.0 &  0.000 &                             366.0 &  0.132 &                            530.0 &  0.552 \\
NN-4-layer\_wide\_no\_dropout\_lr1      &                            0.0 &    NaN &                        162.0 &  0.000 &                             449.0 &  0.439 &                            466.0 &  0.933 \\
NN-4-layer\_wide\_with\_dropout        &                          162.0 &  0.000 &                          0.0 &    NaN &                             203.0 &  0.000 &                            178.0 &  0.000 \\
NN-4-layer\_wide\_with\_dropout\_lr01   &                          449.0 &  0.439 &                        203.0 &  0.000 &                               0.0 &    NaN &                            438.5 &  0.510 \\
NN-4-layer\_wide\_with\_dropout\_lr1    &                          466.0 &  0.933 &                        178.0 &  0.000 &                             438.5 &  0.510 &                              0.0 &    NaN \\
PassiveAggressiveClassifier         &                          174.0 &  0.000 &                       1705.5 &  0.000 &                             198.0 &  0.000 &                            164.0 &  0.000 \\
RandomForestClassifier              &                           64.0 &  0.000 &                        412.0 &  0.000 &                              77.0 &  0.000 &                             69.0 &  0.000 \\
SVC                                 &                           23.0 &  0.000 &                        501.5 &  0.000 &                              28.0 &  0.000 &                             23.0 &  0.000 \\
\bottomrule
\end{tabular}

%% file: tables/wilcoxon_test7.tex
\begin{tabular}{lrrrrrr}
\toprule
{} & \multicolumn{2}{l}{PassiveAggressiveClassifier} & \multicolumn{2}{l}{RandomForestClassifier} & \multicolumn{2}{l}{SVC} \\
{} &                   statistic &  p\_val &              statistic &  p\_val & statistic &  p\_val \\
\midrule
BaggingClassifier                   &                       980.0 &  0.000 &                 1459.0 &  0.005 &    2571.0 &  0.606 \\
BaselineClassifier                  &                        11.5 &  0.000 &                    0.0 &  0.000 &       0.0 &  0.000 \\
BernoulliNaiveBayes                 &                      1330.0 &  0.000 &                  259.5 &  0.000 &     387.5 &  0.000 \\
GaussianNaiveBayes                  &                       958.5 &  0.000 &                  252.5 &  0.000 &     249.0 &  0.000 \\
GradientBoostingClassifier          &                      2379.5 &  0.062 &                  732.0 &  0.000 &    1797.5 &  0.001 \\
K\_Neighbours                        &                      1178.0 &  0.000 &                 1187.0 &  0.000 &    1867.5 &  0.067 \\
NN-12-layer\_wide\_with\_dropout       &                       435.0 &  0.000 &                  109.0 &  0.000 &      58.0 &  0.000 \\
NN-12-layer\_wide\_with\_dropout\_lr01  &                       229.0 &  0.000 &                   82.0 &  0.000 &      39.0 &  0.000 \\
NN-12-layer\_wide\_with\_dropout\_lr1   &                       173.0 &  0.000 &                   61.0 &  0.000 &      19.0 &  0.000 \\
NN-2-layer-droput-input-layer\_lr001 &                       776.0 &  0.000 &                  210.0 &  0.000 &     254.0 &  0.000 \\
NN-2-layer-droput-input-layer\_lr01  &                       200.0 &  0.000 &                   73.0 &  0.000 &      40.0 &  0.000 \\
NN-2-layer-droput-input-layer\_lr1   &                       288.0 &  0.000 &                  104.0 &  0.000 &      68.0 &  0.000 \\
NN-4-layer-droput-each-layer\_lr0001 &                      1205.0 &  0.000 &                  296.5 &  0.000 &     321.5 &  0.000 \\
NN-4-layer-droput-each-layer\_lr01   &                       207.0 &  0.000 &                   64.0 &  0.000 &      23.0 &  0.000 \\
NN-4-layer-droput-each-layer\_lr1    &                       208.0 &  0.000 &                   73.0 &  0.000 &      29.0 &  0.000 \\
NN-4-layer\_thin\_dropout             &                      1111.0 &  0.000 &                  198.0 &  0.000 &     197.0 &  0.000 \\
NN-4-layer\_thin\_dropout\_lr01        &                       191.0 &  0.000 &                   72.0 &  0.000 &      24.0 &  0.000 \\
NN-4-layer\_thin\_dropout\_lr1         &                       160.0 &  0.000 &                   59.0 &  0.000 &      23.0 &  0.000 \\
NN-4-layer\_wide\_no\_dropout          &                      1667.0 &  0.000 &                  359.0 &  0.000 &     422.0 &  0.000 \\
NN-4-layer\_wide\_no\_dropout\_lr01     &                       170.0 &  0.000 &                   59.0 &  0.000 &      22.0 &  0.000 \\
NN-4-layer\_wide\_no\_dropout\_lr1      &                       174.0 &  0.000 &                   64.0 &  0.000 &      23.0 &  0.000 \\
NN-4-layer\_wide\_with\_dropout        &                      1705.5 &  0.000 &                  412.0 &  0.000 &     501.5 &  0.000 \\
NN-4-layer\_wide\_with\_dropout\_lr01   &                       198.0 &  0.000 &                   77.0 &  0.000 &      28.0 &  0.000 \\
NN-4-layer\_wide\_with\_dropout\_lr1    &                       164.0 &  0.000 &                   69.0 &  0.000 &      23.0 &  0.000 \\
PassiveAggressiveClassifier         &                         0.0 &    NaN &                  668.0 &  0.000 &     601.5 &  0.000 \\
RandomForestClassifier              &                       668.0 &  0.000 &                    0.0 &    NaN &    1828.5 &  0.008 \\
SVC                                 &                       601.5 &  0.000 &                 1828.5 &  0.008 &       0.0 &    NaN \\
\bottomrule
\end{tabular}

%% file: tables/nemeniy_test1.tex
\begin{tabular}{lrrrr}
\toprule
{} &  BaggingClassifier &  BaselineClassifier &  BernoulliNaiveBayes &  GaussianNaiveBayes \\
\midrule
BaggingClassifier                   &             -1.000 &               0.000 &                0.938 &               0.597 \\
BaselineClassifier                  &              0.000 &              -1.000 &                0.000 &               0.000 \\
BernoulliNaiveBayes                 &              0.938 &               0.000 &               -1.000 &               1.000 \\
GaussianNaiveBayes                  &              0.597 &               0.000 &                1.000 &              -1.000 \\
GradientBoostingClassifier          &              1.000 &               0.000 &                1.000 &               0.973 \\
K\_Neighbours                        &              1.000 &               0.000 &                0.988 &               0.822 \\
NN-12-layer\_wide\_with\_dropout       &              0.000 &               0.986 &                0.509 &               0.903 \\
NN-12-layer\_wide\_with\_dropout\_lr01  &              0.000 &               1.000 &                0.010 &               0.121 \\
NN-12-layer\_wide\_with\_dropout\_lr1   &              0.000 &               1.000 &                0.002 &               0.036 \\
NN-2-layer-droput-input-layer\_lr001 &              0.338 &               0.002 &                1.000 &               1.000 \\
NN-2-layer-droput-input-layer\_lr01  &              0.000 &               0.969 &                0.624 &               0.947 \\
NN-2-layer-droput-input-layer\_lr1   &              0.000 &               1.000 &                0.149 &               0.567 \\
NN-4-layer-droput-each-layer\_lr0001 &              0.574 &               0.000 &                1.000 &               1.000 \\
NN-4-layer-droput-each-layer\_lr01   &              0.000 &               1.000 &                0.016 &               0.163 \\
NN-4-layer-droput-each-layer\_lr1    &              0.000 &               1.000 &                0.025 &               0.212 \\
NN-4-layer\_thin\_dropout             &              0.308 &               0.002 &                1.000 &               1.000 \\
NN-4-layer\_thin\_dropout\_lr01        &              0.000 &               1.000 &                0.048 &               0.314 \\
NN-4-layer\_thin\_dropout\_lr1         &              0.000 &               1.000 &                0.003 &               0.054 \\
NN-4-layer\_wide\_no\_dropout          &              0.946 &               0.000 &                1.000 &               1.000 \\
NN-4-layer\_wide\_no\_dropout\_lr01     &              0.000 &               1.000 &                0.032 &               0.245 \\
NN-4-layer\_wide\_no\_dropout\_lr1      &              0.000 &               1.000 &                0.024 &               0.209 \\
NN-4-layer\_wide\_with\_dropout        &              0.942 &               0.000 &                1.000 &               1.000 \\
NN-4-layer\_wide\_with\_dropout\_lr01   &              0.000 &               1.000 &                0.008 &               0.100 \\
NN-4-layer\_wide\_with\_dropout\_lr1    &              0.000 &               1.000 &                0.015 &               0.158 \\
PassiveAggressiveClassifier         &              1.000 &               0.000 &                1.000 &               1.000 \\
RandomForestClassifier              &              1.000 &               0.000 &                0.871 &               0.447 \\
SVC                                 &              1.000 &               0.000 &                0.977 &               0.749 \\
\bottomrule
\end{tabular}

%% file: tables/nemeniy_test2.tex
\begin{tabular}{lrrrr}
\toprule
{} &  GradientBoostingClassifier &  K\_Neighbours &  NN-12-layer\_wide\_with\_dropout &  NN-12-layer\_wide\_with\_dropout\_lr01 \\
\midrule
BaggingClassifier                   &                       1.000 &         1.000 &                          0.000 &                               0.000 \\
BaselineClassifier                  &                       0.000 &         0.000 &                          0.986 &                               1.000 \\
BernoulliNaiveBayes                 &                       1.000 &         0.988 &                          0.509 &                               0.010 \\
GaussianNaiveBayes                  &                       0.973 &         0.822 &                          0.903 &                               0.121 \\
GradientBoostingClassifier          &                      -1.000 &         1.000 &                          0.000 &                               0.000 \\
K\_Neighbours                        &                       1.000 &        -1.000 &                          0.000 &                               0.000 \\
NN-12-layer\_wide\_with\_dropout       &                       0.000 &         0.000 &                         -1.000 &                               1.000 \\
NN-12-layer\_wide\_with\_dropout\_lr01  &                       0.000 &         0.000 &                          1.000 &                              -1.000 \\
NN-12-layer\_wide\_with\_dropout\_lr1   &                       0.000 &         0.000 &                          1.000 &                               1.000 \\
NN-2-layer-droput-input-layer\_lr001 &                       0.885 &         0.594 &                          0.979 &                               0.299 \\
NN-2-layer-droput-input-layer\_lr01  &                       0.000 &         0.000 &                          1.000 &                               1.000 \\
NN-2-layer-droput-input-layer\_lr1   &                       0.000 &         0.000 &                          1.000 &                               1.000 \\
NN-4-layer-droput-each-layer\_lr0001 &                       0.969 &         0.806 &                          0.914 &                               0.133 \\
NN-4-layer-droput-each-layer\_lr01   &                       0.000 &         0.000 &                          1.000 &                               1.000 \\
NN-4-layer-droput-each-layer\_lr1    &                       0.000 &         0.000 &                          1.000 &                               1.000 \\
NN-4-layer\_thin\_dropout             &                       0.867 &         0.561 &                          0.983 &                               0.329 \\
NN-4-layer\_thin\_dropout\_lr01        &                       0.000 &         0.000 &                          1.000 &                               1.000 \\
NN-4-layer\_thin\_dropout\_lr1         &                       0.000 &         0.000 &                          1.000 &                               1.000 \\
NN-4-layer\_wide\_no\_dropout          &                       1.000 &         0.990 &                          0.486 &                               0.009 \\
NN-4-layer\_wide\_no\_dropout\_lr01     &                       0.000 &         0.000 &                          1.000 &                               1.000 \\
NN-4-layer\_wide\_no\_dropout\_lr1      &                       0.000 &         0.000 &                          1.000 &                               1.000 \\
NN-4-layer\_wide\_with\_dropout        &                       1.000 &         0.990 &                          0.496 &                               0.009 \\
NN-4-layer\_wide\_with\_dropout\_lr01   &                       0.000 &         0.000 &                          1.000 &                               1.000 \\
NN-4-layer\_wide\_with\_dropout\_lr1    &                       0.000 &         0.000 &                          1.000 &                               1.000 \\
PassiveAggressiveClassifier         &                       1.000 &         1.000 &                          0.006 &                               0.000 \\
RandomForestClassifier              &                       1.000 &         1.000 &                          0.000 &                               0.000 \\
SVC                                 &                       1.000 &         1.000 &                          0.000 &                               0.000 \\
\bottomrule
\end{tabular}

%% file: tables/nemeniy_test3.tex
\begin{tabular}{lrrrr}
\toprule
{} &  NN-12-layer\_wide\_with\_dropout\_lr1 &  NN-2-layer-droput-input-layer\_lr001 &  NN-2-layer-droput-input-layer\_lr01 &  NN-2-layer-droput-input-layer\_lr1 \\
\midrule
BaggingClassifier                   &                              0.000 &                                0.338 &                               0.000 &                              0.000 \\
BaselineClassifier                  &                              1.000 &                                0.002 &                               0.969 &                              1.000 \\
BernoulliNaiveBayes                 &                              0.002 &                                1.000 &                               0.624 &                              0.149 \\
GaussianNaiveBayes                  &                              0.036 &                                1.000 &                               0.947 &                              0.567 \\
GradientBoostingClassifier          &                              0.000 &                                0.885 &                               0.000 &                              0.000 \\
K\_Neighbours                        &                              0.000 &                                0.594 &                               0.000 &                              0.000 \\
NN-12-layer\_wide\_with\_dropout       &                              1.000 &                                0.979 &                               1.000 &                              1.000 \\
NN-12-layer\_wide\_with\_dropout\_lr01  &                              1.000 &                                0.299 &                               1.000 &                              1.000 \\
NN-12-layer\_wide\_with\_dropout\_lr1   &                             -1.000 &                                0.122 &                               1.000 &                              1.000 \\
NN-2-layer-droput-input-layer\_lr001 &                              0.122 &                               -1.000 &                               0.991 &                              0.803 \\
NN-2-layer-droput-input-layer\_lr01  &                              1.000 &                                0.991 &                              -1.000 &                              1.000 \\
NN-2-layer-droput-input-layer\_lr1   &                              1.000 &                                0.803 &                               1.000 &                             -1.000 \\
NN-4-layer-droput-each-layer\_lr0001 &                              0.041 &                                1.000 &                               0.953 &                              0.590 \\
NN-4-layer-droput-each-layer\_lr01   &                              1.000 &                                0.369 &                               1.000 &                              1.000 \\
NN-4-layer-droput-each-layer\_lr1    &                              1.000 &                                0.441 &                               1.000 &                              1.000 \\
NN-4-layer\_thin\_dropout             &                              0.139 &                                1.000 &                               0.993 &                              0.827 \\
NN-4-layer\_thin\_dropout\_lr01        &                              1.000 &                                0.570 &                               1.000 &                              1.000 \\
NN-4-layer\_thin\_dropout\_lr1         &                              1.000 &                                0.164 &                               1.000 &                              1.000 \\
NN-4-layer\_wide\_no\_dropout          &                              0.001 &                                1.000 &                               0.601 &                              0.136 \\
NN-4-layer\_wide\_no\_dropout\_lr01     &                              1.000 &                                0.487 &                               1.000 &                              1.000 \\
NN-4-layer\_wide\_no\_dropout\_lr1      &                              1.000 &                                0.437 &                               1.000 &                              1.000 \\
NN-4-layer\_wide\_with\_dropout        &                              0.002 &                                1.000 &                               0.611 &                              0.141 \\
NN-4-layer\_wide\_with\_dropout\_lr01   &                              1.000 &                                0.260 &                               1.000 &                              1.000 \\
NN-4-layer\_wide\_with\_dropout\_lr1    &                              1.000 &                                0.361 &                               1.000 &                              1.000 \\
PassiveAggressiveClassifier         &                              0.000 &                                0.997 &                               0.011 &                              0.000 \\
RandomForestClassifier              &                              0.000 &                                0.216 &                               0.000 &                              0.000 \\
SVC                                 &                              0.000 &                                0.498 &                               0.000 &                              0.000 \\
\bottomrule
\end{tabular}

%% file: tables/nemeniy_test4.tex
\begin{tabular}{lrrrr}
\toprule
{} &  NN-4-layer-droput-each-layer\_lr0001 &  NN-4-layer-droput-each-layer\_lr01 &  NN-4-layer-droput-each-layer\_lr1 &  NN-4-layer\_thin\_dropout \\
\midrule
BaggingClassifier                   &                                0.574 &                              0.000 &                             0.000 &                    0.308 \\
BaselineClassifier                  &                                0.000 &                              1.000 &                             1.000 &                    0.002 \\
BernoulliNaiveBayes                 &                                1.000 &                              0.016 &                             0.025 &                    1.000 \\
GaussianNaiveBayes                  &                                1.000 &                              0.163 &                             0.212 &                    1.000 \\
GradientBoostingClassifier          &                                0.969 &                              0.000 &                             0.000 &                    0.867 \\
K\_Neighbours                        &                                0.806 &                              0.000 &                             0.000 &                    0.561 \\
NN-12-layer\_wide\_with\_dropout       &                                0.914 &                              1.000 &                             1.000 &                    0.983 \\
NN-12-layer\_wide\_with\_dropout\_lr01  &                                0.133 &                              1.000 &                             1.000 &                    0.329 \\
NN-12-layer\_wide\_with\_dropout\_lr1   &                                0.041 &                              1.000 &                             1.000 &                    0.139 \\
NN-2-layer-droput-input-layer\_lr001 &                                1.000 &                              0.369 &                             0.441 &                    1.000 \\
NN-2-layer-droput-input-layer\_lr01  &                                0.953 &                              1.000 &                             1.000 &                    0.993 \\
NN-2-layer-droput-input-layer\_lr1   &                                0.590 &                              1.000 &                             1.000 &                    0.827 \\
NN-4-layer-droput-each-layer\_lr0001 &                               -1.000 &                              0.177 &                             0.229 &                    1.000 \\
NN-4-layer-droput-each-layer\_lr01   &                                0.177 &                             -1.000 &                             1.000 &                    0.401 \\
NN-4-layer-droput-each-layer\_lr1    &                                0.229 &                              1.000 &                            -1.000 &                    0.475 \\
NN-4-layer\_thin\_dropout             &                                1.000 &                              0.401 &                             0.475 &                   -1.000 \\
NN-4-layer\_thin\_dropout\_lr01        &                                0.335 &                              1.000 &                             1.000 &                    0.604 \\
NN-4-layer\_thin\_dropout\_lr1         &                                0.060 &                              1.000 &                             1.000 &                    0.185 \\
NN-4-layer\_wide\_no\_dropout          &                                1.000 &                              0.014 &                             0.022 &                    1.000 \\
NN-4-layer\_wide\_no\_dropout\_lr01     &                                0.264 &                              1.000 &                             1.000 &                    0.521 \\
NN-4-layer\_wide\_no\_dropout\_lr1      &                                0.225 &                              1.000 &                             1.000 &                    0.470 \\
NN-4-layer\_wide\_with\_dropout        &                                1.000 &                              0.015 &                             0.023 &                    1.000 \\
NN-4-layer\_wide\_with\_dropout\_lr01   &                                0.110 &                              1.000 &                             1.000 &                    0.287 \\
NN-4-layer\_wide\_with\_dropout\_lr1    &                                0.172 &                              1.000 &                             1.000 &                    0.392 \\
PassiveAggressiveClassifier         &                                1.000 &                              0.000 &                             0.000 &                    0.996 \\
RandomForestClassifier              &                                0.424 &                              0.000 &                             0.000 &                    0.193 \\
SVC                                 &                                0.730 &                              0.000 &                             0.000 &                    0.464 \\
\bottomrule
\end{tabular}

%% file: tables/nemeniy_test5.tex
\begin{tabular}{lrrrr}
\toprule
{} &  NN-4-layer\_thin\_dropout\_lr01 &  NN-4-layer\_thin\_dropout\_lr1 &  NN-4-layer\_wide\_no\_dropout &  NN-4-layer\_wide\_no\_dropout\_lr01 \\
\midrule
BaggingClassifier                   &                         0.000 &                        0.000 &                       0.946 &                            0.000 \\
BaselineClassifier                  &                         1.000 &                        1.000 &                       0.000 &                            1.000 \\
BernoulliNaiveBayes                 &                         0.048 &                        0.003 &                       1.000 &                            0.032 \\
GaussianNaiveBayes                  &                         0.314 &                        0.054 &                       1.000 &                            0.245 \\
GradientBoostingClassifier          &                         0.000 &                        0.000 &                       1.000 &                            0.000 \\
K\_Neighbours                        &                         0.000 &                        0.000 &                       0.990 &                            0.000 \\
NN-12-layer\_wide\_with\_dropout       &                         1.000 &                        1.000 &                       0.486 &                            1.000 \\
NN-12-layer\_wide\_with\_dropout\_lr01  &                         1.000 &                        1.000 &                       0.009 &                            1.000 \\
NN-12-layer\_wide\_with\_dropout\_lr1   &                         1.000 &                        1.000 &                       0.001 &                            1.000 \\
NN-2-layer-droput-input-layer\_lr001 &                         0.570 &                        0.164 &                       1.000 &                            0.487 \\
NN-2-layer-droput-input-layer\_lr01  &                         1.000 &                        1.000 &                       0.601 &                            1.000 \\
NN-2-layer-droput-input-layer\_lr1   &                         1.000 &                        1.000 &                       0.136 &                            1.000 \\
NN-4-layer-droput-each-layer\_lr0001 &                         0.335 &                        0.060 &                       1.000 &                            0.264 \\
NN-4-layer-droput-each-layer\_lr01   &                         1.000 &                        1.000 &                       0.014 &                            1.000 \\
NN-4-layer-droput-each-layer\_lr1    &                         1.000 &                        1.000 &                       0.022 &                            1.000 \\
NN-4-layer\_thin\_dropout             &                         0.604 &                        0.185 &                       1.000 &                            0.521 \\
NN-4-layer\_thin\_dropout\_lr01        &                        -1.000 &                        1.000 &                       0.043 &                            1.000 \\
NN-4-layer\_thin\_dropout\_lr1         &                         1.000 &                       -1.000 &                       0.003 &                            1.000 \\
NN-4-layer\_wide\_no\_dropout          &                         0.043 &                        0.003 &                      -1.000 &                            0.028 \\
NN-4-layer\_wide\_no\_dropout\_lr01     &                         1.000 &                        1.000 &                       0.028 &                           -1.000 \\
NN-4-layer\_wide\_no\_dropout\_lr1      &                         1.000 &                        1.000 &                       0.021 &                            1.000 \\
NN-4-layer\_wide\_with\_dropout        &                         0.045 &                        0.003 &                       1.000 &                            0.029 \\
NN-4-layer\_wide\_with\_dropout\_lr01   &                         1.000 &                        1.000 &                       0.007 &                            1.000 \\
NN-4-layer\_wide\_with\_dropout\_lr1    &                         1.000 &                        1.000 &                       0.013 &                            1.000 \\
PassiveAggressiveClassifier         &                         0.000 &                        0.000 &                       1.000 &                            0.000 \\
RandomForestClassifier              &                         0.000 &                        0.000 &                       0.884 &                            0.000 \\
SVC                                 &                         0.000 &                        0.000 &                       0.981 &                            0.000 \\
\bottomrule
\end{tabular}

%% file: tables/nemeniy_test6.tex
\begin{tabular}{lrrrr}
\toprule
{} &  NN-4-layer\_wide\_no\_dropout\_lr1 &  NN-4-layer\_wide\_with\_dropout &  NN-4-layer\_wide\_with\_dropout\_lr01 &  NN-4-layer\_wide\_with\_dropout\_lr1 \\
\midrule
BaggingClassifier                   &                           0.000 &                         0.942 &                              0.000 &                             0.000 \\
BaselineClassifier                  &                           1.000 &                         0.000 &                              1.000 &                             1.000 \\
BernoulliNaiveBayes                 &                           0.024 &                         1.000 &                              0.008 &                             0.015 \\
GaussianNaiveBayes                  &                           0.209 &                         1.000 &                              0.100 &                             0.158 \\
GradientBoostingClassifier          &                           0.000 &                         1.000 &                              0.000 &                             0.000 \\
K\_Neighbours                        &                           0.000 &                         0.990 &                              0.000 &                             0.000 \\
NN-12-layer\_wide\_with\_dropout       &                           1.000 &                         0.496 &                              1.000 &                             1.000 \\
NN-12-layer\_wide\_with\_dropout\_lr01  &                           1.000 &                         0.009 &                              1.000 &                             1.000 \\
NN-12-layer\_wide\_with\_dropout\_lr1   &                           1.000 &                         0.002 &                              1.000 &                             1.000 \\
NN-2-layer-droput-input-layer\_lr001 &                           0.437 &                         1.000 &                              0.260 &                             0.361 \\
NN-2-layer-droput-input-layer\_lr01  &                           1.000 &                         0.611 &                              1.000 &                             1.000 \\
NN-2-layer-droput-input-layer\_lr1   &                           1.000 &                         0.141 &                              1.000 &                             1.000 \\
NN-4-layer-droput-each-layer\_lr0001 &                           0.225 &                         1.000 &                              0.110 &                             0.172 \\
NN-4-layer-droput-each-layer\_lr01   &                           1.000 &                         0.015 &                              1.000 &                             1.000 \\
NN-4-layer-droput-each-layer\_lr1    &                           1.000 &                         0.023 &                              1.000 &                             1.000 \\
NN-4-layer\_thin\_dropout             &                           0.470 &                         1.000 &                              0.287 &                             0.392 \\
NN-4-layer\_thin\_dropout\_lr01        &                           1.000 &                         0.045 &                              1.000 &                             1.000 \\
NN-4-layer\_thin\_dropout\_lr1         &                           1.000 &                         0.003 &                              1.000 &                             1.000 \\
NN-4-layer\_wide\_no\_dropout          &                           0.021 &                         1.000 &                              0.007 &                             0.013 \\
NN-4-layer\_wide\_no\_dropout\_lr01     &                           1.000 &                         0.029 &                              1.000 &                             1.000 \\
NN-4-layer\_wide\_no\_dropout\_lr1      &                          -1.000 &                         0.022 &                              1.000 &                             1.000 \\
NN-4-layer\_wide\_with\_dropout        &                           0.022 &                        -1.000 &                              0.007 &                             0.014 \\
NN-4-layer\_wide\_with\_dropout\_lr01   &                           1.000 &                         0.007 &                             -1.000 &                             1.000 \\
NN-4-layer\_wide\_with\_dropout\_lr1    &                           1.000 &                         0.014 &                              1.000 &                            -1.000 \\
PassiveAggressiveClassifier         &                           0.000 &                         1.000 &                              0.000 &                             0.000 \\
RandomForestClassifier              &                           0.000 &                         0.878 &                              0.000 &                             0.000 \\
SVC                                 &                           0.000 &                         0.979 &                              0.000 &                             0.000 \\
\bottomrule
\end{tabular}

%% file: tables/nemeniy_test7.tex
\begin{tabular}{lrrr}
\toprule
{} &  PassiveAggressiveClassifier &  RandomForestClassifier &    SVC \\
\midrule
BaggingClassifier                   &                        1.000 &                   1.000 &  1.000 \\
BaselineClassifier                  &                        0.000 &                   0.000 &  0.000 \\
BernoulliNaiveBayes                 &                        1.000 &                   0.871 &  0.977 \\
GaussianNaiveBayes                  &                        1.000 &                   0.447 &  0.749 \\
GradientBoostingClassifier          &                        1.000 &                   1.000 &  1.000 \\
K\_Neighbours                        &                        1.000 &                   1.000 &  1.000 \\
NN-12-layer\_wide\_with\_dropout       &                        0.006 &                   0.000 &  0.000 \\
NN-12-layer\_wide\_with\_dropout\_lr01  &                        0.000 &                   0.000 &  0.000 \\
NN-12-layer\_wide\_with\_dropout\_lr1   &                        0.000 &                   0.000 &  0.000 \\
NN-2-layer-droput-input-layer\_lr001 &                        0.997 &                   0.216 &  0.498 \\
NN-2-layer-droput-input-layer\_lr01  &                        0.011 &                   0.000 &  0.000 \\
NN-2-layer-droput-input-layer\_lr1   &                        0.000 &                   0.000 &  0.000 \\
NN-4-layer-droput-each-layer\_lr0001 &                        1.000 &                   0.424 &  0.730 \\
NN-4-layer-droput-each-layer\_lr01   &                        0.000 &                   0.000 &  0.000 \\
NN-4-layer-droput-each-layer\_lr1    &                        0.000 &                   0.000 &  0.000 \\
NN-4-layer\_thin\_dropout             &                        0.996 &                   0.193 &  0.464 \\
NN-4-layer\_thin\_dropout\_lr01        &                        0.000 &                   0.000 &  0.000 \\
NN-4-layer\_thin\_dropout\_lr1         &                        0.000 &                   0.000 &  0.000 \\
NN-4-layer\_wide\_no\_dropout          &                        1.000 &                   0.884 &  0.981 \\
NN-4-layer\_wide\_no\_dropout\_lr01     &                        0.000 &                   0.000 &  0.000 \\
NN-4-layer\_wide\_no\_dropout\_lr1      &                        0.000 &                   0.000 &  0.000 \\
NN-4-layer\_wide\_with\_dropout        &                        1.000 &                   0.878 &  0.979 \\
NN-4-layer\_wide\_with\_dropout\_lr01   &                        0.000 &                   0.000 &  0.000 \\
NN-4-layer\_wide\_with\_dropout\_lr1    &                        0.000 &                   0.000 &  0.000 \\
PassiveAggressiveClassifier         &                       -1.000 &                   1.000 &  1.000 \\
RandomForestClassifier              &                        1.000 &                  -1.000 &  1.000 \\
SVC                                 &                        1.000 &                   1.000 & -1.000 \\
\bottomrule
\end{tabular}

%% file: mlaut.bbl
\begin{thebibliography}{47}
\providecommand{\natexlab}[1]{#1}
\providecommand{\url}[1]{\texttt{#1}}
\expandafter\ifx\csname urlstyle\endcsname\relax
  \providecommand{\doi}[1]{doi: #1}\else
  \providecommand{\doi}{doi: \begingroup \urlstyle{rm}\Url}\fi

\bibitem[skl()]{skll}
scikit-learn laboratory.
\newblock \url{https://skll.readthedocs.io}.
\newblock URL \url{https://skll.readthedocs.io}.

\bibitem[Abadi et~al.(2015)Abadi, Agarwal, Barham, Brevdo, Chen, Citro,
  Corrado, Davis, Dean, Devin, Ghemawat, Goodfellow, Harp, Irving, Isard, Jia,
  Jozefowicz, Kaiser, Kudlur, Levenberg, Man\'{e}, Monga, Moore, Murray, Olah,
  Schuster, Shlens, Steiner, Sutskever, Talwar, Tucker, Vanhoucke, Vasudevan,
  Vi\'{e}gas, Vinyals, Warden, Wattenberg, Wicke, Yu, and
  Zheng]{martin_abadi_tensorflow:_2015}
Mart\'{\i}n Abadi, Ashish Agarwal, Paul Barham, Eugene Brevdo, Zhifeng Chen,
  Craig Citro, Greg~S. Corrado, Andy Davis, Jeffrey Dean, Matthieu Devin,
  Sanjay Ghemawat, Ian Goodfellow, Andrew Harp, Geoffrey Irving, Michael Isard,
  Yangqing Jia, Rafal Jozefowicz, Lukasz Kaiser, Manjunath Kudlur, Josh
  Levenberg, Dandelion Man\'{e}, Rajat Monga, Sherry Moore, Derek Murray, Chris
  Olah, Mike Schuster, Jonathon Shlens, Benoit Steiner, Ilya Sutskever, Kunal
  Talwar, Paul Tucker, Vincent Vanhoucke, Vijay Vasudevan, Fernanda Vi\'{e}gas,
  Oriol Vinyals, Pete Warden, Martin Wattenberg, Martin Wicke, Yuan Yu, and
  Xiaoqiang Zheng.
\newblock {TensorFlow}: Large-scale machine learning on heterogeneous systems,
  2015.

\bibitem[Ba and Caruana(2013)]{ba_deep_2013}
Lei~Jimmy Ba and Rich Caruana.
\newblock Do {Deep} {Nets} {Really} {Need} to be {Deep}?
\newblock \emph{arXiv:1312.6184 [cs]}, 2013.

\bibitem[Bengio(2012)]{bengio_practical_2012}
Yoshua Bengio.
\newblock Practical recommendations for gradient-based training of deep
  architectures.
\newblock \emph{arXiv:1206.5533 [cs]}, 2012.

\bibitem[Bergstra and Bengio(2012)]{bergstra_random_2012}
James Bergstra and Yoshua Bengio.
\newblock Random {Search} for {Hyper}-{Parameter} {Optimization}.
\newblock \emph{Journal of Machine Learning Research}, 2012.

\bibitem[Bischl et~al.(2016)Bischl, Lang, Kotthoff, Schiffner, Richter,
  Studerus, Casalicchio, and Jones]{bischl_mlr:_2016}
Bernd Bischl, Michel Lang, Lars Kotthoff, Julia Schiffner, Jakob Richter, Erich
  Studerus, Giuseppe Casalicchio, and Zachary~M. Jones.
\newblock mlr: Machine learning in r.
\newblock \emph{Journal of Machine Learning Research}, 17\penalty0
  (170):\penalty0 1--5, 2016.
\newblock URL \url{http://www.jmlr.org/papers/v17/15-066.html}.

\bibitem[Bishop(2006)]{bishop_pattern_2006}
Christopher Bishop.
\newblock \emph{Pattern {Recognition} and {Machine} {Learning}}.
\newblock Springer-Verlag New York, 2006.
\newblock ISBN 78-1-4939-3843-8.

\bibitem[Breiman(1996)]{breiman_bagging_1996}
Leo Breiman.
\newblock Bagging predictors.
\newblock \emph{Machine Learning}, 1996.

\bibitem[Breiman(2001)]{breiman_random_2001}
Leo Breiman.
\newblock Random {Forests}.
\newblock \emph{Machine Learning}, 2001.

\bibitem[Buitinck et~al.(2013)Buitinck, Louppe, Blondel, Pedregosa, Mueller,
  Grisel, Niculae, Prettenhofer, Gramfort, Grobler, Layton, Vanderplas, Joly,
  Holt, and Varoquaux]{buitinck_api_2013}
Lars Buitinck, Gilles Louppe, Mathieu Blondel, Fabian Pedregosa, Andreas
  Mueller, Olivier Grisel, Vlad Niculae, Peter Prettenhofer, Alexandre
  Gramfort, Jaques Grobler, Robert Layton, Jake Vanderplas, Arnaud Joly, Brian
  Holt, and Gaël Varoquaux.
\newblock Api design for machine learning software: experiences from the
  scikit-learn project.
\newblock \emph{CoRR}, 2013.

\bibitem[Chen et~al.(2015)Chen, Li, Li, Lin, Wang, Wang, Xiao, Xu, Zhang, and
  Zhang]{chen_mxnet:_2015}
Tianqi Chen, Mu~Li, Yutian Li, Min Lin, Naiyan Wang, Minjie Wang, Tianjun Xiao,
  Bing Xu, Chiyuan Zhang, and Zheng Zhang.
\newblock {MXNet}: {A} {Flexible} and {Efficient} {Machine} {Learning}
  {Library} for {Heterogeneous} {Distributed} {Systems}.
\newblock \emph{arXiv:1512.01274 [cs]}, 2015.

\bibitem[Chollet(2015)]{chollet_keras_2015}
François Chollet.
\newblock Keras, 2015.
\newblock URL \url{https://keras.io}.

\bibitem[Cortes and Vapnik(1995)]{cortes_support-vector_1995}
Corinna Cortes and Vladimir Vapnik.
\newblock Support-{Vector} {Networks}.
\newblock \emph{Machine Learning}, 1995.

\bibitem[Cover and Hart(1967)]{cover_nearest_1967}
Thomas Cover and Peter Hart.
\newblock Nearest neighbor pattern classification.
\newblock \emph{IEEE Transactions on Information Theory}, 1967.

\bibitem[Crammer et~al.(2006)Crammer, Dekel, Keshet, Shalev-Shwartz, and
  Singer]{crammer_online_2006}
Koby Crammer, Ofer Dekel, Joseph Keshet, Shai Shalev-Shwartz, and Yoram Singer.
\newblock Online passive-aggressive algorithms.
\newblock \emph{The Journal of Machine Learning Research}, 2006.

\bibitem[Demšar(2006)]{demsar_statistical_2006}
Janez Demšar.
\newblock Statistical comparisons of classifiers over multiple data sets.
\newblock \emph{Journal of Machine Learning Research}, 2006.

\bibitem[Demšar et~al.(2013)Demšar, Curk, Erjavec, Gorup, Hočevar,
  Milutinovič, Možina, Polajnar, Toplak, Starič, Štajdohar, Umek, Žagar,
  Žbontar, Žitnik, and Zupan]{demsar_orange:_2013}
Janez Demšar, Tomaž Curk, Aleš Erjavec, Črt Gorup, Tomaž Hočevar, Mitar
  Milutinovič, Martin Možina, Matija Polajnar, Marko Toplak, Anže Starič,
  Miha Štajdohar, Lan Umek, Lan Žagar, Jure Žbontar, Marinka Žitnik, and
  Blaž Zupan.
\newblock Orange: {Data} {Mining} {Toolbox} in {Python}.
\newblock \emph{Journal of Machine Learning Research}, 2013.

\bibitem[Efron and Hastie(2016)]{efron_computer_2016}
Bradley Efron and Trevor Hastie.
\newblock \emph{Computer {Age} {Statistical} {Inference}: {Algorithms},
  {Evidence} and {Data} {Science}}.
\newblock Institute of {Mathematical} {Statistics} {Monographs}. Cambridge
  University Press, Cambridge, 2016.

\bibitem[Fernández-Delgado et~al.(2014)Fernández-Delgado, Cernadas, Barro,
  and Amorim]{fernandez-delgado_we_2014}
Manuel Fernández-Delgado, Eva Cernadas, Senén Barro, and Dinani Amorim.
\newblock Do we {Need} {Hundreds} of {Classifiers} to {Solve} {Real} {World}
  {Classification} {Problems}?
\newblock \emph{Journal of Machine Learning Research}, 2014.

\bibitem[Feurer et~al.(2015)Feurer, Klein, Eggensperger, Springenberg, Blum,
  and Hutter]{feurer2015efficient}
Matthias Feurer, Aaron Klein, Katharina Eggensperger, Jost Springenberg, Manuel
  Blum, and Frank Hutter.
\newblock Efficient and robust automated machine learning.
\newblock In \emph{Advances in Neural Information Processing Systems}, pages
  2962--2970, 2015.

\bibitem[Friedman(2001)]{friedman_greedy_2001}
Jerome Friedman.
\newblock Greedy {Function} {Approximation}: {A} {Gradient} {Boosting}
  {Machine}.
\newblock \emph{The Annals of Statistics}, 2001.

\bibitem[Goodfellow et~al.(2016)Goodfellow, Bengio, and
  Courville]{goodfellow_deep_2016}
Ian Goodfellow, Yoshua Bengio, and Aaron Courville.
\newblock \emph{Deep Learning}.
\newblock {MIT} Press, 2016.

\bibitem[Gupta and Raza(2018)]{gupta_optimizing_2018}
Tarun~Kumar Gupta and Khalid Raza.
\newblock Optimizing {Deep} {Neural} {Network} {Architecture}: {A} {Tabu}
  {Search} {Based} {Approach}.
\newblock \emph{arXiv:1808.05979 [cs, stat]}, 2018.

\bibitem[Hall et~al.(2009)Hall, Frank, Holmes, Pfahringer, Reutemann, and
  Witten]{hall2009weka}
Mark Hall, Eibe Frank, Geoffrey Holmes, Bernhard Pfahringer, Peter Reutemann,
  and Ian~H Witten.
\newblock The weka data mining software: an update.
\newblock \emph{ACM SIGKDD explorations newsletter}, 11\penalty0 (1):\penalty0
  10--18, 2009.

\bibitem[Hasanpour et~al.(2016)Hasanpour, Rouhani, Fayyaz, and
  Sabokrou]{hasanpour_lets_2016}
Seyyed~Hossein Hasanpour, Mohammad Rouhani, Mohsen Fayyaz, and Mohammad
  Sabokrou.
\newblock Lets keep it simple, {Using} simple architectures to outperform
  deeper and more complex architectures.
\newblock \emph{arXiv:1608.06037 [cs]}, 2016.

\bibitem[Hinton et~al.(2012)Hinton, Srivastava, Krizhevsky, Sutskever, and
  Salakhutdinov]{hinton_improving_2012}
Geoffrey Hinton, Nitish Srivastava, Alex Krizhevsky, Ilya Sutskever, and Ruslan
  Salakhutdinov.
\newblock Improving neural networks by preventing co-adaptation of feature
  detectors.
\newblock \emph{arXiv:1207.0580 [cs]}, 2012.

\bibitem[Hsu et~al.(2003)Hsu, Chang, and Lin]{hsu_practical_2016}
Chih-Wei Hsu, Chih-Chung Chang, and Chih-Jen Lin.
\newblock A {Practical} {Guide} to {Support} {Vector} {Classification}.
\newblock page~16, 2003.

\bibitem[Huang et~al.(2003)Huang, Lu, and Ling]{huang_comparing_2003}
Jin Huang, Jingjing Lu, and Charles Ling.
\newblock Comparing naive {Bayes}, decision trees, and {SVM} with {AUC} and
  accuracy.
\newblock IEEE Comput. Soc, 2003.

\bibitem[Jagtap and Kodge(2013)]{jagtap_census_2013}
Sudhir Jagtap and Bheemashankar Kodge.
\newblock Census {Data} {Mining} and {Data} {Analysis} using {WEKA}.
\newblock \emph{arXiv:1310.4647 [cs]}, 2013.

\bibitem[James et~al.(2013)James, Witten, Hastie, and
  Tibshirani]{james_introduction_2013}
Gareth James, Daniela Witten, Trevor Hastie, and Robert Tibshirani.
\newblock \emph{Introduction to {Statistical} {Learning}}.
\newblock Springer Publishing Company, Incorporated, 2013.
\newblock ISBN 978-1-4614-7137-0.

\bibitem[Jia et~al.(2014)Jia, Shelhamer, Donahue, Karayev, Long, Girshick,
  Guadarrama, and Darrell]{shelhamer_caffe_nodate}
Yangqing Jia, Evan Shelhamer, Jeff Donahue, Sergey Karayev, Jonathan Long, Ross
  Girshick, Sergio Guadarrama, and Trevor Darrell.
\newblock Caffe: Convolutional architecture for fast feature embedding, 2014.

\bibitem[Kazakov and Kir\'{a}ly(2018)]{mlaut_2018}
Viktor Kazakov and Franz Kir\'{a}ly.
\newblock mlaut: {Machine Learning} automation toolbox, 2018.
\newblock URL \url{https://github.com/alan-turing-institute/mlaut}.

\bibitem[Krizhevsky et~al.(2017)Krizhevsky, Sutskever, and
  Hinton]{krizhevsky_imagenet_2017}
Alex Krizhevsky, Ilya Sutskever, and Geoffrey~E. Hinton.
\newblock {ImageNet} classification with deep convolutional neural networks.
\newblock \emph{Communications of the ACM}, 2017.

\bibitem[Ojha et~al.(2017)Ojha, Abraham, and Snášel]{ojha_metaheuristic_2017}
Varun~Kumar Ojha, Ajith Abraham, and Václav Snášel.
\newblock Metaheuristic {Design} of {Feedforward} {Neural} {Networks}: {A}
  {Review} of {Two} {Decades} of {Research}.
\newblock \emph{Engineering Applications of Artificial Intelligence}, 2017.

\bibitem[Pedregosa et~al.(2011)Pedregosa, Varoquaux, Gramfort, Michel, Thirion,
  Grisel, Blondel, Prettenhofer, Weiss, Dubourg, Vanderplas, Passos,
  Cournapeau, Brucher, Perrot, and Duchesnay]{pedregosa_scikit-learn:_2011}
Fabian Pedregosa, Gaël Varoquaux, Alexandre Gramfort, Vincent Michel, Bertrand
  Thirion, Olivier Grisel, Mathieu Blondel, Peter Prettenhofer, Ron Weiss,
  Vincent Dubourg, Jake Vanderplas, Alexandre Passos, David Cournapeau,
  Matthieu Brucher, Matthieu Perrot, and Édouard Duchesnay.
\newblock Scikit-learn: {Machine} {Learning} in {Python}.
\newblock \emph{Journal of Machine Learning Research}, 2011.

\bibitem[Probst et~al.(2018)Probst, Bischl, and
  Boulesteix]{probst_tunability:_2018}
Philipp Probst, Bernd Bischl, and Anne-Laure Boulesteix.
\newblock Tunability: {Importance} of {Hyperparameters} of {Machine} {Learning}
  {Algorithms}.
\newblock \emph{{arXiv}:1802.09596 [stat]}, 2018.

\bibitem[Ross(2010)]{ross_introductory_2010}
Sheldon~M Ross.
\newblock \emph{Introductory Statistics - 3rd Edition}.
\newblock Academic Press, 2010.

\bibitem[Sansone and De~Natale(2017)]{sansone_training_2017}
Emanuele Sansone and Francesco G.~B. De~Natale.
\newblock Training {Feedforward} {Neural} {Networks} with {Standard} {Logistic}
  {Activations} is {Feasible}.
\newblock \emph{arXiv:1710.01013 [cs, stat]}, 2017.

\bibitem[Scikit-Learn(2018)]{scikit-learn_model_2018}
Scikit-Learn.
\newblock Model selection: choosing estimators and their parameters —
  scikit-learn 0.20.0 documentation, 2018.

\bibitem[Seide and Agarwal(2016)]{seide_cntk:_2016}
Frank Seide and Amit Agarwal.
\newblock {CNTK}: {Microsoft}'s {Open}-{Source} {Deep}-{Learning} {Toolkit}.
\newblock In \emph{Proceedings of the 22Nd {ACM} {SIGKDD} {International}
  {Conference} on {Knowledge} {Discovery} and {Data} {Mining}}, {KDD} '16,
  pages 2135--2135, New York, NY, USA, 2016. ACM.
\newblock ISBN 978-1-4503-4232-2.
\newblock \doi{10.1145/2939672.2945397}.

\bibitem[Sonnenburg et~al.(2010)Sonnenburg, Rätsch, Henschel, Widmer, Behr,
  Zien, Bona, Binder, Gehl, and Franc]{sonnenburg_shogun_2010}
Sören Sonnenburg, Gunnar Rätsch, Sebastian Henschel, Christian Widmer, Jonas
  Behr, Alexander Zien, Fabio~de Bona, Alexander Binder, Christian Gehl, and
  Vojtěch Franc.
\newblock The {SHOGUN} {Machine} {Learning} {Toolbox}.
\newblock \emph{Journal of Machine Learning Research}, pages 1799--1802, 2010.

\bibitem[Srivastava et~al.(2014)Srivastava, Hinton, Krizhevsky, Sutskever, and
  Salakhutdinov]{srivastava_dropout:_2014}
Nitish Srivastava, Geoffrey Hinton, Alex Krizhevsky, Ilya Sutskever, and Ruslan
  Salakhutdinov.
\newblock Dropout: {A} {Simple} {Way} to {Prevent} {Neural} {Networks} from
  {Overfitting}.
\newblock \emph{Journal of Machine Learning Research}, 2014.

\bibitem[Terpilowski(2018)]{terpilowski_scikit-posthocs:_nodate}
Maksim Terpilowski.
\newblock scikit-posthocs: Statistical post-hoc analysis and outlier detection
  algorithms, 2018.
\newblock URL \url{http://github.com/maximtrp/scikit-posthocs}.

\bibitem[Thomas et~al.(2018)Thomas, Coors, and Bischl]{thomas2018automatic}
Janek Thomas, Stefan Coors, and Bernd Bischl.
\newblock Automatic gradient boosting.
\newblock \emph{arXiv preprint arXiv:1807.03873}, 2018.

\bibitem[Wainer(2016)]{wainer_comparison_2016}
Jacques Wainer.
\newblock Comparison of 14 different families of classification algorithms on
  115 binary datasets.
\newblock \emph{arXiv:1606.00930 [cs]}, 2016.

\bibitem[Wilcoxon(1945)]{wilcoxon_individual_1945}
Frank Wilcoxon.
\newblock Individual comparisons by ranking methods.
\newblock \emph{Biometrics Bulletin}, 1945.

\bibitem[Wing et~al.(2018)Wing, Weston, Williams, Keefer, Engelhardt, Cooper,
  Mayer, Kenkel, Team, Benesty, Lescarbeau, Ziem, Scrucca, Tang, Candan, and
  Hunt]{wing_caret:_2018}
Max Kuhn Contributions from~Jed Wing, Steve Weston, Andre Williams, Chris
  Keefer, Allan Engelhardt, Tony Cooper, Zachary Mayer, Brenton Kenkel, the
  R.~Core Team, Michael Benesty, Reynald Lescarbeau, Andrew Ziem, Luca Scrucca,
  Yuan Tang, Can Candan, and {and}~Tyler Hunt.
\newblock caret: {Classification} and {Regression} {Training}, 2018.

\end{thebibliography}
